\pdfoutput=1

\documentclass[11pt]{article}

\usepackage[preprint]{acl}

\usepackage{times}
\usepackage{latexsym}

\usepackage[T1]{fontenc}

\usepackage[utf8]{inputenc}

\usepackage{microtype}

\usepackage{inconsolata}

\usepackage{graphicx}
\usepackage{xcolor}

\usepackage{amsmath}
\usepackage{hyperref}
\usepackage{float}
\usepackage{multicol}
\usepackage{array}
\usepackage{tabularx}
\usepackage{longtable}
\usepackage{geometry}
\usepackage{makecell}
\usepackage{multirow}
\usepackage{booktabs}
\usepackage{tikz}
\usepackage{graphicx}
\usepackage{subcaption}
\usepackage{caption}
\usepackage{xcolor}
\usepackage{soul}
\usepackage{listings}
\lstdefinestyle{MyListingStyle_0}{
    basicstyle=\normalsize\ttfamily, 
    frame=single, 
    breaklines=true, 
    breakindent=0pt, 
    xleftmargin=10pt, 
    moredelim=**[is][\color{blue}]{@}{@}, 
    moredelim=**[is][\color{red}]{!}{!},
    moredelim=**[is][\color{orange}]{+}{+},
}
\lstdefinestyle{MyListingStyle_1}{
    basicstyle=\scriptsize\ttfamily, 
    frame=single, 
    breaklines=true, 
    breakindent=0pt, 
    xleftmargin=10pt, 
    moredelim=**[is][\color{blue}]{@}{@}, 
    moredelim=**[is][\color{red}]{!}{!},
    moredelim=**[is][\color{orange}]{+}{+},
}

\title{How Much Do Large Language Models Know about Human Motion?\\A Case Study in 3D Avatar Control}

\author{%
  Kunhang Li${}^{1}$ Jason Naradowsky${}^{1}$ Yansong Feng${}^{2}$ Yusuke Miyao${}^{1,3}$\\
${}^{1}$The University of Tokyo ${}^{2}$Peking University ${}^{3}$NII LLMC\\
\texttt{\{kunhangli, narad, yusuke\}@is.s.u-tokyo.ac.jp}\\\texttt{fengyansong@pku.edu.cn}}

\begin{document}
\maketitle
\begin{abstract}
We explore the human motion knowledge of Large Language Models (LLMs) through 3D avatar control. Given a motion instruction, we prompt LLMs to first generate a high-level movement plan with consecutive steps (\textbf{High-level Planning}), then specify body part positions in each step (\textbf{Low-level Planning}), which we linearly interpolate into avatar animations. Using 20 representative motion instructions that cover fundamental movements and balance body part usage, we conduct comprehensive evaluations, including human and automatic scoring of both high-level movement plans and generated animations, as well as automatic comparison with oracle positions in low-level planning. Our findings show that LLMs are strong at interpreting high-level body movements but struggle with precise body part positioning. While decomposing motion queries into atomic components improves planning, LLMs face challenges in multi-step movements involving high-degree-of-freedom body parts. Furthermore, LLMs provide reasonable approximations for general spatial descriptions, but fall short in handling precise spatial specifications. Notably, LLMs demonstrate promise in conceptualizing creative motions and distinguishing culturally specific motion patterns.~\footnote{\href{https://github.com/KunhangL/MotionDecomposition}{https://github.com/KunhangL/MotionDecomposition}}
\end{abstract}

\section{Introduction}

Recent approaches in text-conditioned human motion generation attempt to improve the generalization to unseen instructions by leveraging Large Language Models (LLMs) to extract motion-relevant information, such as active body parts \citep{SINC}, detailed body part descriptions \citep{huang_control}, and keyframe coordinates \citep{text2animation}. However, these methods only utilize LLMs as auxiliary components, leaving the extent of their human motion knowledge largely unexplored.

\begin{figure}[t]
    \centering
    \includegraphics[width=\columnwidth]{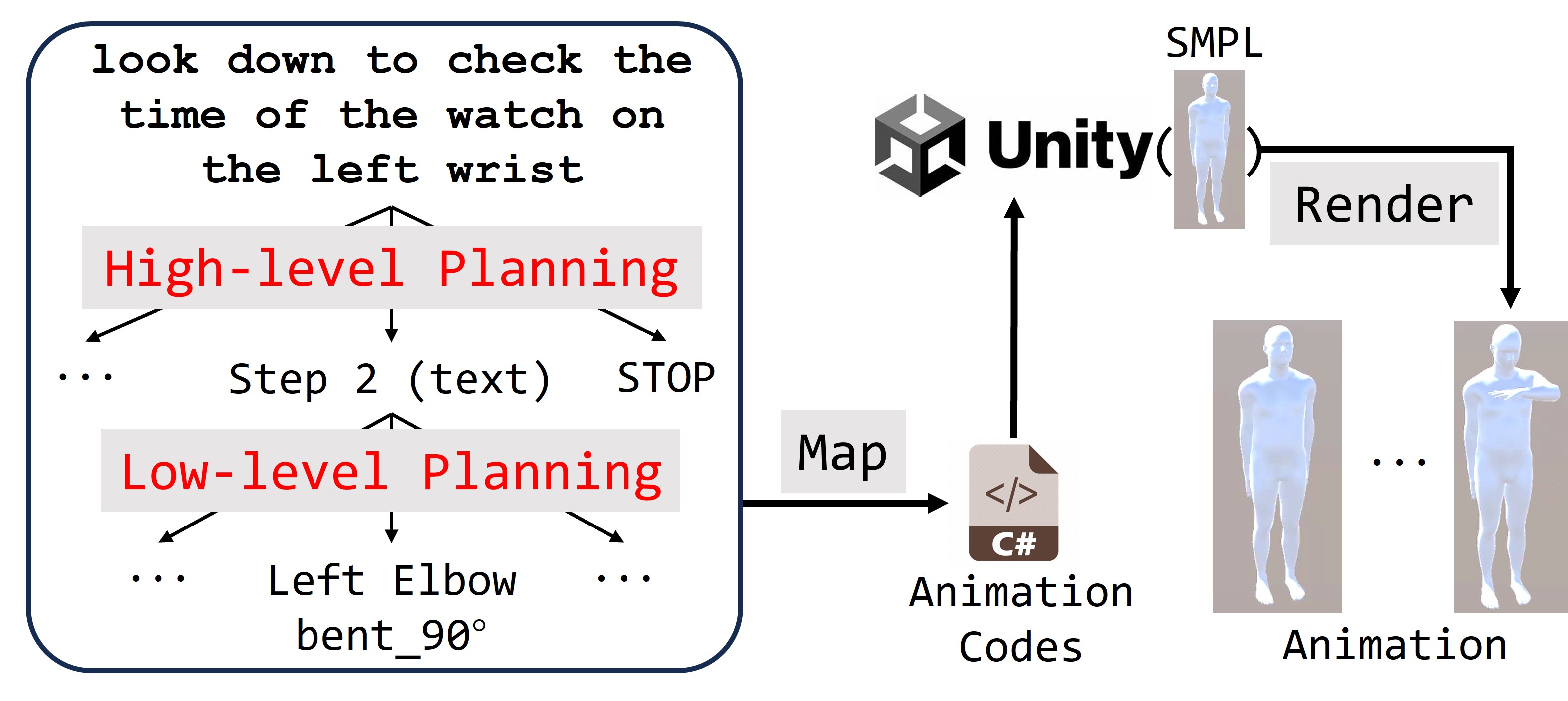}
    \caption{Our motion knowledge grounding pipeline converts the instruction into a high-level movement plan, followed by body part position predictions, which are mapped to animation codes and rendered in Unity.}
    \label{fig:intro}
\end{figure}

In this paper, we explore LLMs' knowledge of human motion through their capabilities to drive a 3D human avatar. Following the natural hierarchy from action sequences to body part movements \citep{motion_hierarchy}, our approach (Figure \ref{fig:intro}) consists of two stages: (1) \textbf{High-level Planning}, where an LLM generates step-by-step body part movements in natural language from the input motion instruction, and (2) \textbf{Low-level Planning}, where the LLM selects a position for every body part from a predefined set of poses within each step (e.g., \texttt{neutral}, \texttt{bent\_in\_90\_degrees} for left elbow). These predictions are then converted into animation codes for the Skinned Multi-Person Linear (SMPL; \citealp{SMPL}) 3D human model in Unity,~\footnote{\href{https://unity.com/}{https://unity.com/}} using predefined rules that map body part positions to SMPL joint rotations (e.g., bending left elbow to 90 degrees maps to \texttt{m\_avg\_L\_Elbow} rotation from \texttt{(0, 0, 0)} to \texttt{(0, 90, 0)}). The rendered animations from linearly interpolated LLM-selected poses provide a clear verification lens for human evaluators.

We carefully design 20 representative motion instructions with full coverage of basic movement primitives and balanced body part usage, and evaluate both commercial (e.g., Claude 3.5 Sonnet) and open-source (e.g., Llama-3.1-70B) LLMs through three complementary approaches: (1) human and GPT-4.1-based evaluation of high-level planning feasibility and key movement completeness, assessing the conceptual understanding of motion; (2) quantitative comparison of low-level body part positioning against oracle annotations for scalability and reproducibility; and (3) human and Gemini 2.5 Pro-based judgement of animation quality for both complete motion and individual body parts, capturing multiple valid motion variations with holistic feedback on naturalness.

Firstly, we find that LLMs demonstrate high competence in generating high-level plans with physically proper key movements. However, they struggle with precise body part positioning, especially for multi-step motions involving high-degree-of-freedom body parts (e.g., upper arm). These positioning errors often accumulate across multiple body parts, resulting in low-quality animations.

Secondly, breaking down motion queries into atomic components enhances performance compared to single-round generation. For high-level planning, iterative querying of individual motion aspects (e.g., body movements, states) proves more effective. For low-level planning, hierarchical decomposition and position-by-position selection consistently yield superior results.

Finally, while LLMs provide reasonable approximations for general spatial descriptions (e.g., the bending motion for wiping a one-meter high table), they fail to handle precise spatial specifications in text (e.g., pick up the object by foot), and fall short in generating accurate spatial and temporal parameters for avatar control. However, LLMs show promise in conceptualizing creative motions (e.g., strut like a peacock showing off feathers) and distinguishing culturally-specific motion patterns (e.g., differentiating normal kneel to bow and kneel to perform a Japanese bow), suggesting their potential to provide enhanced semantic understanding when combined with high-quality low-level motion generators from traditional supervised approaches.
\section{Related Work}

Contemporary generative models show remarkable progress in synthesizing realistic human body movements from natural language instructions \citep{Guo2022text2motion, MDM, motiondiffuse, momask}. However, these models often fail on novel motion instructions out of the limited training datasets, such as compositional instructions, rare activities, or nuanced movement descriptions. To address this generalization problem, recent work uses LLMs to extract specific motion-relevant information, indicating that LLMs might contain rich human motion knowledge.

\citet{SINC} use LLMs to identify relevant body parts for action labels like ``stroll'', showing LLMs' understanding of the anatomical requirements for different movements. However, they only focus on simple action-to-body-part mapping without exploring complex motion reasoning capabilities. Later research further prompts LLMs to decompose abstract motion descriptions into sequential, step-by-step movement specifications \citep{finemotiondiffuse}. More advanced approaches leverage LLMs for hierarchical motion planning and control. For instance, CoMo \citep{huang_control} and Fg-T2M++ \citep{fg_t2m++} employ LLMs to parse ambiguous instructions into structured descriptions targeting specific body parts, enabling fine-grained control over motion generation. More comprehensively, \citet{atomic-text-motion} propose a framework that converts instructions into atomic motion plans organized by predefined body segments, such as spine, left upper limb, etc. Instead of using LLMs for motion-related text generation, recent work also shows that LLMs can directly generate keyframe coordinates to be interpolated as motions \citep{free-motion, text2animation}.

The aforementioned approaches focus on leveraging LLMs as auxiliary tools to optimize text-to-motion systems. However, they overlook the fundamental question of what motion knowledge LLMs actually possess and how accurately they understand human movement principles. We address this research gap by grounding LLM responses into 3D avatar animations, and probing their motion knowledge across multiple levels of abstraction.
\begin{figure*}[t]
    \centering
    \includegraphics[width=\textwidth]{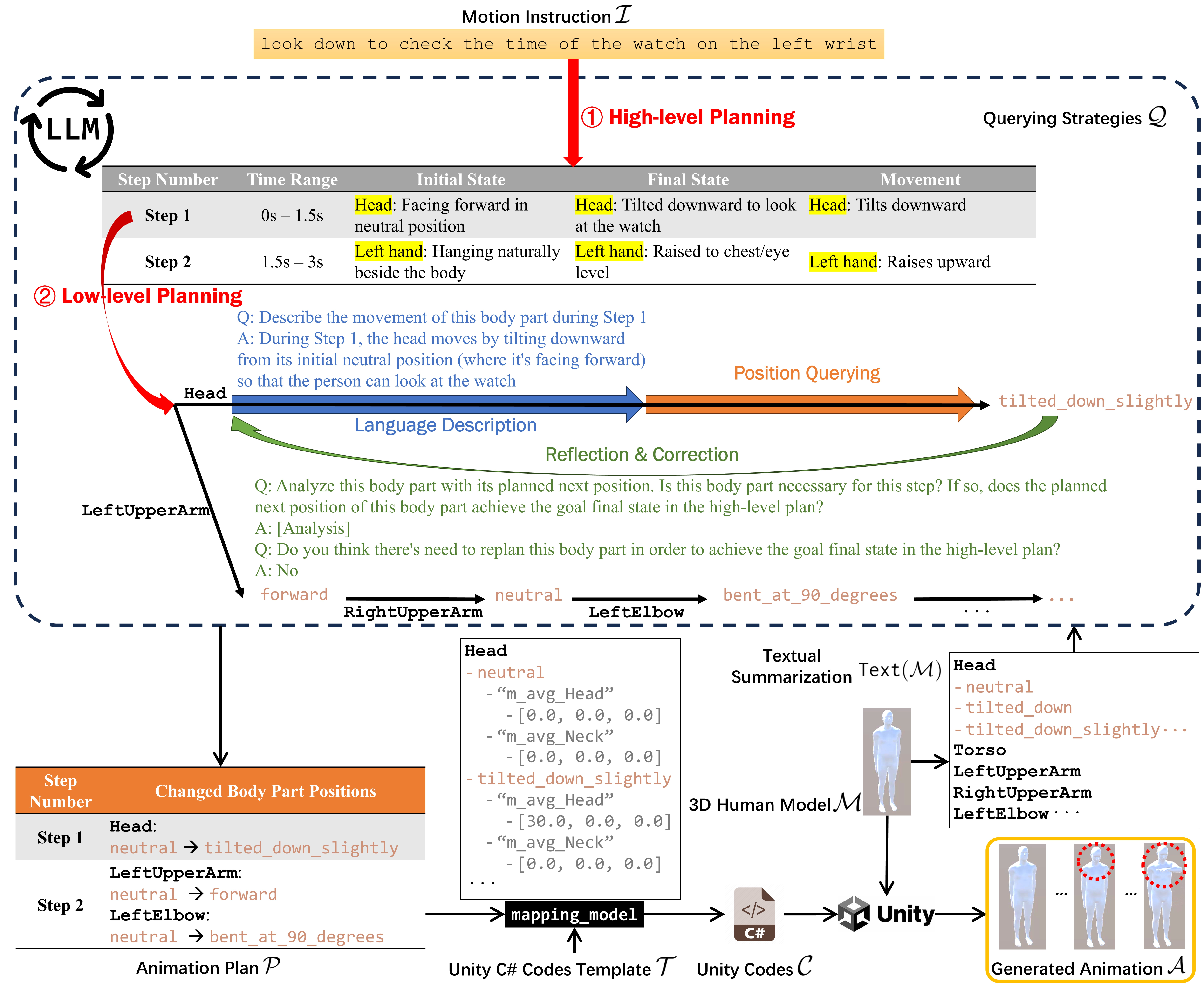}
    \caption{Our motion knowledge grounding pipeline: (1) An LLM processes natural language motion instructions using hierarchical querying strategies to generate an animation plan with specific body part positions, (2) A mapping model converts the animation plan into Unity-compatible codes by transforming body part positions into SMPL joint rotations, (3) Unity renders the final animation by executing the codes.}
    \label{fig:pipeline}
\end{figure*}

\section{Methodology}

This section presents our comprehensive methodology for motion knowledge grounding and evaluation. We introduce a hierarchical pipeline that converts natural language instructions into Unity animations (\S \ref{sec:pipeline}). Within the pipeline, we develop systematic querying strategies that guide LLMs through high-level motion decomposition and low-level body part specification (\S \ref{sec:strategies}). Finally, we design a novel evaluation framework to assess LLM capabilities at multiple levels of the pipeline (\S \ref{sec:eval}).

\subsection{Motion Knowledge Grounding Pipeline}
\label{sec:pipeline}

Figure \ref{fig:pipeline} illustrates our motion knowledge grounding pipeline using schematic prompts. Firstly, based on the hierarchical joint structure of the SMPL model (e.g., \texttt{m\_avg\_Head}'s child joint is \texttt{m\_avg\_Neck}), we define a finite set of textually summarized positions for preset body parts. For example, we define the body part \texttt{Head} to incorporate \texttt{m\_avg\_Head} and \texttt{m\_avg\_Neck} with available positions like \texttt{neutral}, \texttt{tilted\_down}, \texttt{tilted\_down\_slightly}, etc. We design hierarchical querying strategies that first decompose the input motion instruction into sequential high-level steps, then for each step, iteratively specify body part positions by selecting from the textually summarized positions. These strategies guide the LLM to generate the animation plan.

Secondly, we employ a mapping model to convert the animation plan into Unity codes by transforming the specified body part positions into joint rotations on SMPL, and incorporating them into a code template.

Finally, we render the animation by executing Unity codes on SMPL, where joint rotations are linearly interpolated between consecutive steps.~\footnote{We provide the full prompting details in Appendix \ref{appendix:prompts}, and the details of preset body parts and their mapping rules to predefined positions in Appendix \ref{appendix:mapping_rules}.}

\subsection{Querying Strategies}
\label{sec:strategies}

\paragraph{High-level Planning.} We employ two strategies to translate the instruction into step-by-step descriptions: (1) \textit{piece\_by\_piece} --- querying sequentially about body movements, states, timing, and completion status per step, and (2) \textit{in\_one\_go} --- generating the complete motion plan in one query. This comparison evaluates the effectiveness of atomic versus holistic planning approaches.

\vspace{-0.15em}
\paragraph{Low-level Planning.} To convert high-level plans into low-level body part positions for SMPL mapping, we first prompt the LLM to generate a language description of the queried body part, then implement three approaches: (1) \textit{hierarchical} --- querying from large to atomic components (e.g., first determining if elbow is straight or bent, then if bent, specifying slightly, 90 degrees, or fully), (2) \textit{one\_by\_one} --- offering positions sequentially (e.g., first whether straight, then whether slightly bent, etc.), and (3) \textit{all} --- presenting all positions simultaneously. These approaches balance structural guidance against decision complexity. We query related body parts sequentially (e.g., \texttt{LeftUpperArm}, \texttt{RightUpperArm}, \texttt{LeftElbow}) for symmetry and employ self-reflection for improved accuracy.

\subsection{Evaluation Framework}
\label{sec:eval}

We propose an evaluation framework aligned with our motion knowledge grounding pipeline, covering high-level planning, low-level planning, and complete animation generation. In the absence of existing metrics for this purpose, we introduce new ones tailored to each level: for high-level plans and animations, human judgement captures nuanced aspects of instruction adherence and naturalness, while multimodal LLMs enable scalable and consistent assessment; for low-level plans, automated metrics provide reproducible evaluation of body part positioning against annotated ground truth. Based on this framework, we design representative instructions to assess LLMs' motion understanding in diverse scenarios.

\paragraph{High-level Planning.} To evaluate LLMs' abilities to conceptually decompose motion instructions, we introduce \textbf{High-level Plan Score (HPS)}, a human-scored 5-point Likert scale metric assessing the physical feasibility and completeness of generated movement plans. Each high-level plan is scored by three independent annotators following rubrics.~\footnote{Evaluation details are provided in Appendix \ref{appendix:human_evaluation_details_high_level_planning}.} Additionally, GPT-4.1 is utilized as the automated evaluator, employing the identical evaluation guidelines provided to the human evaluators, with three evaluations per high-level plan.

\paragraph{Low-level Planning.} To consistently evaluate LLMs' capabilities to specify body part positions, we use fixed high-level plans from GPT-4o with the \textit{in\_one\_go} strategy followed by manual correction. LLMs predict body part positions based on these plans. For scalable and reproducible evaluation, we annotate oracle positions of all body parts across steps, and calculate \textbf{Body Part Position Accuracy (BPPA)} by comparing LLM-generated positions against the annotated ones.

\paragraph{Complete Animation Generation.} We conduct human evaluation of the complete pipeline's animations to better accommodate valid motion variations and assess overall naturalness. Five independent annotators rate each animation using: (1) \textbf{Whole Body Score (WBS)} --- a 5-point Likert scale measuring adherence to the motion instruction, and (2) \textbf{Body Part Quality (BPQ)} --- classification of six key body parts (\texttt{Head}, \texttt{Torso}, \texttt{Left Arm}, \texttt{Right Arm}, \texttt{Left Leg}, \texttt{Right Leg}) into ``Good'', ``Partially Good'', ``Bad'', or ``Not Relevant'' categories. We introduce ``Not Relevant'' to distinguish between motion-critical body parts (e.g., arms during throwing) and those with little involvement in the action (e.g., legs during a standing wave), while still marking any unnatural movement as ``Bad'', helping evaluators provide targeted feedback on the quality of key motion components. Oracle animations are evaluated separately to establish performance bounds without biasing annotators.~\footnote{Full animation evaluation details are in Appendix \ref{appendix:human_evaluation_details_complete_generation}.} Additionally, we employ Gemini 2.5 Pro as an automated evaluator using the same criteria provided to humans. Each animation is evaluated five times, with frames sampled at 1-second intervals as image inputs.~\footnote{We observe that using full video inputs results in notably low agreement with humans, likely due to the unnatural motion artifacts introduced through linear interpolation.}

\paragraph{Motion Instructions.} Given the extensive human evaluation required, we design a focused set of instructions that maximize coverage while remaining feasible for thorough assessment, following HCI practices that emphasize human-in-the-loop assessment through focused, representative examples \citep{hci_eval_design}. We create 20 motion instructions fully covering basic primitives from which complex motions can be composed, with balanced coverage across body parts (head: 15, torso: 16, arms: 16 each, legs: 13 each).~\footnote{The details of these motion instructions are provided in Appendix \ref{appendix:motion_instructions}.}
\section{Results and Analysis}

Using the designed motion instructions, we run the motion knowledge grounding pipeline on both commercial and open-source LLMs including Claude 3.5 Sonnet, GPT-4o, GPT-4o-mini, GPT-3.5-turbo and Llama-3.1-70B.~\footnote{Llama-3.1-8B struggles to follow the output schema, and is easily distracted by body part positions in the chat history. LLM hyperparameters and costs are in Appendix \ref{appendix:model_running_details}.} Nine evaluators with artificial intelligence research backgrounds participate in the human evaluation, where we calculate the inter-annotator agreement. For HPS and WBS using Likert scales, we calculate the pairwise weighted kappa \citep{weighted_kappa}. For category-based BPQ, we apply average pairwise agreement, calculated as the mean percentage of matching categories between evaluator pairs.

This section presents our evaluation results and analysis across three key components of the motion knowledge grounding process: high-level planning that decomposes motion descriptions into sequential steps (\S \ref{sec:results_high_level_planning}), low-level planning that predicts precise body part positions (\S \ref{sec:results_low_level_planning}), and complete animation generation that synthesizes animations from the instructions (\S \ref{sec:results_complete_animation_generation}).

\subsection{High-level Planning}
\label{sec:results_high_level_planning}

As shown in Table \ref{tab:results_high_level_planning}, while GPT-4.1 tends to give lower scores, the \textit{piece\_by\_piece} approach consistently outperforms \textit{in\_one\_go} across LLMs in both human and GPT-4.1 evaluations. Under \textit{piece\_by\_piece}, Claude 3.5 Sonnet and GPT-4o variants achieve similarly high HPS, while Llama-3.1-70B exceeds GPT-3.5-turbo by a large margin of 0.57. These results suggest that while most LLMs possess sophisticated understanding of high-level body movements, this knowledge is more effectively accessed through incremental guidance rather than one-round generation. To assess the consistency between humans and GPT-4.1 judgements, we compute the average HPS per high-level plan and report a Pearson correlation coefficient of 0.665 ($p = 2.47 \times 10^{-24}$), a Spearman correlation coefficient of 0.549 ($p = 1.49 \times 10^{-15}$), and a Krippendorff’s alpha of 0.653. For the inter-annotator agreement, we report an average kappa of 0.74, indicating substantial agreement based on the interpretation of \citet{kappa_interpretation}.

\begin{table}[h]
  \centering
  \small
    \begin{tabular}{ccc}
      \toprule
      \textbf{LLM} & \makecell{\textbf{HPS}\\(\textit{piece\_by\_piece})} & \makecell{\textbf{HPS}\\(\textit{in\_one\_go})} \\
      \midrule
      Claude 3.5 Sonnet & 4.57 / \textbf{4.55} & 4.42 / \textbf{4.53} \\
      GPT-4o & \textbf{4.68} / 4.53 & \textbf{4.55} / 4.28 \\
      GPT-4o-mini & 4.67 / 4.28 & 3.93 / 3.73 \\
      GPT-3.5-turbo & 3.50 / 3.35 & 3.33 / 3.13 \\
      Llama-3.1-70B & 4.07 / 3.92 & - \\
      \bottomrule
    \end{tabular}%
  \caption{HPS for each tested LLM across two high-level planning strategies. Each score pair represents the mean HPS rated by human annotators (left) and by GPT-4.1 (right). Llama-3.1-70B is excluded from \textit{in\_one\_go} due to output schema compliance issues. In addition, we report motion-wise mean scores with standard deviation and variance in Appendix \ref{appendix:statistical_measures_hps}.}
  \label{tab:results_high_level_planning}
\end{table}

\subsection{Low-level Planning}
\label{sec:results_low_level_planning}

As shown in Table \ref{tab:results_low_level_planning}, Claude 3.5 Sonnet and GPT-4o maintain top across three low-level planning strategies. Llama-3.1-70B falls far behind all closed-source LLMs, with large gaps of 9\%--15\% from GPT-3.5-turbo. Exceptionally, GPT-3.5-turbo receives a low BPPA of 21.70\% when running the \textit{all} strategy, possibly due to its limitation in long context understanding.

\begin{table}[t]
  \centering
  \resizebox{\columnwidth}{!}{
    \begin{tabular}{cccc}
      \toprule
      \textbf{LLM} &
      \makecell{\textbf{BPPA (\%)}\\(\textit{hierarchical})} &
      \makecell{\textbf{BPPA (\%)}\\(\textit{one\_by\_one})} &
      \makecell{\textbf{BPPA (\%)}\\(\textit{all})} \\
      \midrule
      Claude 3.5 Sonnet & \textbf{73.52} & \textcolor{blue}{71.23} & \textbf{70.75} \\
      GPT-4o & \textcolor{blue}{70.87} & \textbf{71.70} & \textcolor{blue}{67.49} \\
      GPT-4o-mini & 68.10 & 67.80 & 65.32 \\
      GPT-3.5-turbo & 67.19 & 62.76 & 21.70 \\
      Llama-3.1-70B & 52.60 & 53.34 & 45.87 \\
      \bottomrule
    \end{tabular}
  }
  \caption{BPPA across three low-level planning strategies for each LLM. Each value is averaged from two runs. Bold and blue values respectively indicate the highest and second-highest accuracy for the same strategy. In addition, we report motion-wise mean scores with standard deviation and variance in Appendix \ref{appendix:statistical_measures_bppa}.}
  \label{tab:results_low_level_planning}
\end{table}

\begin{figure}[t]
    \centering
    \includegraphics[width=0.85\linewidth]{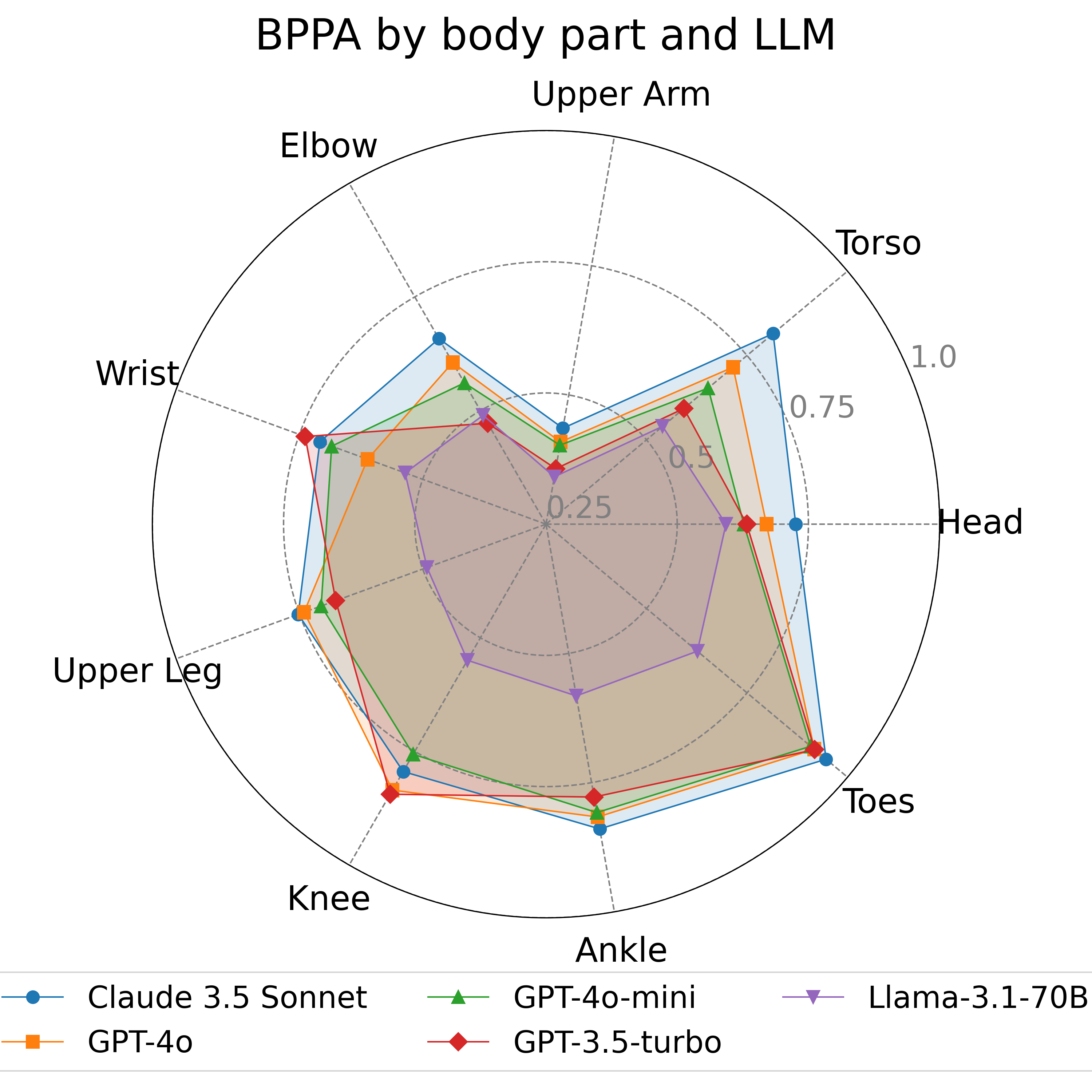}
        \caption{Body-part-wise BPPA for LLMs with the \textit{hierarchical} strategy. BPPA is averaged for paired body parts, e.g., ``Elbow'' for ``LeftElbow'' and ``RightElbow''.}
        \label{fig:BPPA_by_body_part_and_llm}
\end{figure}

\begin{figure}[t]
    \centering
    \includegraphics[width=0.85\linewidth]{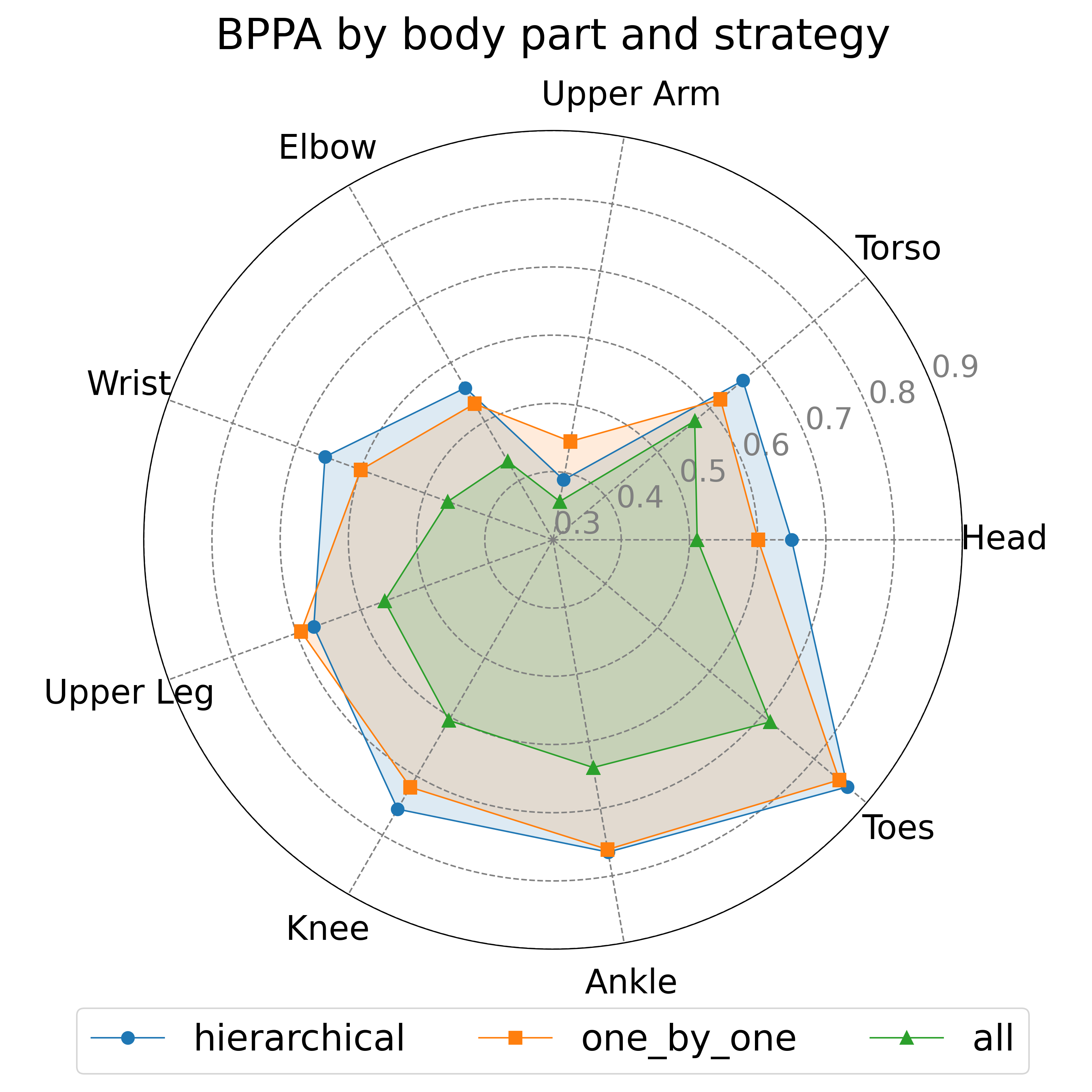}
    \caption{Body-part-wise BPPA for strategies averaged across LLMs. BPPA is averaged for paired body parts.}
    \label{fig:BPPA_by_body_part_and_strategy}
\end{figure}

To examine what drives BPPA differences across LLMs, we visualize body-part-wise BPPA for LLMs using the \textit{hierarchical} strategy in Figure \ref{fig:BPPA_by_body_part_and_llm}. The relative performance of LLMs remains consistent with the overall BPPA shown in Table \ref{tab:results_low_level_planning} for most body parts, except that GPT-3.5-turbo shows notably higher BPPA for \texttt{Knee} and \texttt{Wrist}. \texttt{Upper Arm} shows notably low BPPA, possibly due to its high degree-of-freedom. We discuss the correlation between body part complexity and BPPA in \S \ref{sec:discussion_low_level_planning}.~\footnote{Additionally, higher-ranked LLMs show higher accuracy in identifying and correcting inaccurate body part positions during self-reflection (Appendix \ref{appendix:reflection}).}

For the same LLM, the performance ranking of strategies from highest to lowest (Table \ref{tab:results_low_level_planning}) is mostly \textit{hierarchical}, \textit{one\_by\_one}, and \textit{all}, which indicates that increased structural guidance in prompting enables LLMs to generate more precise body part positions, suggesting enhanced utilization of their inherent understanding of human motion details. Furthermore, we visualize the body-part-wise BPPA for strategies in Figure \ref{fig:BPPA_by_body_part_and_strategy}. The \textit{all} strategy falls far behind across all body parts. The \textit{hierarchical} strategy outperforms \textit{one\_by\_one} in most body parts, except for high-degree-of-freedom body parts \texttt{Upper Arm} and \texttt{Upper Leg}. This indicates that heuristically defined querying structures might be too rigid compared to more flexible position-by-position selection for complex articulations.

\subsection{Complete Animation Generation}
\label{sec:results_complete_animation_generation}

We compare LLMs' complete generation performance under \textit{piece\_by\_piece} high-level planning and \textit{hierarchical} low-level planning. As shown in Table \ref{tab:results_complete_generation_0}, LLMs mostly maintain their relative rankings from BPPA (Table \ref{tab:results_low_level_planning}) when evaluated with WBS, except for GPT-3.5-turbo and Llama-3.1-70B's rankings are reversed in Gemini 2.5 Pro evaluation. While GPT-3.5-turbo largely outperforms Llama-3.1-70B in BPPA, both achieve similarly low WBS, likely because evaluators assign similarly low scores when animation quality falls below a certain threshold. The best performing LLM, Claude 3.5 Sonnet, scores well below the oracle animations, indicating considerable room for motion understanding improvement.

To quantify alignment between human and model evaluations for WBS scoring, we report a Pearson correlation coefficient of 0.585 ($p = 2.24 \times 10^{-12}$), a Spearman correlation coefficient of 0.597 ($p = 6.15 \times 10^{-13}$), and a Krippendorff’s alpha of 0.578. We also observe that Gemini 2.5 Pro tends to produce more conservative and averaged scores compared to human evaluators --- it assigns notably lower scores to oracle animations with a difference of 0.6, while assigning comparatively higher scores to LLM-generated motions. For the inter-annotator agreement, WBS achieves a moderate kappa of 0.531.~\footnote{Most human agreement pairs fall in the moderate to substantial agreement range (0.41--0.80), as shown in Figure \ref{fig:agreement_overall_scores}.} We hypothesize this moderate agreement stems from the inherent uncertainty in human motion, where people can move and express themselves in various valid ways.

\begin{figure}[t]
    \centering
    \small
    \begin{tabular}{cc}
      \toprule
      \textbf{LLM} & \textbf{WBS} \\
      \midrule
      Claude 3.5 Sonnet & \textbf{3.29} / \textbf{3.65} \\
      GPT-4o & 3.13 / 3.22 \\
      GPT-4o-mini & 2.87 / 2.73 \\
      GPT-3.5-turbo & 2.14 / 2.20 \\
      Llama-3.1-70B & 2.13 / 2.29 \\
      \midrule
      (Oracle Annotation) & 4.57 / 3.97 \\
      \bottomrule
    \end{tabular}
    \captionof{table}{WBS with \textit{piece\_by\_piece} high-level planning and \textit{hierarchical} low-level planning. Each score pair shows human-rated (left) and Gemini 2.5 Pro-rated (right) WBS. Motion-wise statistics are reported in Appendix \ref{appendix:statistical_measures_wbs}.}
    \label{tab:results_complete_generation_0}
\end{figure}

\begin{table*}[t]
\centering
\small
\resizebox{\textwidth}{!}{%
\setlength{\tabcolsep}{2.5pt}
\begin{tabular*}{\textwidth}{@{\extracolsep{\fill}}c|ccc|ccc|ccc|ccc|ccc|ccc@{}}
\toprule
\multirow{2}{*}{\textbf{LLM}} & \multicolumn{3}{c|}{\texttt{\normalsize{Head}}} & \multicolumn{3}{c|}{\texttt{\normalsize{Torso}}} & \multicolumn{3}{c|}{\texttt{\normalsize{Left Arm}}} & \multicolumn{3}{c|}{\texttt{\normalsize{Right Arm}}} & \multicolumn{3}{c|}{\texttt{\normalsize{Left Leg}}} & \multicolumn{3}{c}{\texttt{\normalsize{Right Leg}
}} \\
& \textbf{G} & \textbf{PG} & \textbf{B} & \textbf{G} & \textbf{PG} & \textbf{B} & \textbf{G} & \textbf{PG} & \textbf{B} & \textbf{G} & \textbf{PG} & \textbf{B} & \textbf{G} & \textbf{PG} & \textbf{B} & \textbf{G} & \textbf{PG} & \textbf{B} \\
\midrule
Claude 3.5 Sonnet & 74.1 & 22.2 & 3.7 & \sethlcolor{pink}\hl{72.6} & 17.7 & 9.7 & \sethlcolor{pink}\hl{25.0} & 53.9 & 21.1 & \sethlcolor{pink}\hl{29.3} & 53.3 & 17.3 & 38.6 & 31.8 & 29.5 & \sethlcolor{pink}\hl{31.7} & 29.3 & 39.0 \\
GPT-4o & 63.8 & 19.1 & 17.0 & 60.7 & 25.0 & 14.3 & 15.2 & \hl{58.2} & 26.6 & 16.9 & \hl{64.9} & 18.2 & \sethlcolor{pink}\hl{46.8} & \hl{36.2} & 17.0 & 29.5 & \hl{47.7} & 22.7 \\
GPT-4o-mini & \sethlcolor{pink}\hl{80.7} & 8.8 & 10.5 & 59.4 & 28.1 & 12.5 & 12.8 & 47.4 & 39.7 & 12.2 & 52.7 & 35.1 & 17.9 & 33.3 & 48.7 & 11.1 & 33.3 & 55.6 \\
GPT-3.5-turbo & 34.2 & 13.2 & \sethlcolor{lightgray}\hl{52.6} & 29.1 & 16.4 & \sethlcolor{lightgray}\hl{54.5} & 3.8 & 41.8 & \sethlcolor{lightgray}\hl{54.4} & 3.8 & 46.2 & 50.0 & 10.3 & 30.8 & 59.0 & 5.4 & 18.9 & 75.7 \\
Llama-3.1-70B & 44.0 & \hl{32.0} & 24.0 & 34.8 & \hl{34.8} & 30.4 & 6.9 & 41.4 & 51.7 & 9.4 & 38.8 & \sethlcolor{lightgray}\hl{51.8} & 15.5 & 7.0 & \sethlcolor{lightgray}\hl{77.5} & 5.9 & 5.9 & \sethlcolor{lightgray}\hl{88.2} \\
\midrule
(Average) & 59.4 & 19.0 & 21.6 & 51.3 & 24.4 & 24.3 & 12.8 & 48.6 & 38.7 & 14.3 & 51.2 & 34.5 & 25.8 & 27.8 & 46.3 & 16.7 & 27.0 & 56.2 \\
(Oracle) & 89.6 & 10.4 & 0.0 & 80.3 & 18.2 & 1.5 & 74.0 & 19.5 & 6.5 & 76.3 & 19.7 & 4.0 & 76.6 & 14.9 & 8.5 & 76.1 & 13.0 & 10.9 \\
\bottomrule
\end{tabular*}
}
\caption{Percentage (\textbf{\%}) of Body Part Quality (BPQ) from human evaluation after excluding ``Not Relevant'' across evaluated LLMs. \textbf{G}, \textbf{PG}, and \textbf{B} respectively stand for ``Good'', ``Partially Good'', and ``Bad''. Highest percentages for each category are highlighted in \sethlcolor{pink}\hl{pink} (\textbf{G}), \sethlcolor{yellow}\hl{yellow} (\textbf{PG}), and \sethlcolor{lightgray}\hl{gray} (\textbf{B}).}
\label{tab:model_comparison_body_part_labels}
\end{table*}

A detailed analysis of BPQ reveals distinct performance tendencies across different body parts (Table \ref{tab:model_comparison_body_part_labels}). When comparing the percentages of averaged results among LLMs and oracle animations, head and torso movements demonstrate relatively smaller deficits, while arm and leg motions exhibit notably larger inaccuracies. Among the LLMs, Claude 3.5 Sonnet and GPT-4o consistently achieve higher percentages in the ``Good'' and ``Partially Good'' categories, whereas GPT-3.5-turbo and Llama-3.1-70B show higher frequencies in the ``Bad'' category across all body parts. Furthermore, we find that Gemini 2.5 Pro has limited alignment with nuanced human judgements of body parts.~\footnote{The agreement scores between Gemini 2.5 Pro and human majority votes for body parts remain below 0.8 (Table \ref{tab:agreement_body_parts}), far from the expected near-perfect agreement.}
\section{Discussion}
To reveal both the capabilities and limitations of LLMs in each stage of motion knowledge grounding in more details, we analyze the performance patterns in high-level planning, low-level planning, and complete animation generation through quantitative error analysis and representative case studies.

\subsection{High-level Planning}

To better understand how LLMs perform differently in HPS (Table \ref{tab:results_high_level_planning}), we count the numbers of high-level plans with wrong or incomplete action descriptions generated using the \textit{piece\_by\_piece} strategy (Table \ref{tab:high_level_planning_cnts}). The ranking of combined error counts aligns with the HPS results. Specifically, Llama-3.1-70B and GPT-3.5-turbo show inferior performance due to higher combined error counts, with GPT-3.5-turbo receiving the lowest HPS primarily due to having the most incomplete plans. Two representative cases demonstrate the two error types: For the motion instruction \underline{lift the right shoe} \underline{with both hands and put it on in the air}, the wrong plan only specifies lifting the foot while reaching down to grab the shoe, deviating from the intended sequence of lifting the shoe then putting it on (Figure \ref{fig:high_level_plan_case_wrong}). For \underline{look down to check the time of the} \underline{watch on the left wrist}, the incomplete plan omits the crucial action of positioning the left arm to make the watch visible (Figure \ref{fig:high_level_plan_case_incomplete}).

\begin{table}[h]
  \centering
  \small
    \begin{tabular}{ccc}
      \toprule
      \textbf{LLM} & \textbf{\#Wrong} & \textbf{\#Incomplete} \\
      \midrule
      Claude 3.5 Sonnet & 1 & 5 \\
      GPT-4o & 1 & 4 \\
      GPT-4o-mini & 2 & 3 \\
      GPT-3.5-turbo & 4 & 9 \\
      Llama-3.1-70B & 5 & 5 \\
      \bottomrule
    \end{tabular}
  \caption{Counts of high-level plans with wrong or incomplete action descriptions for each LLM with the \textit{piece\_by\_piece} strategy.}
  \label{tab:high_level_planning_cnts}
\end{table}

\subsection{Low-level Planning}
\label{sec:discussion_low_level_planning}
\begin{figure}[t]
    \centering
    \begin{subfigure}[b]{0.3\textwidth}
        \centering
        \includegraphics[width=0.95\textwidth]{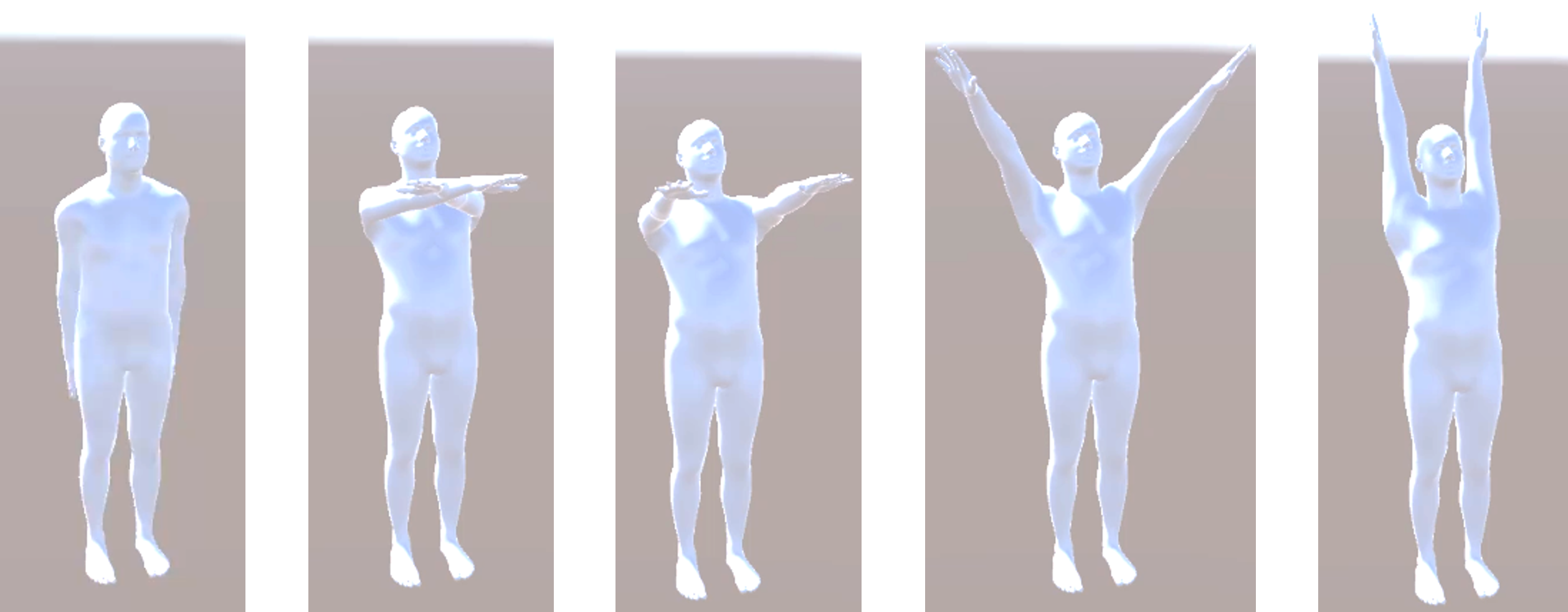}
        \caption{``lean back fully and toss the ball into the air at a 45-degree angle using both hands''}
        \label{fig:low_level_planning_case_0}
    \end{subfigure}
    \hspace{0.1\textwidth}
    \begin{subfigure}[b]{0.3\textwidth}
        \centering
        \includegraphics[width=0.7\textwidth]{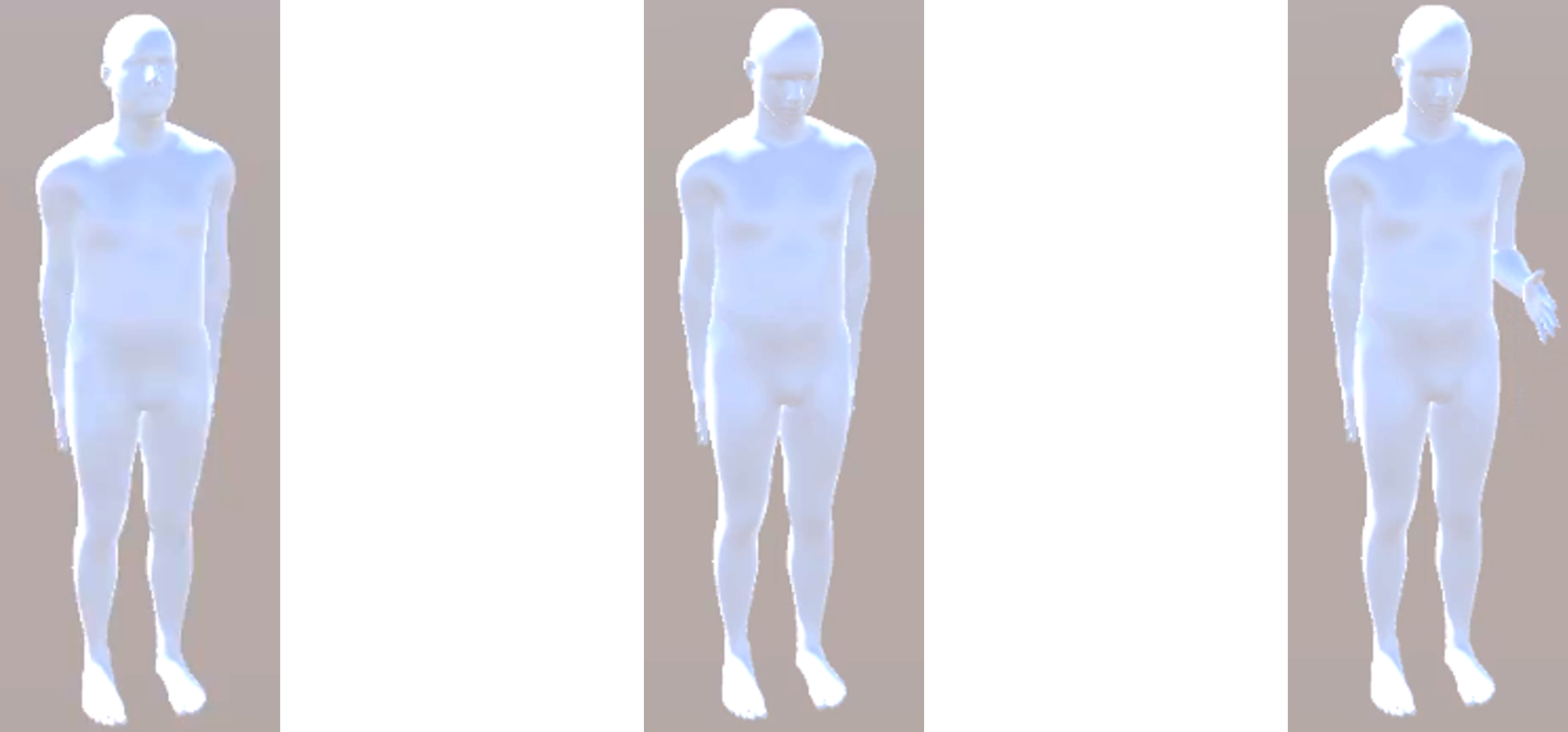}
        \caption{``look down to check the time of the watch on the left wrist''}
        \label{fig:low_level_planning_case_1}
    \end{subfigure}
    \caption{Key frames at one-second intervals from the example animations predicted by Claude 3.5 Sonnet using the \textit{hierarchical} strategy.}
\end{figure}

LLMs typically achieve BPPA between 50\% and 75\% (Table \ref{tab:results_low_level_planning}), revealing their limitations in predicting precise body part positions. When observing animations predicted using the \textit{hierarchical} strategy from the low-level planning evaluation, we find that although BPPA is high, there might be critical errors in animations. The example animation (Figure \ref{fig:low_level_planning_case_0}), despite achieving BPPA of 0.7812, wrongly crosses the arms and fails to toss the ball, showing that positioning errors often accumulate across multiple body parts, resulting in low-quality animations. In another example (Figure \ref{fig:low_level_planning_case_1}), despite achieving high BPPA of 0.9688, the animation fails to rotate the shoulder to align the wrist with the face, demonstrating that errors in key articulated joints can severely impact the overall motion quality regardless of high BPPA.

Furthermore, to clearly understand how different body parts and motions affect BPPA, we analyze the correlation between their complexity and BPPA when using the \textit{hierarchical} strategy.

\paragraph{Body Part Complexity.} We observe two phenomena in the correlation between the number of possible positions and BPPA for different body parts (Figure \ref{fig:low_level_planning_body_part_DoF_accuracy_correlation}). First, BPPA tends to inversely correlate with movement flexibility --- body parts with more possible positions show lower accuracy compared to more constrained parts (e.g., upper arm versus upper leg). Second, LLMs demonstrate higher accuracy for lower body parts compared to their upper body counterparts.

\paragraph{Motion Complexity.} LLM performance in position prediction declines as motions become more complex with increased steps and body parts (Figure \ref{fig:low_level_planning_motion_complexity_accuracy_correlation}). This degradation likely stems from two factors: LLMs' difficulty in maintaining spatial relationships across extended movement sequences, and training data bias where humans describe only core movements (e.g., ``raise arms'') while omitting auxiliary ones (e.g., shoulder and elbow adjustments). Unlike humans who can intuitively infer these auxiliary movements, LLMs appear limited in developing such implicit understanding.

\subsection{Complete Animation Generation}

\begin{figure*}[t]
    \centering
    \begin{minipage}[t]{0.32\textwidth}
        \centering
        \begin{subfigure}[t]{\textwidth}
            \centering
            \includegraphics[height=1.6cm]{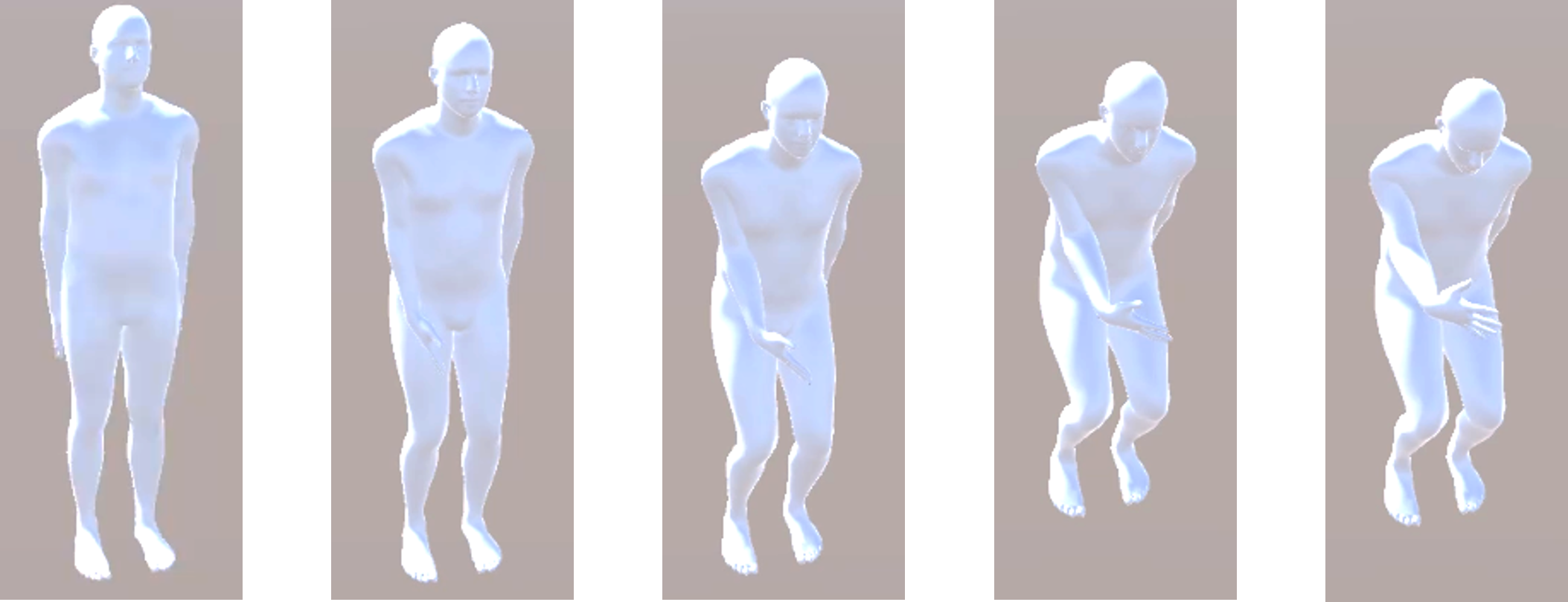}
            \includegraphics[height=1.6cm]{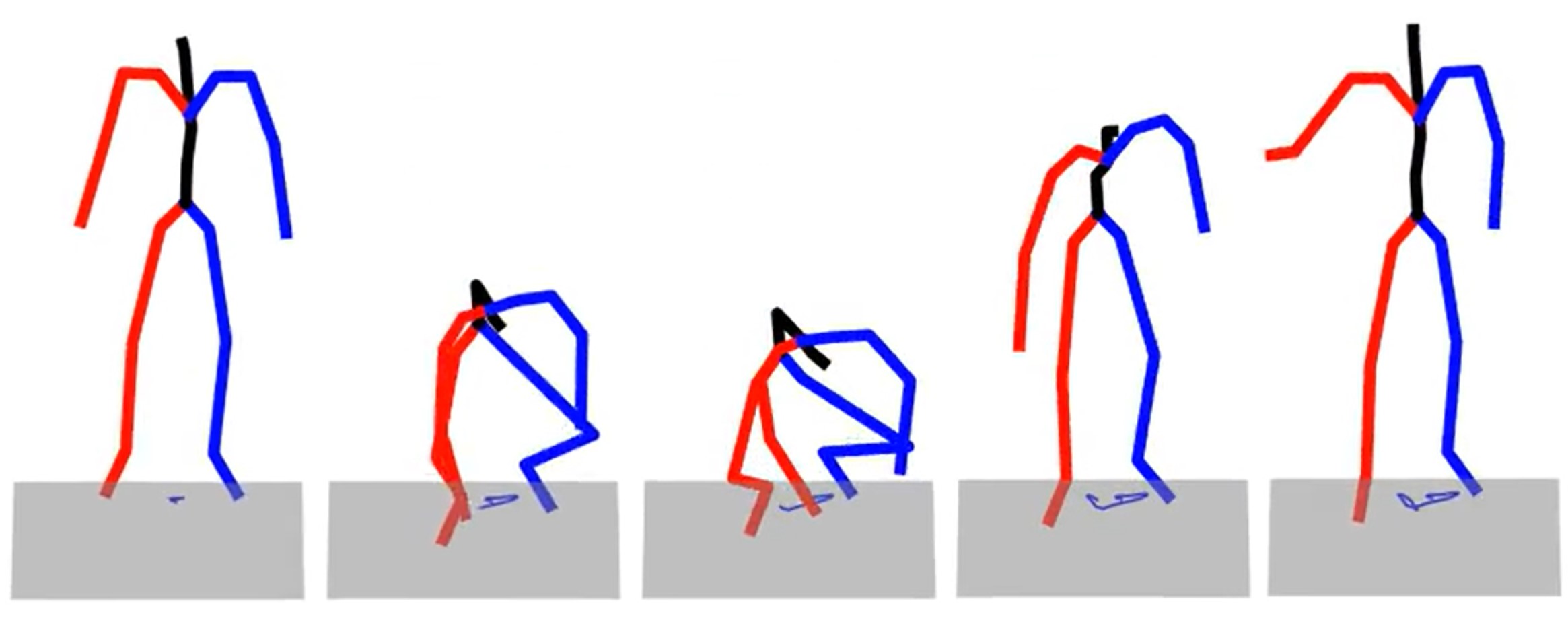}
            \caption{``squat to pick up litter by the right foot with the right hand''}
            \label{fig:complete_generation_precision_case_0}
        \end{subfigure}
        \vfill
        \begin{subfigure}[t]{\textwidth}
            \centering
            \includegraphics[height=1.6cm]{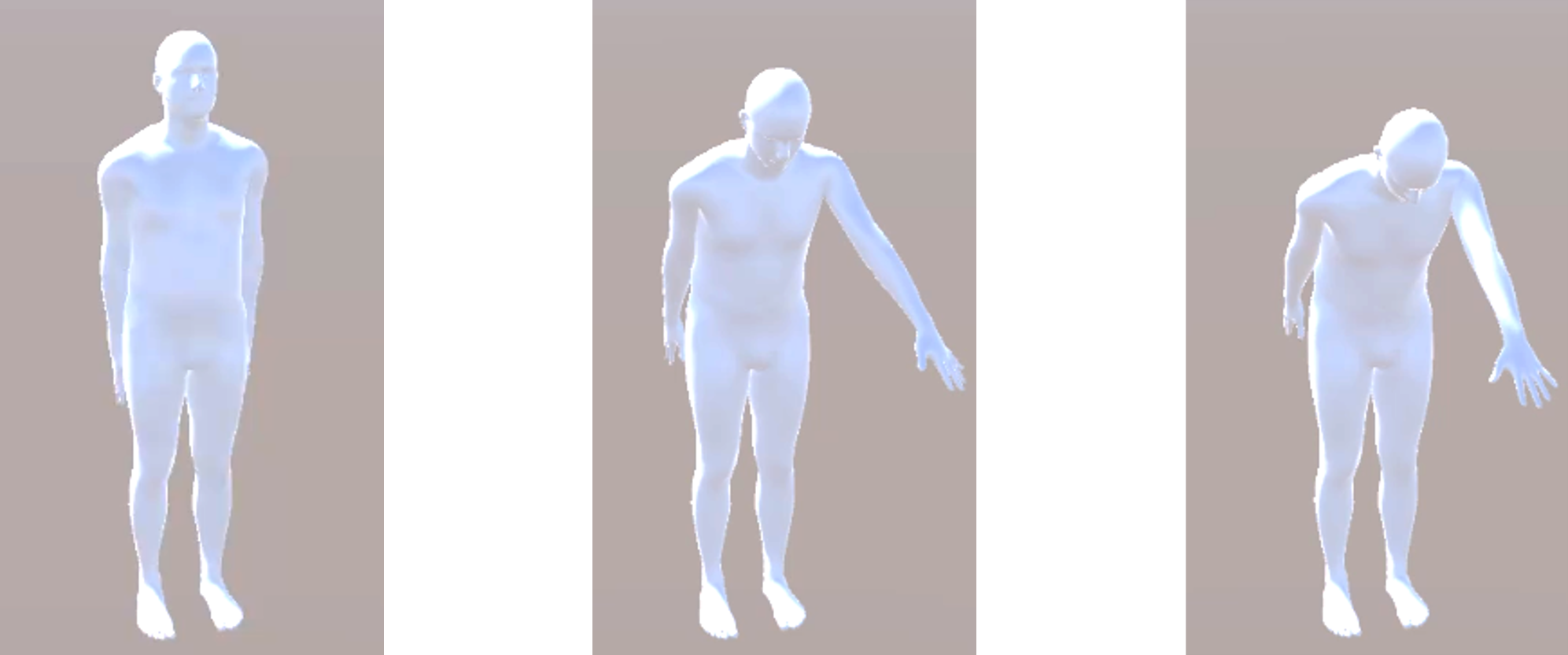}
            \includegraphics[height=1.6cm]{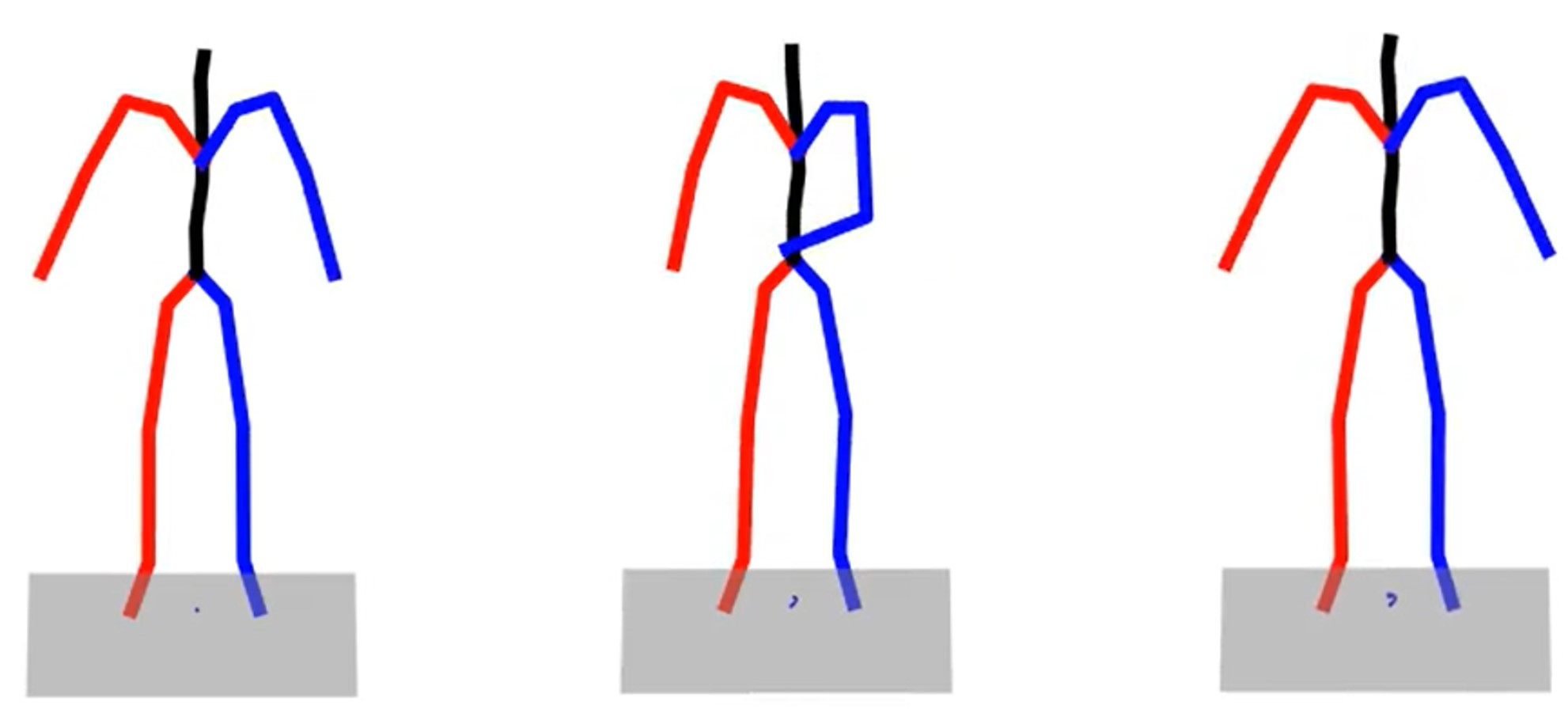}
            \caption{``wipe down the 1-meter-high table in front of you with a cloth in the left hand''}
            \label{fig:complete_generation_precision_case_1}
        \end{subfigure}
    \end{minipage}
    \hfill
    \begin{minipage}[t]{0.3\textwidth}
        \centering
        \begin{subfigure}[t]{\textwidth}
            \centering
            \includegraphics[height=1.6cm]{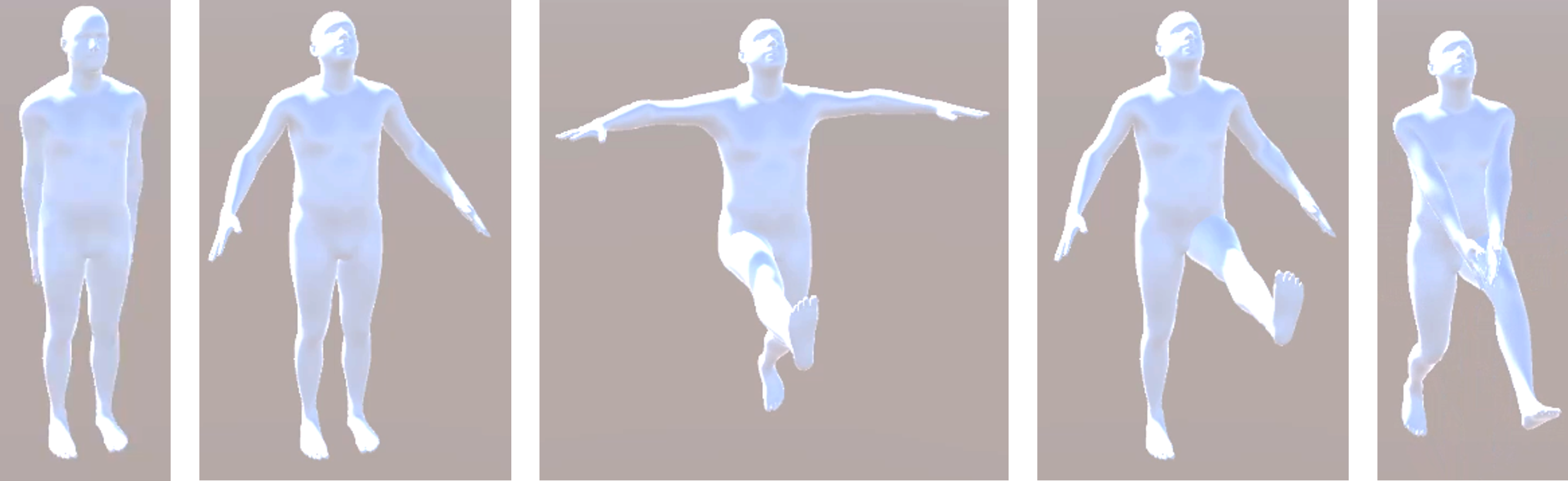}
            \includegraphics[height=1.6cm]{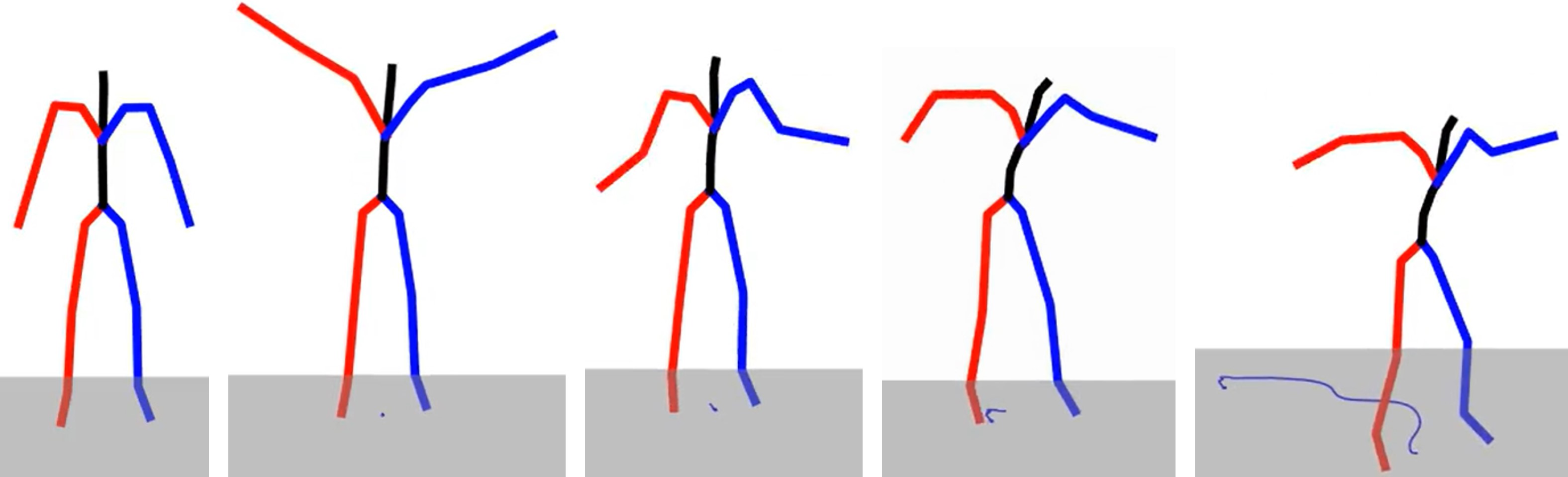}
            \caption{``strut like a peacock showing off its feathers''}
            \label{fig:complete_generation_imagination_case_0}
        \end{subfigure}
        \vfill
        \begin{subfigure}[t]{\textwidth}
            \centering
            \includegraphics[height=1.6cm]{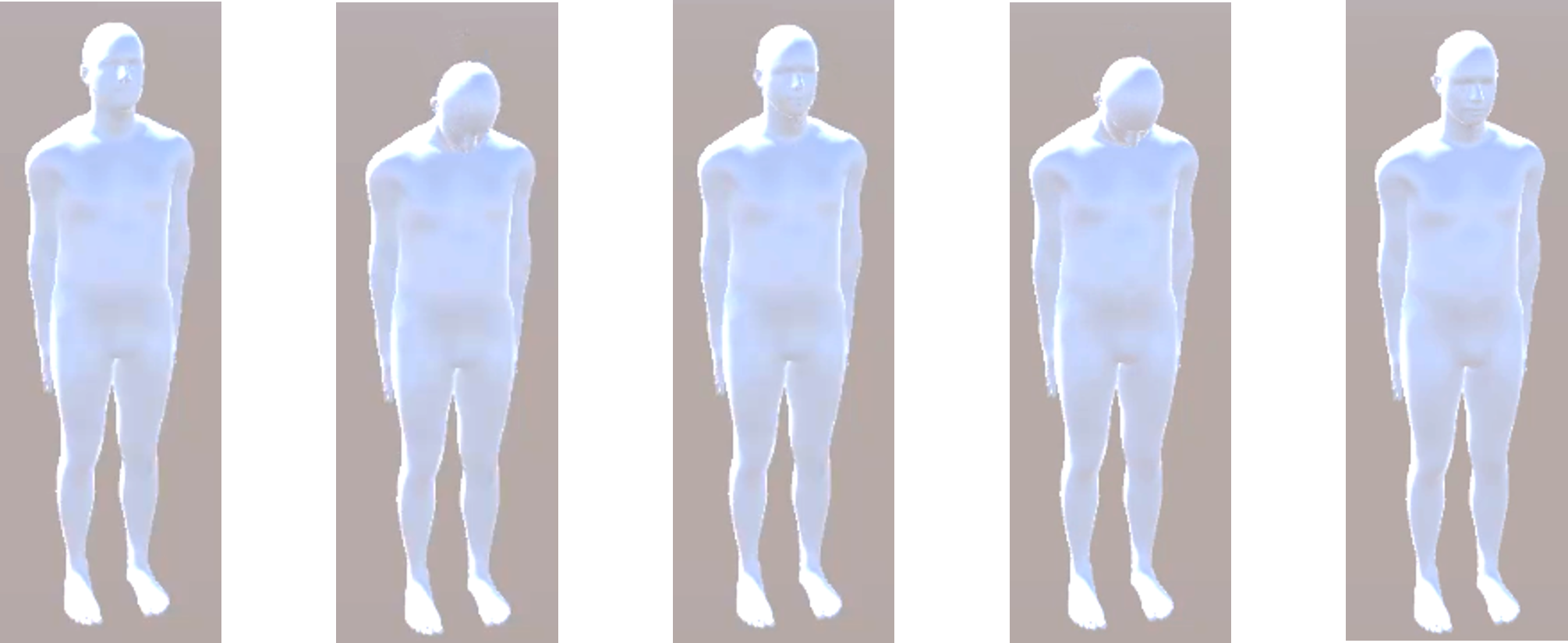}
            \includegraphics[height=1.6cm]{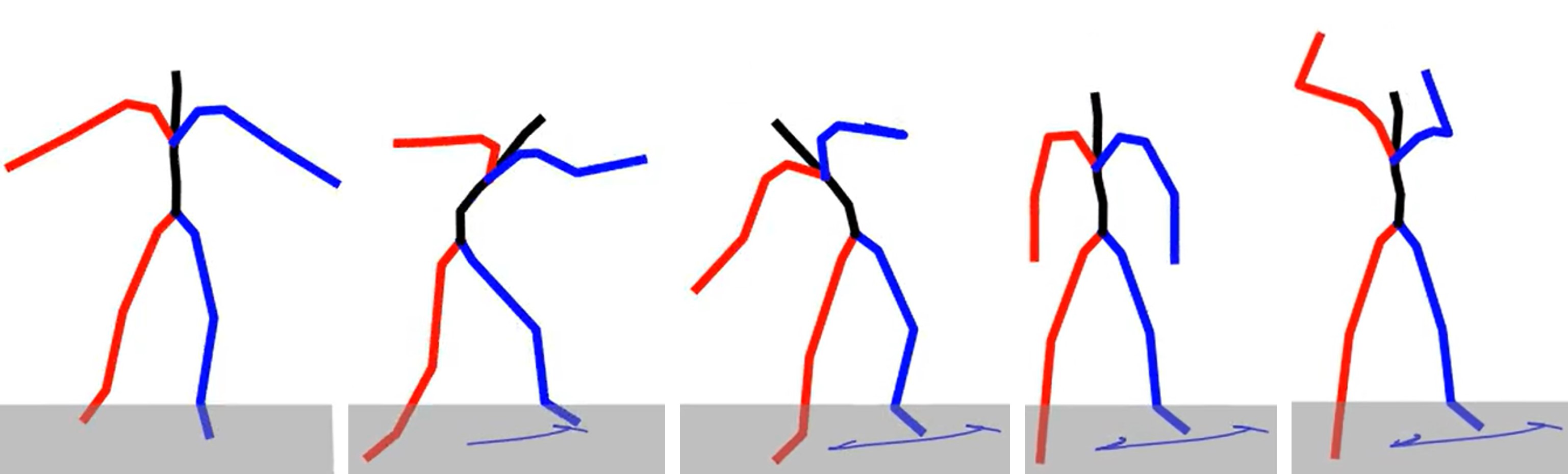}
            \caption{``tap like a woodpecker on a tree''}
            \label{fig:complete_generation_imagination_case_1}
        \end{subfigure}
    \end{minipage}
    \hfill
    \begin{minipage}[t]{0.3\textwidth}
        \centering
        \begin{subfigure}[t]{\textwidth}
            \centering
            \includegraphics[height=1.6cm]{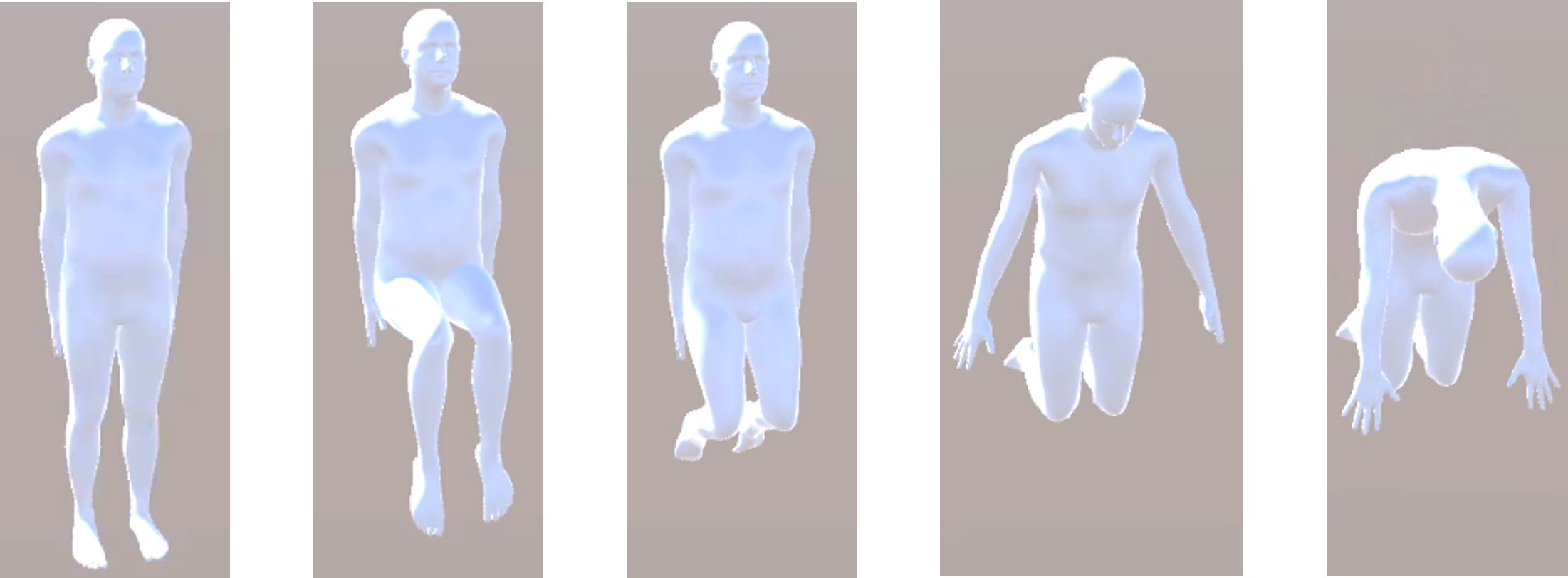}
            \includegraphics[height=1.6cm]{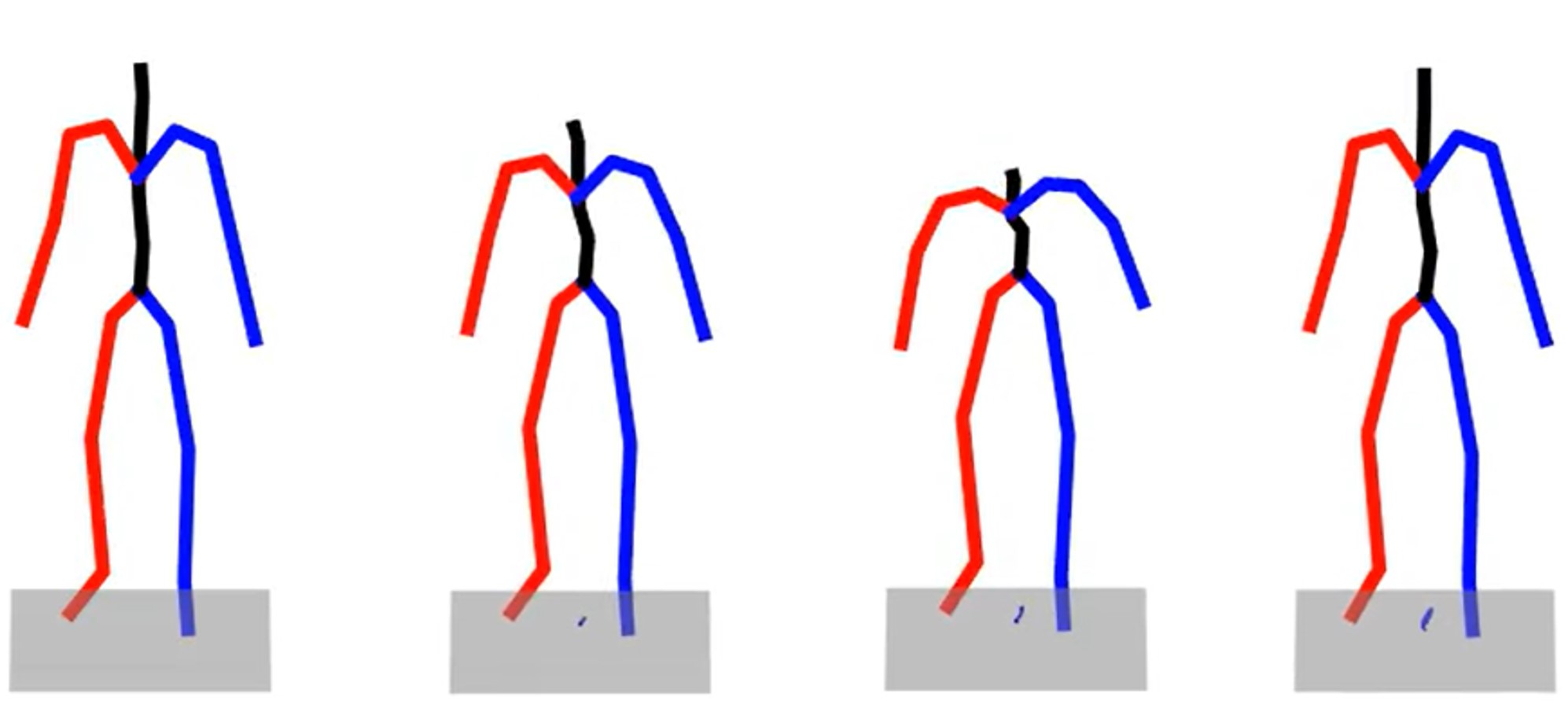}
            \caption{``kneel in a traditional Japanese bow''}
            \label{fig:complete_generation_culture_case_0}
        \end{subfigure}
        \vfill
        \begin{subfigure}[t]{\textwidth}
            \centering
            \includegraphics[height=1.6cm]{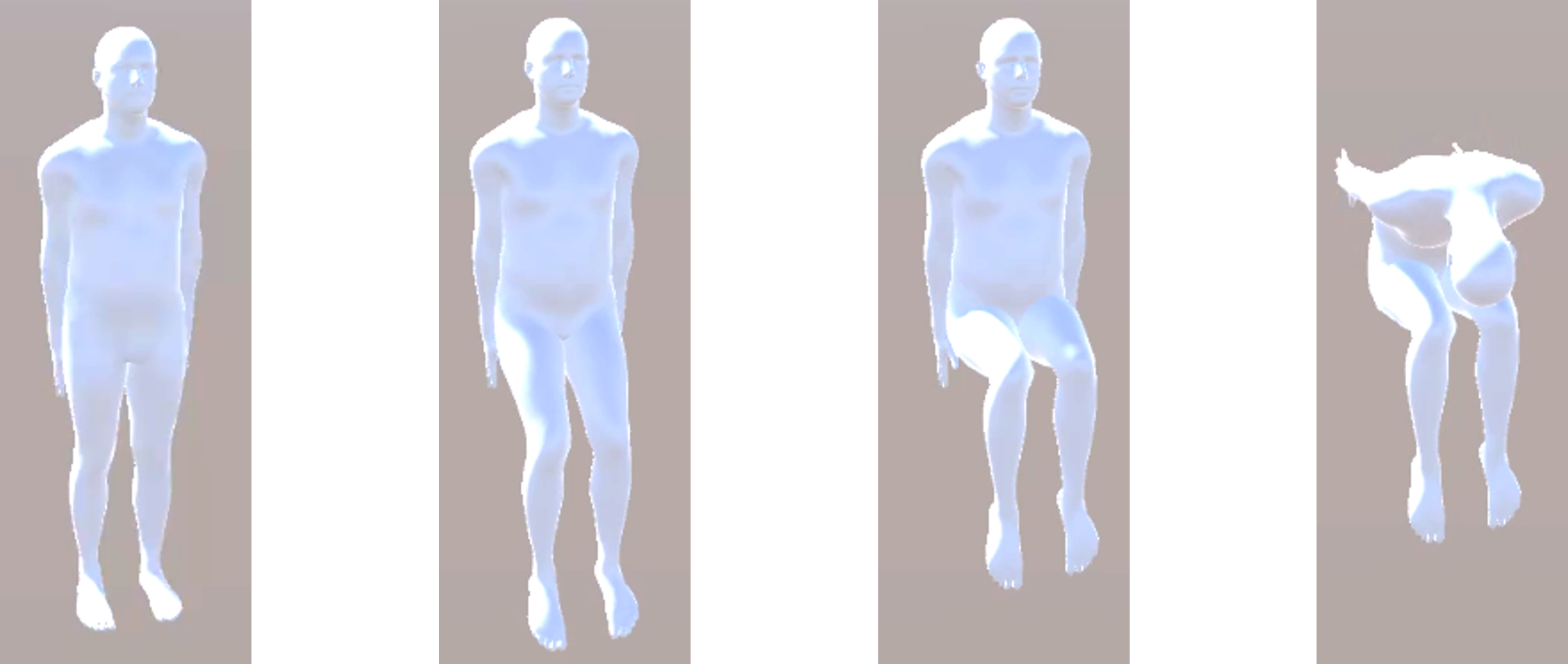}
            \includegraphics[height=1.6cm]{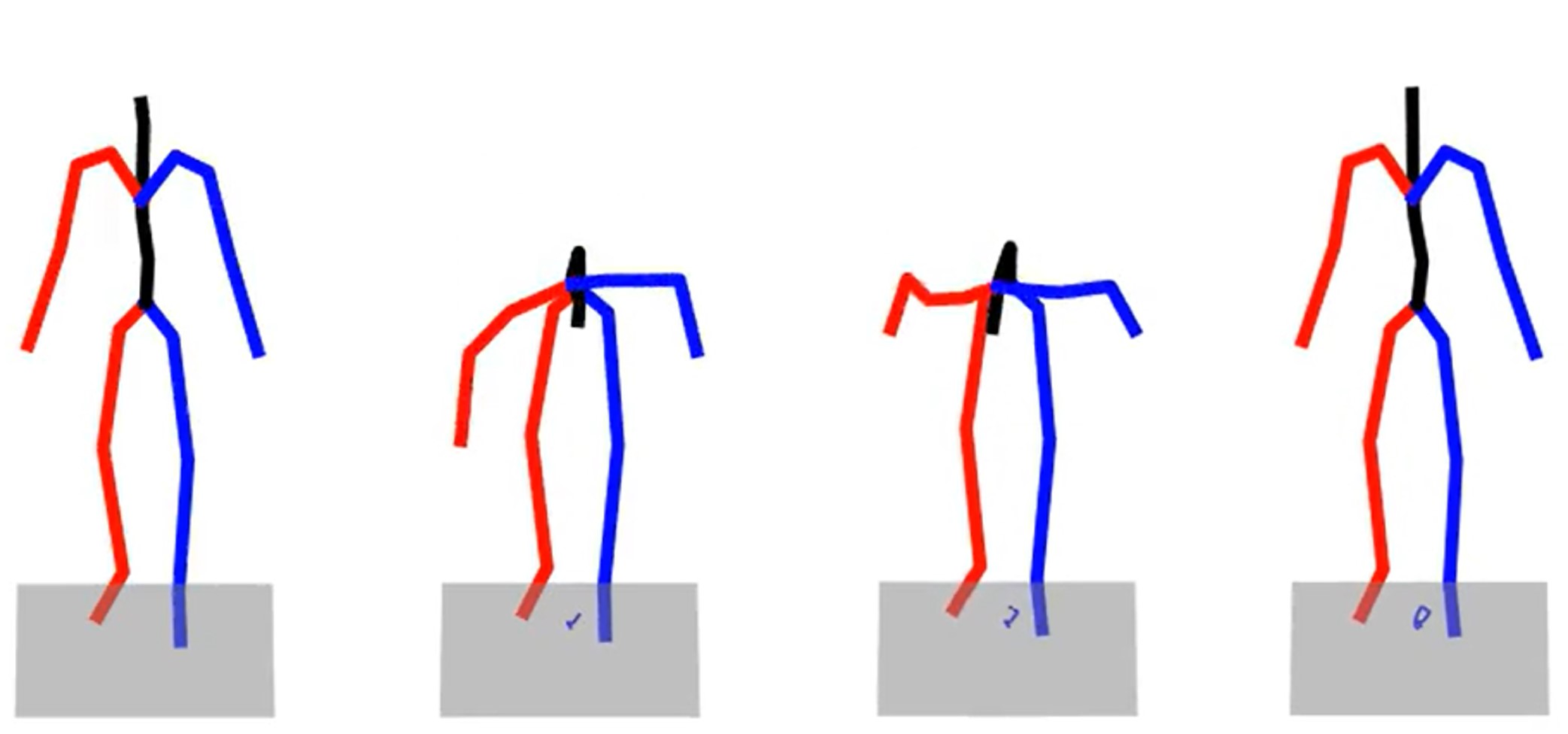}
            \caption{``kneel to bow''}
            \label{fig:complete_generation_culture_case_1}
        \end{subfigure}
    \end{minipage}
    \caption{Key frames for example animations reflecting spatial precision (a, b), imagination by animal imitation (c, d), and cultural awareness (e, f). The frames in human figures are generated using our pipeline, while the frames in stick figures are produced by the text-to-motion generative model MoMask \citep{momask}.}
\end{figure*}

We present a comparative case study between the keyframes extracted from animations generated by our pipeline and those produced by MoMask \citep{momask}, a state-of-the-art text-to-motion generative model. This comparison provides insights into how LLMs' motion understanding differs from specialized motion generative models.~\footnote{Instructions for spatial precision case study are taken from the main experiments, while the others are newly created.}

\paragraph{Spatial Precision.} LLMs show mixed spatial understanding capabilities. For precise positioning requirements such as picking up objects near feet, LLMs consistently fail where MoMask succeeds (Figure \ref{fig:complete_generation_precision_case_0}). However, LLMs provide reasonable approximations for less stringent specifications, such as bending to wipe a one-meter high table (Figure \ref{fig:complete_generation_precision_case_1}), while MoMask generates generic wiping motions without proper height adaptation, likely because it learns average motion patterns from training data rather than interpreting specific height measurements. LLMs demonstrate solid comprehension of basic directional concepts like left or right, forward or backward, and up or down.

\paragraph{Imagination.} We find that LLMs can predict animations from animal imitation instructions, demonstrating sign of creative motion imagination. For instance, GPT-4o successfully predicts a coherent sequence of arm and leg movements that mimic a peacock displaying feathers (Figure \ref{fig:complete_generation_imagination_case_0}). While MoMask also performs adequately on this instruction, its success likely stems from exposure to similar motions in its training data (specifically, humans imitating birds with wing-like movements). Interestingly, Claude 3.5 Sonnet demonstrates more flexible conceptualization by generating plausible woodpecker-like head movements that MoMask completely fails to reproduce (Figure \ref{fig:complete_generation_imagination_case_1}). These observations suggest that LLMs possess a fundamental understanding of motion-related semantics that extends beyond simple pattern matching.

\paragraph{Cultural Awareness.} LLMs demonstrate the capability in distinguishing culturally specific motion patterns. For instance, when prompted about Japanese bowing, Claude 3.5 Sonnet correctly generates the formal full-body bow with a kneeling posture and hands positioned on the ground (Figure \ref{fig:complete_generation_culture_case_0}), differentiating it from a simple knee-bending bow (Figure \ref{fig:complete_generation_culture_case_1}). In contrast, MoMask produces similar general bowing motions for both instructions, suggesting its limited cultural understanding.

\vspace{-0.15em}
\paragraph{Generating Raw Avatar Control Parameters.} We further examine LLMs' capabilities to directly generate SMPL control parameters, with prompting strategies illustrated in Appendix \ref{appendix:raw_control_params}. Our analysis reveals three key limitations. First, LLMs generate timing for high-level steps primarily in whole and half seconds (Figure \ref{fig:raw_control_params_timing_0}, \ref{fig:raw_control_params_timing_1}), lacking precise temporal control. Second, with GPT-4o as an example, the LLM shows poor comprehension of joint rotations, failing in both directional recognition and movement quantity generation (Figure \ref{fig:complete_generation_raw_params_case_0}). Third, while the LLM can roughly identify root movement directions (Figure \ref{fig:complete_generation_raw_params_case_1}), they fail to generate reasonable values, likely because human language rarely specifies body root concepts and precise global movements.
\section{Conclusion}

Through our hierarchical framework for 3D avatar control, we evaluate LLMs' human motion understanding on carefully designed representative instructions. Our findings reveal that LLMs possess substantial motion-related semantic knowledge despite limitations in precise spatial understanding, particularly for multi-step motions involving high-degree-of-freedom body parts. Breaking down movements into atomic components improves both high-level conceptual planning and low-level body part coordination, yet LLMs fall short in handling precise spatial specifications and generating accurate spatial-temporal parameters for avatar control. Notably, LLMs show promise in conceptualizing creative motions and distinguishing culturally-specific motion patterns.

\paragraph{Practical Value \& Future Directions.} Our work suggests LLMs could enhance natural language interfaces for avatar control and culturally-aware, creative motion synthesis. To address spatial precision limitations, we propose a hybrid approach: using LLMs for high-level planning and motion decomposition, while delegating low-level execution to specialized motion models \citep{MDM, momask} that handle joint dynamics (e.g., velocity, acceleration) and physical constraints (e.g., foot contact, collision). This leverages LLMs' semantic strengths with motion models' spatial precision, enabling interpretable language-based control with enhanced motion realism. While implementing such a system exceeds our current scope of evaluating LLMs' human motion knowledge, we hope our findings provide insights for future development.
\section*{Limitations}

Our work, as the first exploration of LLMs' motion understanding through avatar control, faces two main limitations. First, due to the high workload and cost of human evaluation, we focus on twenty representative motion instructions. While providing initial insights into LLMs' motion understanding, they are limited in scale. A more comprehensive evaluation would require a larger dataset covering a broader range of motion scenarios and edge cases. Second, while our choice of linear interpolation between keyframes enables clear verification of LLMs' human motion understanding, it produces mechanical movements not immediately ready for practical applications. Further research could investigate hybrid approaches that combine LLMs' sophisticated high-level motion understanding with specialized motion synthesis models to generate more natural animations.
\section*{Ethics Statement}
This work demonstrates both the potential and limitations of using LLMs for human body movement prediction, which has implications for various fields including animation, robotics, and human-computer interaction. While the ability to generate human movements from natural language could democratize animation creation, it also raises potential risks. The technology could be misused to create misleading or deceptive content, particularly in combination with other AI tools for digital human generation. There are also ethical considerations around consent and representation, as such systems could potentially reproduce and amplify biases in human movement patterns. Additionally, as these technologies become more sophisticated, there may be privacy concerns regarding the capture and reproduction of distinctive individual movement styles. Therefore, future development in this area should carefully consider these ethical implications and incorporate appropriate safeguards.
\section*{Acknowledgments}
This work was partially supported by the “R\&D Hub Aimed at Ensuring Transparency and Reliability of Generative AI Models” project of the Ministry of Education, Culture, Sports, Science and Technology.
We acknowledge the assistance of Claude 3.5 Sonnet and GPT-4o for manuscript polishing and visualization code development. We would like to thank Bowen Chen for helpful discussion. We would also like to express our appreciation to all annotators for their valuable contributions and dedication to this work.

\bibliography{custom}

\appendix
\newpage
\onecolumn
\section{Appendix}
\label{sec:appendix}

\renewcommand{\thefigure}{A\arabic{figure}}
\renewcommand{\thetable}{A\arabic{table}}
\setcounter{figure}{0}
\setcounter{table}{0}

\subsection{Prompts}
\label{appendix:prompts}

\captionof{figure}{The list of prompt templates, with \texttt{\textcolor{blue}{[blue]}} indicating the prompt type, \texttt{\textcolor{orange}{<orange>}} as an illustrative label indicating the prompt function (not used in the actual LLM querying), and \texttt{\textcolor{red}{\{red\}}} indicating the placeholder for corresponding contents. \texttt{\textcolor{red}{\{position\}}}, \texttt{\textcolor{red}{\{description\}}} and \texttt{\textcolor{red}{\{positions with descriptions\}}} are all taken from Table \ref{tab:body_part_positions_descriptions}.}
\label{fig:prompts}

\lstset{style=MyListingStyle_0}
\begin{lstlisting}[escapeinside={(*@}{@*)}]
@[System Prompt]@

You will be given a textual human motion instruction, followed by a sequence of clarification questions about different aspects about the motion. You should use your daily knowledge about human motions to answer the questions accurately and concisely.



@[High-level Planning Prompts] (piece_by_piece)@

+<setup>+
The human initially stands naturally with arms hanging beside the body. The textual human motion instruction is "!{motion instruction}!".

+<movement>+
What are the movements of relevant body parts in Step!{step number}!? The movements should be simple enough to be only **single-directional**.

+<initial_state>+
What are the initial states of relevant body parts in Step!{step number}!?

+<final_state>+
What are the final states of relevant body parts in Step!{step number}!?

+<timing>+
How long does Step!{step number}! last in the second unit?

+<is_end>+
Is it the end of this motion?



@[High-level Planning Prompts] (in_one_go)@

The human initially stands naturally with arms hanging beside the body. The textual human motion instruction is "!{motion instruction}!". Decompose it step-by-step with three language descriptions for each step (one for the initial state of moved body parts, one for the final state of moved body parts and one for the movement). Each step should be simple enough to include only **single-direction** motions for all moved body parts. Estimate a time range in the second unit for each step (the end time of the last step should exactly be the start time of the next step).



@[Low-level Planning Prompts] (hierarchical)@

+<step_setup>+
The human initially stands naturally with arms hanging beside the body. The textual human motion instruction is "!{motion instruction}!". In the high-leve plan of Step!{step number}!, the initial states of relevant body parts are "!{initial states}!", the final states of relevant body parts are "!{final states}!", and the movements of relevant body parts are "!{movements}!".

+<language_description>+
The last position of !{body part}! is **!{position}!** (!{description}!). Describe the movement of this body part during Step!{step number}! and final position at the end of the step in language.

+<position_choice>+
Details in Figure (*@\ref{fig:hierarchical_prompts}@*)

+<reflection_analysis>+
Analyze this body part with its planned next position. Is this body part necessary for this step? If so, does the planned next position of this body part achieve the goal final state in the high-level plan?

+<reflection_judgement>+
Do you think there's need to replan this body part in order to achieve the goal final state in the high-level plan? Give your judgement.

+<correction>+
You think that: !{reflection}!. So the next position of !{body part}! should not be **!{position}!**.\newline Based on the thought, replan this body part in Step!{step number}!.



@[Low-level Planning Prompts] (one_by_one)@

+<step_setup>+
The human initially stands naturally with arms hanging beside the body. The textual human motion instruction is "!{motion instruction}!". In the high-leve plan of Step!{step number}!, the initial states of relevant body parts are "!{initial states}!", the final states of relevant body parts are "!{final states}!", and the movements of relevant body parts are "!{movements}!".

+<language_description>+
The last position of !{body part}! is **!{position}!** (!{description}!). Describe the movement of this body part during Step!{step number}! and final position at the end of the step in language.

+<position_choice>+
The last position of !{body part}! is **!{position}!** (!{description}!). Is the next position **!{position}!** (!{description}!)?

+<reflection_analysis>+
Analyze this body part with its planned next position. Is this body part necessary for this step? If so, does the planned next position of this body part achieve the goal final state in the high-level plan?

+<reflection_judgement>+
Do you think there's need to replan this body part in order to achieve the goal final state in the high-level plan? Give your judgement.

+<correction>+
You think that: !{reflection}!. So the next position of !{body part}! should not be **!{position}!**.\newline Based on the thought, replan this body part in Step!{step number}!.



@[Low-level Planning Prompts] (all)@

+<step_setup>+
The human initially stands naturally with arms hanging beside the body. The textual human motion instruction is "!{motion instruction}!". In the high-leve plan of Step!{step number}!, the initial states of relevant body parts are "!{initial states}!", the final states of relevant body parts are "!{final states}!", and the movements of relevant body parts are "!{movements}!".

+<language_description>+
The last position of !{body part}! is **!{position}!** (!{description}!). Describe the movement of this body part during Step!{step number}! and final position at the end of the step in language.

+<position_choice>+
There are multiple possible positions for !{body part}!:
!{positions with descriptions}!
The last position of this body part is **!{position}!**. Choose the next position from the options above.

+<reflection_analysis>+
Analyze this body part with its planned next position. Is this body part necessary for this step? If so, does the planned next position of this body part achieve the goal final state in the high-level plan?

+<reflection_judgement>+
Do you think there's need to replan this body part in order to achieve the goal final state in the high-level plan? Give your judgement.

+<correction>+
You think that: !{reflection}!. So the next position of !{body part}! should not be **!{position}!**.\newline Based on the thought, replan this body part in Step!{step number}!.
\end{lstlisting}

\captionof{figure}{The hierarchical prompts for \texttt{\textcolor{orange}{<position\_choice>}} of \texttt{\textcolor{blue}{[Low-level Planning Prompts] (hierarchical)}}. Each question comes with several options. If the value of the selected option is one position string, the hierarchical querying returns this position. Otherwise the LLM continues to ask the nested question.}
\label{fig:hierarchical_prompts}
\lstset{style=MyListingStyle_1}
\begin{lstlisting}
"Head": {
    "question": "At the end of this step, is the head upright in the neutral position, tilted left, tilted right, tilted down, tilted up, turned left or turned right? Choose one from %s",
    "options": {
        "neutral": "neutral",
        "tilted_left": {
            "question": "Is the head tilted left slightly or fully? Choose one from %s",
            "options": {
                "tilted_left_slightly": "tilted_left_slightly",
                "tilted_left_fully": "tilted_left_fully"
            }
        },
        "tilted_right": {
            "question": "Is the head tilted right slightly or fully? Choose one from %s",
            "options": {
                "tilted_right_slightly": "tilted_right_slightly",
                "tilted_right_fully": "tilted_right_fully"
            }
        },
        "tilted_down": {
            "question": "Is the head tilted down slightly or fully? Choose one from %s",
            "options": {
                "tilted_down_slightly": "tilted_down_slightly",
                "tilted_down_fully": "tilted_down_fully"
            }
        },
        "tilted_up": {
            "question": "Is the head tilted up slightly or fully? Choose one from %s",
            "options": {
                "tilted_up_slightly": "tilted_up_slightly",
                "tilted_up_fully": "tilted_up_fully"
            }
        },
        "turned_left": {
            "question": "Is the head turned left slightly or fully? Choose one from %s",
            "options": {
                "turned_left_slightly": "turned_left_slightly",
                "turned_left_fully": "turned_left_fully"
            }
        },
        "turned_right": {
            "question": "Is the head turned right slightly or fully? Choose one from %s",
            "options": {
                "turned_right_slightly": "turned_right_slightly",
                "turned_right_fully": "turned_right_fully"
            }
        }
    }
},
"Torso": {
    "question": "At the end of this step, is the torso upright in the neutral position, bent forward, bent backward, tilted left, tilted right, twisted left or twisted right? Choose one from %s",
    "options": {
        "neutral": "neutral",
        "bent_backward": "bent_backward",
        "bent_forward": {
            "question": "Is the torso bent forward slightly or fully? Choose one from %s",
            "options": {
                "bent_forward_slightly": "bent_forward_slightly",
                "bent_forward_fully": "bent_forward_fully"
            }
        },
        "tilted_left": {
            "question": "Is the torso tilted left slightly or fully? Choose one from %s",
            "options": {
                "tilted_left_slightly": "tilted_left_slightly",
                "tilted_left_fully": "tilted_left_fully"
            }
        },
        "tilted_right": {
            "question": "Is the torso tilted right slightly or fully? Choose one from %s",
            "options": {
                "tilted_right_slightly": "tilted_right_slightly",
                "tilted_right_fully": "tilted_right_fully"
            }
        },
        "twisted_left": {
            "question": "Is the torso twisted left slightly or fully? Choose one from %s",
            "options": {
                "twisted_left_slightly": "twisted_left_slightly",
                "twisted_left_fully": "twisted_left_fully"
            }
        },
        "twisted_right": {
            "question": "Is the torso twisted right slightly or fully? Choose one from %s",
            "options": {
                "twisted_right_slightly": "twisted_right_slightly",
                "twisted_right_fully": "twisted_right_fully"
            }
        }
    }
},
"LeftUpperArm": {
    "question": "At the end of this step, relative to the torso, is the left upper arm neutrally resting by the side of the body, straight upward, straight forward, straight out to the side forming a right angle with the torso, or in other in-between positions? Choose one from %s",
    "options": {
        "neutral": "neutral",
        "upward": "upward",
        "forward": {
            "question": "Is the left elbow pit facing inward to the midline of the body or upward relative to the torso? Choose one from %s",
            "options": {
                "forward_elbowpit_inward": "forward_elbowpit_inward",
                "forward_elbowpit_upward": "forward_elbowpit_upward"
            }
        },
        "side": {
            "question": "Is the left elbow pit facing forward or upward relative to the torso? Choose one from %s",
            "options": {
                "side_elbowpit_forward": "side_elbowpit_forward",
                "side_elbowpit_upward": "side_elbowpit_upward"
            }
        },
        "in_between_positions": {
            "question": "Relative to the torso, is the left upper arm out to the side, towards the midline of the body, or neither? Choose one from %s",
            "options": {
                "out_to_side": {
                    "question": "Relative to the torso, is the left upper arm in front of the body, behind the body, or neither? Choose one from %s",
                    "options": {
                        "front": {
                            "question": "Relative to the torso, is the left upper arm above the left shoulder, below the left shoulder, or neither? Choose one from %s",
                            "options": {
                                "above": "forward_to_upward_side",
                                "below": "neutral_to_forward_side",
                                "neither": "forward_to_side"
                            }
                        },
                        "behind": {
                            "question": "Relative to the torso, is the left upper arm above the left shoulder, below the left shoulder, or neither? Choose one from %s",
                            "options": {
                                "above": "side_to_upward_back",
                                "below": "neutral_to_backward_side",
                                "neither": "side_to_back"
                            }
                        },
                        "neither": {
                            "question": "Relative to the torso, is the left upper arm above the left shoulder or below the left shoulder? Choose one from %s",
                            "options": {
                                "above": "upward_to_side",
                                "below": "neutral_to_side"
                            }
                        }
                    }
                },
                "towards_midline": {
                    "question": "Relative to the torso, is the left upper arm above the left shoulder, below the left shoulder, or neither? Choose one from %s",
                    "options": {
                        "above": "forward_to_upward_midline",
                        "below": "neutral_to_forward_midline",
                        "neither": "forward_to_midline"
                    }
                },
                "neither": {
                    "question": "Relative to the torso, is the left upper arm in front of the body or behind the body? Choose one from %s",
                    "options": {
                        "front": {
                            "question": "Relative to the torso, is the left upper arm above the left shoulder or below the left shoulder? Choose one from %s",
                            "options": {
                                "above": "forward_to_upward",
                                "below": "neutral_to_forward"
                            }
                        },
                        "behind": "neutral_to_back"
                    }
                }
            }
        }
    }
},
"RightUpperArm": {
    "question": "At the end of this step, relative to the torso, is the right upper arm neutrally resting by the side of the body, straight upward, straight forward, straight out to the side forming a right angle with the torso, or in other in-between positions? Choose one from %s",
    "options": {
        "neutral": "neutral",
        "upward": "upward",
        "forward": {
            "question": "Is the right elbow pit facing inward to the midline of the body or upward relative to the torso? Choose one from %s",
            "options": {
                "forward_elbowpit_inward": "forward_elbowpit_inward",
                "forward_elbowpit_upward": "forward_elbowpit_upward"
            }
        },
        "side": {
            "question": "Is the right elbow pit facing forward or upward relative to the torso? Choose one from %s",
            "options": {
                "side_elbowpit_forward": "side_elbowpit_forward",
                "side_elbowpit_upward": "side_elbowpit_upward"
            }
        },
        "in_between_positions": {
            "question": "Relative to the torso, is the right upper arm out to the side, towards the midline of the body, or neither? Choose one from %s",
            "options": {
                "out_to_side": {
                    "question": "Relative to the torso, is the right upper arm in front of the body, behind the body, or neither? Choose one from %s",
                    "options": {
                        "front": {
                            "question": "Relative to the torso, is the right upper arm above the right shoulder, below the right shoulder, or neither? Choose one from %s",
                            "options": {
                                "above": "forward_to_upward_side",
                                "below": "neutral_to_forward_side",
                                "neither": "forward_to_side"
                            }
                        },
                        "behind": {
                            "question": "Relative to the torso, is the right upper arm above the right shoulder, below the right shoulder, or neither? Choose one from %s",
                            "options": {
                                "above": "side_to_upward_back",
                                "below": "neutral_to_backward_side",
                                "neither": "side_to_back"
                            }
                        },
                        "neither": {
                            "question": "Relative to the torso, is the right upper arm above the right shoulder or below the right shoulder? Choose one from %s",
                            "options": {
                                "above": "upward_to_side",
                                "below": "neutral_to_side"
                            }
                        }
                    }
                },
                "towards_midline": {
                    "question": "Relative to the torso, is the right upper arm above the right shoulder, below the right shoulder, or neither? Choose one from %s",
                    "options": {
                        "above": "forward_to_upward_midline",
                        "below": "neutral_to_forward_midline",
                        "neither": "forward_to_midline"
                    }
                },
                "neither": {
                    "question": "Relative to the torso, is the right upper arm in front of the body or behind the body? Choose one from %s",
                    "options": {
                        "front": {
                            "question": "Relative to the torso, is the right upper arm above the right shoulder or below the right shoulder? Choose one from %s",
                            "options": {
                                "above": "forward_to_upward",
                                "below": "neutral_to_forward"
                            }
                        },
                        "behind": "neutral_to_back"
                    }
                }
            }
        }
    }
},
"LeftElbow": {
    "question": "At the end of this step, is the left elbow stright or bent? Choose one from %s",
    "options": {
        "straight": "neutral",
        "bent": {
            "question": "Is the left elbow slightly bent in, bent in 90 degrees or fully bent? Choose one from %s",
            "options": {
                "slightly_bent_in": "slightly_bent_in",
                "bent_in_90_degrees": "bent_in_90_degrees",
                "fully_bent": "fully_bent"
            }
        }
    }
},
"RightElbow": {
    "question": "At the end of this step, is the right elbow stright or bent? Choose one from %s",
    "options": {
        "straight": "neutral",
        "bent": {
            "question": "Is the right elbow slightly bent in, bent in 90 degrees or fully bent? Choose one from %s",
            "options": {
                "slightly_bent_in": "slightly_bent_in",
                "bent_in_90_degrees": "bent_in_90_degrees",
                "fully_bent": "fully_bent"
            }
        }
    }
},
"LeftWrist": {
    "question": "At the end of this step, is the left wrist straight in the neutral position, bent vertically, or tilted sideways? Choose one from %s",
    "options": {
        "neutral": "neutral",
        "bent_vertically": {
            "question": "Is the left wrist bent upward so that the back of the left hand is closer to the back of the left forearm, with the muscles of the back of the left forearm contracted? Or is the left wrist bent downward so that the left palm moves towards the left forearm? Choose one from %s",
            "options": {
                "bent_upward": "bent_upward",
                "bent_downward": {
                    "question": "Is the left wrist slightly bent downward or fully? Choose one from %s",
                    "options": {
                        "bent_slightly_downward": "bent_slightly_downward",
                        "fully_bent_downward": "fully_bent_downward"
                    }
                }
            }
        },
        "tilted_sideways": {
            "question": "Is the left wrist tilted laterally towards the left thumb, or the left little finger? Choose one from %s",
            "options": {
                "tilted_towards_thumb_side": "tilted_towards_thumb_side",
                "tilted_towards_pinky_side": "tilted_towards_pinky_side"
            }
        }
    }
},
"RightWrist": {
    "question": "At the end of this step, is the right wrist straight in the neutral position, bent vertically, or tilted sideways? Choose one from %s",
    "options": {
        "neutral": "neutral",
        "bent_vertically": {
            "question": "Is the right wrist bent upward so that the back of the right hand is closer to the back of the right forearm, with the muscles of the back of the right forearm contracted? Or is the right wrist bent downward so that the right palm moves towards the right forearm? Choose one from %s",
            "options": {
                "bent_upward": "bent_upward",
                "bent_downward": {
                    "question": "Is the right wrist slightly bent downward or fully? Choose one from %s",
                    "options": {
                        "bent_slightly_downward": "bent_slightly_downward",
                        "fully_bent_downward": "fully_bent_downward"
                    }
                }
            }
        },
        "tilted_sideways": {
            "question": "Is the right wrist tilted laterally towards the right thumb, or the right little finger? Choose one from %s",
            "options": {
                "tilted_towards_thumb_side": "tilted_towards_thumb_side",
                "tilted_towards_pinky_side": "tilted_towards_pinky_side"
            }
        }
    }
},
"LeftUpperLeg": {
    "question": "At the end of this step, is the left upper leg neutrally aligned with the body midline, straight forward, straight out to the side forming a right angle with the torso, or in other in-between positions? Choose one from %s",
    "options": {
        "neutral": "neutral",
        "forward": "forward",
        "side": "side",
        "in_between_positions": {
            "question": "Relative to the torso, is the left upper leg out to the side, towards the midline of the body, or neither? Choose one from %s",
            "options": {
                "out_to_side" : {
                    "question": "Relative to the torso, is the left upper leg in front of the body, behind the body, or neither? Choose one from %s",
                    "options": {
                        "front": {
                            "question": "Relative to the torso, is the left upper leg above the left pelvis, below the left pelvis, or neither? Choose one from %s",
                            "options": {
                                "above": "forward_to_upward_side",
                                "below": "neutral_to_forward_side",
                                "neither": "forward_to_side"
                            }
                        },
                        "behind": "neutral_to_backward_side",
                        "neither": "neutral_to_side"
                    }
                },
                "towards_midline": {
                    "question": "Relative to the torso, is the left upper leg in front of the body or behind the body? Choose one from %s",
                    "options": {
                        "front": {
                            "question": "Relative to the torso, is the left upper leg above the left pelvis, below the left pelvis, or neither? Choose one from %s",
                            "options": {
                                "above": "forward_to_upward_midline",
                                "below": "neutral_to_forward_midline",
                                "neither": "forward_to_midline"
                            }
                        },
                        "behind": "neutral_to_backward_midline"
                    }
                },
                "neither": {
                    "question": "Relative to the torso, is the left upper leg in front of the body or behind the body? Choose one from %s",
                    "options": {
                        "front": {
                            "question": "Relative to the torso, is the left upper leg above the left pelvis or below the left pelvis? Choose one from %s",
                            "options": {
                                "above": "forward_to_upward",
                                "below": "neutral_to_forward"
                            }
                        },
                        "behind": "neutral_to_backward"
                    }
                }
            }
        }
    }
},
"RightUpperLeg": {
    "question": "At the end of this step, is the right upper leg neutrally aligned with the body midline, straight forward, straight out to the side forming a right angle with the torso, or in other in-between positions? Choose one from %s",
    "options": {
        "neutral": "neutral",
        "forward": "forward",
        "side": "side",
        "in_between_positions": {
            "question": "Relative to the torso, is the right upper leg out to the side, towards the midline of the body, or neither? Choose one from %s",
            "options": {
                "out_to_side" : {
                    "question": "Relative to the torso, is the right upper leg in front of the body, behind the body, or neither? Choose one from %s",
                    "options": {
                        "front": {
                            "question": "Relative to the torso, is the right upper leg above the right pelvis, below the right pelvis, or neither? Choose one from %s",
                            "options": {
                                "above": "forward_to_upward_side",
                                "below": "neutral_to_forward_side",
                                "neither": "forward_to_side"
                            }
                        },
                        "behind": "neutral_to_backward_side",
                        "neither": "neutral_to_side"
                    }
                },
                "towards_midline": {
                    "question": "Relative to the torso, is the right upper leg in front of the body or behind the body? Choose one from %s",
                    "options": {
                        "front": {
                            "question": "Relative to the torso, is the right upper leg above the right pelvis, below the right pelvis, or neither? Choose one from %s",
                            "options": {
                                "above": "forward_to_upward_midline",
                                "below": "neutral_to_forward_midline",
                                "neither": "forward_to_midline"
                            }
                        },
                        "behind": "neutral_to_backward_midline"
                    }
                },
                "neither": {
                    "question": "Relative to the torso, is the right upper leg in front of the body or behind the body? Choose one from %s",
                    "options": {
                        "front": {
                            "question": "Relative to the torso, is the right upper leg above the right pelvis or below the right pelvis? Choose one from %s",
                            "options": {
                                "above": "forward_to_upward",
                                "below": "neutral_to_forward"
                            }
                        },
                        "behind": "neutral_to_backward"
                    }
                }
            }
        }
    }
},
"LeftKnee": {
    "question": "At the end of this step, is the left knee stright or bent? Choose one from %s",
    "options": {
        "straight": "neutral",
        "bent": {
            "question": "Is the left knee slightly bent, bent at 90 degrees or fully bent? Choose one from %s",
            "options": {
                "slightly_bent": "slightly_bent",
                "bent_at_90_degrees": "bent_at_90_degrees",
                "fully_bent": "fully_bent"
            }
        }
    }
},
"RightKnee": {
    "question": "At the end of this step, is the right knee stright or bent? Choose one from %s",
    "options": {
        "straight": "neutral",
        "bent": {
            "question": "Is the right knee slightly bent, bent at 90 degrees or fully bent? Choose one from %s",
            "options": {
                "slightly_bent": "slightly_bent",
                "bent_at_90_degrees": "bent_at_90_degrees",
                "fully_bent": "fully_bent"
            }
        }
    }
},
"LeftAnkle": {
    "question": "At the end of this step, is the left ankle in a neutral position like when standing? Or is the left ankle bent vertically? Or is the left ankle tilted so that the sole moves towards the side or midline? Choose one from %s",
    "options": {
        "neutral": "neutral",
        "bent_vertically": {
            "question": "Is the left ankle bent upward so that the top of the left foot forms an acute angle with the left shin? Or is the left ankle bent downward so that the top of the left foot forms an obtuse angle with the left shin? Choose one from %s",
            "options": {
                "bent_upward": "bent_upward",
                "bent_downward": "bent_downward"
            }
        },
        "tilted_inward_or_outward": {
            "question": "Is the left ankle tilted inwards so that the left sole moves towards the midline of the body? Or is the left ankle tilted outwards so that the left sole moves away from the midline of the body? Choose one from %s",
            "options": {
                "tilted_inward": "tilted_inward",
                "tilted_outward": "tilted_outward"
            }
        }
    }
},
"RightAnkle": {
    "question": "At the end of this step, is the right ankle in a neutral position like when standing? Or is the right ankle bent vertically? Or is the right ankle tilted so that the sole moves towards the side or midline? Choose one from %s",
    "options": {
        "neutral": "neutral",
        "bent_vertically": {
            "question": "Is the right ankle bent upward so that the top of the right foot forms an acute angle with the right shin? Or is the right ankle bent downward so that the top of the right foot forms an obtuse angle with the right shin? Choose one from %s",
            "options": {
                "bent_upward": "bent_upward",
                "bent_downward": "bent_downward"
            }
        },
        "tilted_inward_or_outward": {
            "question": "Is the right ankle tilted inwards so that the right sole moves towards the midline of the body? Or is the right ankle tilted outwards so that the right sole moves away from the midline of the body? Choose one from %s",
            "options": {
                "tilted_inward": "tilted_inward",
                "tilted_outward": "tilted_outward"
            }
        }
    }
},
"LeftToes": {
    "question": "At the end of this step, are the left toes in the neutral position or curled? Choose one from %s",
    "options": {
        "neutral": "neutral",
        "curled": {
            "question": "Are the left toes curled up or down? Choose one from %s",
            "options": {
                "curled_up": "curled_up",
                "curled_down": "curled_down"
            }
        }
    }
},
"RightToes": {
    "question": "At the end of this step, are the right toes in the neutral position or curled? Choose one from %s",
    "options": {
        "neutral": "neutral",
        "curled": {
            "question": "Are the right toes curled up or down? Choose one from %s",
            "options": {
                "curled_up": "curled_up",
                "curled_down": "curled_down"
            }
        }
    }
}
\end{lstlisting}

\newpage
\scriptsize
\begin{longtable}{|c|c|p{10cm}|}
\caption{The language descriptions of body part positions.} \label{tab:body_part_positions_descriptions} \\
\hline
\textbf{Body Part} & \textbf{Position} & \textbf{Description} \\
\hline
\endfirsthead

\hline
\textbf{Body Part} & \textbf{Position} & \textbf{Description} \\
\hline
\endhead

\hline
\endfoot

\hline
\endlastfoot

Head & neutral & The head is level with the spine and faces forward relative to the torso. The chin is neither raised nor lowered, forming a right angle (90 degrees) with the neck. The back of the head is aligned with the upper back, maintaining a straight, neutral posture. \\
\cline{2-3}
& tilted\_up\_slightly & The chin forms a slightly obtuse angle, slightly more than 90 degrees, with the neck. The back of the head is slightly closer to the upper back, making the angle between the back of the head and the upper back slightly more than 90 degrees. \\
\cline{2-3}
& tilted\_up\_fully & The chin forms a significantly obtuse angle with the neck. The back of the head almost touches the upper back. \\
\cline{2-3}
& tilted\_down\_slightly & The chin forms a slightly acute angle, slightly less than 90 degrees, with the neck. \\
\cline{2-3}
& tilted\_down\_fully & The chin nearly touches the chest, forming a very acute angle, often less than 45 degrees, with the neck. \\
\cline{2-3}
& turned\_left\_slightly & The nose points just a bit to the left of the body's midline. The angle between the nose's direction and the body's central axis is small, resulting in a slight turn. \\
\cline{2-3}
& turned\_left\_fully & The nose points directly towards or over the left shoulder, forming a nearly right angle of approximately 90 degrees with the body's midline. \\
\cline{2-3}
& turned\_right\_slightly & The nose points just a bit to the right of the body's midline. The angle between the nose's direction and the body's central axis is small, resulting in a slight turn. \\
\cline{2-3}
& turned\_right\_fully & The nose points directly towards or over the right shoulder, forming a nearly right angle of approximately 90 degrees with the body's midline. \\
\cline{2-3}
& tilted\_left\_slightly & The left ear forms a small angle, around 45 degrees, with the left shoulder. The right ear forms a larger, obtuse angle with the right shoulder. \\
\cline{2-3}
& tilted\_left\_fully & The left ear nearly touches the left shoulder, forming an angle smaller than 20 degrees. The right ear forms an almost straight angle, nearing 180 degrees, with the right shoulder. \\
\cline{2-3}
& tilted\_right\_slightly & The right ear forms a small angle, around 45 degrees, with the right shoulder. The left ear forms a larger, obtuse angle with the left shoulder. \\
\cline{2-3}
& tilted\_right\_fully & The right ear nearly touches the right shoulder, forming an angle smaller than 20 degrees. The left ear forms an almost straight angle, nearing 180 degrees, with the left shoulder. \\
\hline

Torso & neutral & The waist is in an upright position, aligned with the spine. The pelvis and torso maintain a straight posture. \\
\cline{2-3}
& twisted\_left\_slightly & The left side of the waist moves slightly backward, and the right side moves slightly forward, forming a small angle to the left relative to the feet. The muscles on the right side extend mildly, while those on the left side contract slightly. \\
\cline{2-3}
& twisted\_left\_fully & The left side of the waist moves backward, and the right side moves forward, causing a rotational angle to the left relative to the feet. The muscles on the right side of the lower back and abdomen extend slightly, while the muscles on the left side contract lightly. \\
\cline{2-3}
& twisted\_right\_slightly & The right side of the waist moves slightly backward, and the left side moves slightly forward, forming a small angle to the right relative to the feet. The muscles on the left side extend mildly, while those on the right side contract slightly. \\
\cline{2-3}
& twisted\_right\_fully & The right side of the waist moves backward, and the left side moves forward, causing a rotational angle to the right relative to the feet. The muscles on the left side of the lower back and abdomen extend slightly, while the muscles on the right side contract lightly. \\
\cline{2-3}
& bent\_forward\_slightly & The waist forms a slight angle, around 45 degrees, with the thighs. The lower abdomen is closer to the thighs, and the muscles in the lower back lengthen slightly while those in the lower abdomen contract a little. \\
\cline{2-3}
& bent\_forward\_fully & The waist forms a right angle, around 90 degrees, with the thighs. The muscles in the lower back stretch significantly, while those in the abdomen contract fully. \\
\cline{2-3}
& bent\_backward & The waist forms a pronounced outward curve, creating a significant arch in the lower back. The front side angle of the waist increases markedly, with the muscles in the lower back fully contracted and those in front maximally extended. \\
\cline{2-3}
& tilted\_left\_slightly & The left side of the waist moves slightly downward, and the right side is slightly raised, forming a small angle. The right side muscles extend a bit, while the left muscles contract slightly. \\
\cline{2-3}
& tilted\_left\_fully & The left side of the waist moves downward, forming a right angle with the pelvis, with the right side raised. The right body muscles stretch, while the left muscles contract. \\
\cline{2-3}
& tilted\_right\_slightly & The right side of the waist moves slightly downward, and the left side is slightly raised, forming a small angle. The left side muscles extend a bit, while the right muscles contract slightly. \\
\cline{2-3}
& tilted\_right\_fully & The right side of the waist moves downward, forming a right angle with the pelvis, with the left side raised. The left body muscles stretch, while the right muscles contract. \\
\hline

LeftUpperArm & neutral & The left upper arm is relaxed and hanging straight down by the side of the body, parallel to the torso. \\
\cline{2-3}
& forward\_elbowpit\_inward & The left upper arm is extended straight forward relative to the torso, parallel to the ground and perpendicular to the torso. The elbow pit faces inward to the midline of the body. \\
\cline{2-3}
& forward\_elbowpit\_upward & The left upper arm is extended straight forward relative to the torso, parallel to the ground and perpendicular to the torso. The elbow pit faces upward relative to the torso. \\
\cline{2-3}
& upward & The left upper arm is lifted straight upwards, close to the ear, reaching towards the sky. \\
\cline{2-3}
& side\_elbowpit\_forward & The left upper arm is extended straight out to the side, forming a right angle with the torso (horizontally aligned with the shoulders). The elbow pit faces forward. \\
\cline{2-3}
& side\_elbowpit\_upward & The left upper arm is extended straight out to the side, forming a right angle with the torso (horizontally aligned with the shoulders). The elbow pit faces upward. \\
\cline{2-3}
& neutral\_to\_forward & The left upper arm is raised from the neutral position towards the forward position, forming an approximate 45-degree angle in front of the body. \\
\cline{2-3}
& forward\_to\_upward & The left upper arm is raised from the forward position towards the upward position, forming an approximate 45-degree angle from forward towards upwards. \\
\cline{2-3}
& neutral\_to\_back & The left upper arm is raised from the neutral position towards the back, forming an approximate 45-degree angle behind the body. \\
\cline{2-3}
& forward\_to\_midline & The left upper arm is raised forward and slightly crosses the midline of the body. \\
\cline{2-3}
& forward\_to\_side & The left upper arm is extended from the forward position towards the side, forming an approximate 45-degree angle between forward and side. \\
\cline{2-3}
& side\_to\_back & The left upper arm is extended from the side position towards the back, forming an approximate 45-degree angle between side and back. \\
\cline{2-3}
& neutral\_to\_side & The left upper arm is in the neutral position but extends halfway outward to the side, forming a small angle from the torso. \\
\cline{2-3}
& upward\_to\_side & The left upper arm is extended outward from the upwards position towards the side, forming an approximate 45-degree angle between upwards and side. \\
\cline{2-3}
& forward\_to\_upward\_side & The left upper arm is raised upwards and slightly angled towards the side from the forward position, forming a diagonal line above the shoulder. \\
\cline{2-3}
& forward\_to\_upward\_midline & The left upper arm is raised upwards and slightly angled towards the midline from the forward position, forming a diagonal line above the shoulder. \\
\cline{2-3}
& side\_to\_upward\_back & The left upper arm is raised from the side position towards the back and upwards, forming a diagonal reaching upwards and backward. \\
\cline{2-3}
& neutral\_to\_forward\_midline & The left upper arm is raised upwards and slightly angled towards the midline from the neutral position, forming a diagonal line reaching up. \\
\cline{2-3}
& neutral\_to\_forward\_side & The left upper arm is raised upwards and slightly angled towards the side from the neutral position, forming a diagonal line reaching up. \\
\cline{2-3}
& neutral\_to\_backward\_side & The left upper arm is raised upwards and slightly angled towards the back and side from the neutral position, forming a diagonal line reaching up and behind the torso. \\
\hline

RightUpperArm & neutral & The right upper arm is relaxed and hanging straight down by the side of the body, parallel to the torso. \\
\cline{2-3}
& forward\_elbowpit\_inward & The right upper arm is extended straight forward relative to the torso, parallel to the ground and perpendicular to the torso. The elbow pit faces inward to the midline of the body. \\
\cline{2-3}
& forward\_elbowpit\_upward & The right upper arm is extended straight forward relative to the torso, parallel to the ground and perpendicular to the torso. The elbow pit faces upward relative to the torso. \\
\cline{2-3}
& upward & The right upper arm is lifted straight upwards, close to the ear, reaching towards the sky. \\
\cline{2-3}
& side\_elbowpit\_forward & The right upper arm is extended straight out to the side, forming a right angle with the torso (horizontally aligned with the shoulders). The elbow pit faces forward. \\
\cline{2-3}
& side\_elbowpit\_upward & The right upper arm is extended straight out to the side, forming a right angle with the torso (horizontally aligned with the shoulders). The elbow pit faces upward. \\
\cline{2-3}
& neutral\_to\_forward & The right upper arm is raised from the neutral position towards the forward position, forming an approximate 45-degree angle in front of the body. \\
\cline{2-3}
& forward\_to\_upward & The right upper arm is raised from the forward position towards the upward position, forming an approximate 45-degree angle from forward towards upwards. \\
\cline{2-3}
& neutral\_to\_back & The right upper arm is raised from the neutral position towards the back, forming an approximate 45-degree angle behind the body. \\
\cline{2-3}
& forward\_to\_midline & The right upper arm is raised forward and slightly crosses the midline of the body. \\
\cline{2-3}
& forward\_to\_side & The right upper arm is extended from the forward position towards the side, forming an approximate 45-degree angle between forward and side. \\
\cline{2-3}
& side\_to\_back & The right upper arm is extended from the side position towards the back, forming an approximate 45-degree angle between side and back. \\
\cline{2-3}
& neutral\_to\_side & The right upper arm is in the neutral position but extends halfway outward to the side, forming a small angle from the torso. \\
\cline{2-3}
& upward\_to\_side & The right upper arm is extended outward from the upwards position towards the side, forming an approximate 45-degree angle between upwards and side. \\
\cline{2-3}
& forward\_to\_upward\_side & The right upper arm is raised upwards and slightly angled towards the side from the forward position, forming a diagonal line above the shoulder. \\
\cline{2-3}
& forward\_to\_upward\_midline & The right upper arm is raised upwards and slightly angled towards the midline from the forward position, forming a diagonal line above the shoulder. \\
\cline{2-3}
& side\_to\_upward\_back & The right upper arm is raised from the side position towards the back and upwards, forming a diagonal reaching upwards and backward. \\
\cline{2-3}
& neutral\_to\_forward\_midline & The right upper arm is raised upwards and slightly angled towards the midline from the neutral position, forming a diagonal line reaching up. \\
\cline{2-3}
& neutral\_to\_forward\_side & The right upper arm is raised upwards and slightly angled towards the side from the neutral position, forming a diagonal line reaching up. \\
\cline{2-3}
& neutral\_to\_backward\_side & The right upper arm is raised upwards and slightly angled towards the back and side from the neutral position, forming a diagonal line reaching up and behind the torso. \\
\hline

LeftElbow & neutral & The left elbow is extended naturally, forming a straight line from the left shoulder to the left wrist. The left upper arm and left forearm create a nearly straight alignment of about 180 degrees. \\
\cline{2-3}
& slightly\_bent\_in & The left forearm forms a slightly obtuse angle with the left upper arm. The left hand moves slightly closer to the left elbow; the muscles in the left upper arm contract slightly. \\
\cline{2-3}
& bent\_in\_90\_degrees & The left forearm forms a right angle with the left upper arm. The muscles in the left upper arm are moderately contracted. \\
\cline{2-3}
& fully\_bent & The left forearm nearly touches or touches the left upper arm, forming a very acute angle close to zero degrees. The left hand is very close to or touching the left shoulder; the muscles in the left upper arm are fully contracted. \\
\hline

RightElbow & neutral & The right elbow is extended naturally, forming a straight line from the right shoulder to the right wrist. The right upper arm and right forearm create a nearly straight alignment of about 180 degrees. \\
\cline{2-3}
& slightly\_bent\_in & The right forearm forms a slightly obtuse angle with the right upper arm. The right hand moves slightly closer to the right elbow; the muscles in the right upper arm contract slightly. \\
\cline{2-3}
& bent\_in\_90\_degrees & The right forearm forms a right angle with the right upper arm. The muscles in the right upper arm are moderately contracted. \\
\cline{2-3}
& fully\_bent & The right forearm nearly touches or touches the right upper arm, forming a very acute angle close to zero degrees. The right hand is very close to or touching the right shoulder; the muscles in the right upper arm are fully contracted. \\
\hline

LeftWrist & neutral &  The left wrist extends straight, aligned with the left forearm, forming a continuous straight line from the left elbow to the left hand. The angle between the left forearm and left wrist is close to 180 degrees. \\
\cline{2-3}
& bent\_upward & The left wrist forms a right upward angle with the left forearm. The muscles on the back of the left forearm are fully contracted. \\
\cline{2-3}
& bent\_slightly\_downward & The left wrist forms a small downward angle, less than 45 degrees. The left palm moves slightly closer to the left forearm; the muscles on the front of the left forearm contract slightly. \\
\cline{2-3}
& fully\_bent\_downward & The left wrist forms a right downward angle with the left forearm. The muscles on the front of the left forearm are fully contracted. \\
\cline{2-3}
& tilted\_towards\_thumb\_side & The left wrist tilts laterally to form a small angle, less than 20 degrees, moving the thumb side of the left hand closer to the left forearm. The muscles on the thumb side of the left forearm contract slightly. \\
\cline{2-3}
& tilted\_towards\_pinky\_side & The left wrist tilts laterally to form a small angle, less than 20 degrees, moving the little finger side of the left hand closer to the left forearm. The muscles on the pinky side of the left forearm contract slightly. \\
\hline

RightWrist & neutral & The right wrist extends straight, aligned with the right forearm, forming a continuous straight line from the right elbow to the right hand. The angle between the right forearm and right wrist is close to 180 degrees. \\
\cline{2-3}
& bent\_upward & The right wrist forms a right upward angle with the right forearm. The muscles on the back of the right forearm are fully contracted. \\
\cline{2-3}
& bent\_slightly\_downward & The right wrist forms a small downward angle, less than 45 degrees. The right palm moves slightly closer to the right forearm; the muscles on the front of the right forearm contract slightly. \\
\cline{2-3}
& fully\_bent\_downward & The right wrist forms a right downward angle with the right forearm. The muscles on the front of the right forearm are fully contracted. \\
\cline{2-3}
& tilted\_towards\_thumb\_side & The right wrist tilts laterally to form a small angle, less than 20 degrees, moving the thumb side of the right hand closer to the right forearm. The muscles on the thumb side of the right forearm contract slightly. \\
\cline{2-3}
& tilted\_towards\_pinky\_side & The right wrist tilts laterally to form a small angle, less than 20 degrees, moving the little finger side of the right hand closer to the right forearm. The muscles on the pinky side of the right forearm contract slightly. \\
\hline

LeftUpperLeg & neutral & The left upperleg is aligned with the body midline, standing straight with the foot pointing forward relative to the torso. \\
\cline{2-3}
& forward & The left upperleg is extended forward in front of the body, perpendicular to the torso. \\
\cline{2-3}
& side & The left upperleg is extended out to the side, perpendicular to the torso. \\
\cline{2-3}
& forward\_to\_side & The left upperleg is extended fully forward and slightly to the side, forming an approximate 45-degree angle between the forward direction and the side. \\
\cline{2-3}
& forward\_to\_midline & The left upperleg is extended fully forward and slightly towards the midline of the body. \\
\cline{2-3}
& neutral\_to\_forward & The left upperleg is lifted from the neutral position and extended forward in front of the body, in the middle of neutral and forward positions. \\
\cline{2-3}
& neutral\_to\_backward & The left upperleg is lifted from the neutral position and extended backward behind the body, in the middle of neutral and backward positions. \\
\cline{2-3}
& forward\_to\_upward & The left upperleg is slightly raised upwards from the forward position. \\
\cline{2-3}
& neutral\_to\_side & The left upperleg is lifted from the neutral position and extended to the side, in the middle of neutral and side positions. \\
\cline{2-3}
& neutral\_to\_forward\_side & The left upperleg is lifted from the neutral position and extended partly forward and slightly to the side, forming a diagonal line. \\
\cline{2-3}
& neutral\_to\_forward\_midline & The left upperleg is lifted from the neutral position and extended partly forward and slightly towards the midline of the body. \\
\cline{2-3}
& neutral\_to\_backward\_side & The left upperleg is lifted from the neutral position and extended partly backward and slightly to the side, forming a diagonal line. \\
\cline{2-3}
& neutral\_to\_backward\_midline & The left upperleg is lifted from the neutral position and extended partly backward and slightly towards the midline of the body. \\
\cline{2-3}
& forward\_to\_upward\_side & The left upperleg is extended fully forward and raised partly upwards while slightly moving towards the side. \\
\cline{2-3}
& forward\_to\_upward\_midline & The left upperleg is extended fully forward and raised partly upwards while slightly moving towards the midline of the body. \\
\hline

RightUpperLeg & neutral & The right upperleg is aligned with the body midline, standing straight with the foot pointing forward relative to the torso. \\
\cline{2-3}
& forward & The right upperleg is extended forward in front of the body, perpendicular to the torso. \\
\cline{2-3}
& side & The right upperleg is extended out to the side, perpendicular to the torso. \\
\cline{2-3}
& forward\_to\_side & The right upperleg is extended fully forward and slightly to the side, forming an approximate 45-degree angle between the forward direction and the side. \\
\cline{2-3}
& forward\_to\_midline & The right upperleg is extended fully forward and slightly towards the midline of the body. \\
\cline{2-3}
& neutral\_to\_forward & The right upperleg is lifted from the neutral position and extended forward in front of the body, in the middle of neutral and forward positions. \\
\cline{2-3}
& neutral\_to\_backward & The right upperleg is lifted from the neutral position and extended backward behind the body, in the middle of neutral and backward positions. \\
\cline{2-3}
& forward\_to\_upward & The right upperleg is slightly raised upwards from the forward position. \\
\cline{2-3}
& neutral\_to\_side & The right upperleg is lifted from the neutral position and extended to the side, in the middle of neutral and side positions. \\
\cline{2-3}
& neutral\_to\_forward\_side & The right upperleg is lifted from the neutral position and extended partly forward and slightly to the side, forming a diagonal line. \\
\cline{2-3}
& neutral\_to\_forward\_midline & The right upperleg is lifted from the neutral position and extended partly forward and slightly towards the midline of the body. \\
\cline{2-3}
& neutral\_to\_backward\_side & The right upperleg is lifted from the neutral position and extended partly backward and slightly to the side, forming a diagonal line. \\
\cline{2-3}
& neutral\_to\_backward\_midline & The right upperleg is lifted from the neutral position and extended partly backward and slightly towards the midline of the body. \\
\cline{2-3}
& forward\_to\_upward\_side & The right upperleg is extended fully forward and raised partly upwards while slightly moving towards the side. \\
\cline{2-3}
& forward\_to\_upward\_midline & The right upperleg is extended fully forward and raised partly upwards while slightly moving towards the midline of the body. \\
\hline

LeftKnee & neutral & The left leg is straight, the left knee fully extended, forming a continuous line from the left thigh to the left ankle. The angle between the left thigh and calf is close to 180 degrees. Both the front and back thigh muscles maintain a neutral length. \\
\cline{2-3}
& slightly\_bent & The left calf forms a small angle, less than 45 degrees, with the back of the left thigh. The left knee is slightly flexed; the front thigh muscles contract slightly, and the back thigh muscles lengthen slightly. \\
\cline{2-3}
& bent\_at\_90\_degrees & The left calf forms a right angle with the left back thigh. The front thigh muscles are moderately contracted, and the back thigh muscles are moderately stretched. \\
\cline{2-3}
& fully\_bent & The left calf forms a nearly zero-degree angle with the back of the left thigh, with the left heel nearly or fully touching the buttocks. The front thigh muscles are extremely contracted, and the back thigh muscles are maximally stretched. \\
\hline

RightKnee & neutral & The right leg is straight, the right knee fully extended, forming a continuous line from the right thigh to the right ankle. The angle between the right thigh and calf is close to 180 degrees. Both the front and back thigh muscles maintain a neutral length. \\
\cline{2-3}
& slightly\_bent & The right calf forms a small angle, less than 45 degrees, with the back of the right thigh. The right knee is slightly flexed; the front thigh muscles contract slightly, and the back thigh muscles lengthen slightly. \\
\cline{2-3}
& bent\_at\_90\_degrees & The right calf forms a right angle with the right back thigh. The front thigh muscles are moderately contracted, and the back thigh muscles are moderately stretched. \\
\cline{2-3}
& fully\_bent & The right calf forms a nearly zero-degree angle with the back of the right thigh, with the right heel nearly or fully touching the buttocks. The front thigh muscles are extremely contracted, and the back thigh muscles are maximally stretched. \\
\hline

LeftAnkle & neutral & The left toes point straight ahead, aligned with the left foot. The angle between the top of the left foot and the left shin is around 90 degrees. Both the front and back muscles of the left lower leg maintain a neutral length. \\
\cline{2-3}
& bent\_upward & The top of the left foot forms an acute angle, less than 90 degrees, with the left shin. The left toes point closer to the left shin; the muscles on the front of the left lower leg contract to lift the left toes. \\
\cline{2-3}
& bent\_downward & The top of the left foot forms a steep obtuse angle, around 180 degrees, with the left shin. The left toes point significantly downward; the calf muscles are fully contracted. \\
\cline{2-3}
& tilted\_inward & The sole of the left foot moves towards the midline of the body, forming a small inward angle with the left ankle. The muscles on the inside of the left lower leg contract slightly. \\
\cline{2-3}
& tilted\_outward & The sole of the left foot moves away from the midline of the body, forming a small outward angle with the left ankle. The muscles on the outside of the left lower leg contract slightly. \\
\hline

RightAnkle & neutral & The right toes point straight ahead, aligned with the right foot. The angle between the top of the right foot and the right shin is around 90 degrees. Both the front and back muscles of the right lower leg maintain a neutral length. \\
\cline{2-3}
& bent\_upward & The top of the right foot forms an acute angle, less than 90 degrees, with the right shin. The right toes point closer to the right shin; the muscles on the front of the right lower leg contract to lift the right toes. \\
\cline{2-3}
& bent\_downward & The top of the right foot forms a steep obtuse angle, around 180 degrees, with the right shin. The right toes point significantly downward; the calf muscles are fully contracted. \\
\cline{2-3}
& tilted\_inward & The sole of the right foot moves towards the midline of the body, forming a small inward angle with the right ankle. The muscles on the inside of the right lower leg contract slightly. \\
\cline{2-3}
& tilted\_outward & The sole of the right foot moves away from the midline of the body, forming a small outward angle with the right ankle. The muscles on the outside of the right lower leg contract slightly. \\
\hline

LeftToes & neutral & The left toes point straight ahead, aligned with the left foot. The left toes and the left foot form a straight line from the base to the tips, making a right angle. \\
\cline{2-3}
& curled\_up & The left toes lift upwards, forming small upward angles between the left toe tips and the left foot. The muscles on the top of the left foot and left toes contract to lift the left toes. \\
\cline{2-3}
& curled\_down & The left toes curl downward, forming small downward angles with the sole of the left foot. The muscles on the bottom of the left foot and left toes contract to curl the left toes. \\
\hline

RightToes & neutral & The right toes point straight ahead, aligned with the right foot. The right toes and the right foot form a straight line from the base to the tips, making a right angle. \\
\cline{2-3}
& curled\_up & The right toes lift upwards, forming small upward angles between the right toe tips and the right foot. The muscles on the top of the right foot and right toes contract to lift the right toes. \\
\cline{2-3}
& curled\_down & The right toes curl downward, forming small downward angles with the sole of the right foot. The muscles on the bottom of the right foot and right toes contract to curl the right toes. \\
\hline

\end{longtable}

\twocolumn
\normalsize

\onecolumn
\subsection{Details of Preset Body Parts and Predefined Positions}
\label{appendix:mapping_rules}

\begin{scriptsize}
\begin{longtable}{|c|c|c|}
\caption{Preset body parts with their predefined positions and the corresponding coordinates of relevant SMPL joints (Rules).} \label{tab:body_part_positions_details}\\
\hline
\textbf{Body Part} & \textbf{Position} & \textbf{Coordinates of the Relevant SMPL Joints (Rules)} \\
\hline
\endfirsthead

\hline
\textbf{Body Part} & \textbf{Position} & \textbf{Coordinates of the Relevant SMPL Joints (Rules)} \\
\hline
\endhead

\hline
\endfoot

\hline
\endlastfoot

\multirow{13}{*}{Head} & neutral & \{m\_avg\_Head: [0.0, 0.0, 0.0], m\_avg\_Neck: [0.0, 0.0, 0.0]\} \\
& tilted\_up\_slightly & \{m\_avg\_Head: [-30.0, 0.0, 0.0], m\_avg\_Neck: [0.0, 0.0, 0.0]\} \\
& tilted\_up\_fully & \{m\_avg\_Head: [-45.0, 0.0, 0.0], m\_avg\_Neck: [-10.0, 0.0, 0.0]\} \\
& tilted\_down\_slightly & \{m\_avg\_Head: [30.0, 0.0, 0.0], m\_avg\_Neck: [0.0, 0.0, 0.0]\} \\
& tilted\_down\_fully & \{m\_avg\_Head: [45.0, 0.0, 0.0], m\_avg\_Neck: [30.0, 0.0, 0.0]\} \\
& turned\_left\_slightly & \{m\_avg\_Head: [0.0, -45.0, 0.0], m\_avg\_Neck: [0.0, 0.0, 0.0]\} \\
& turned\_left\_fully & \{m\_avg\_Head: [0.0, -60.0, 0.0], m\_avg\_Neck: [0.0, -30.0, 0.0]\} \\
& turned\_right\_slightly & \{m\_avg\_Head: [0.0, 45.0, 0.0], m\_avg\_Neck: [0.0, 0.0, 0.0]\} \\
& turned\_right\_fully & \{m\_avg\_Head: [0.0, 60.0, 0.0], m\_avg\_Neck: [0.0, 30.0, 0.0]\} \\
& tilted\_left\_slightly & \{m\_avg\_Head: [0.0, 0.0, 20.0], m\_avg\_Neck: [0.0, 0.0, 0.0]\} \\
& tilted\_left\_fully & \{m\_avg\_Head: [0.0, 0.0, 45.0], m\_avg\_Neck: [0.0, 0.0, 20.0]\} \\
& tilted\_right\_slightly & \{m\_avg\_Head: [0.0, 0.0, -20.0], m\_avg\_Neck: [0.0, 0.0, 0.0]\} \\
& tilted\_right\_fully & \{m\_avg\_Head: [0.0, 0.0, -45.0], m\_avg\_Neck: [0.0, 0.0, -20.0]\} \\
\hline

\multirow{12}{*}{Torso} & neutral & \{m\_avg\_Spine1: [0.0, 0.0, 0.0], m\_avg\_Spine2: [0.0, 0.0, 0.0], m\_avg\_Spine3: [0.0, 0.0, 0.0]\} \\
& twisted\_left\_slightly & \{m\_avg\_Spine1: [0.0, -20.0, 0.0], m\_avg\_Spine2: [0.0, -10.0, 0.0], m\_avg\_Spine3: [0.0, -5.0, 0.0]\} \\
& twisted\_left\_fully & \{m\_avg\_Spine1: [0.0, -30.0, 0.0], m\_avg\_Spine2: [0.0, -20.0, 0.0], m\_avg\_Spine3: [0.0, -10.0, 0.0]\} \\
& twisted\_right\_slightly & \{m\_avg\_Spine1: [0.0, 20.0, 0.0], m\_avg\_Spine2: [0.0, 10.0, 0.0], m\_avg\_Spine3: [0.0, 5.0, 0.0]\} \\
& twisted\_right\_fully & \{m\_avg\_Spine1: [0.0, 30.0, 0.0], m\_avg\_Spine2: [0.0, 20.0, 0.0], m\_avg\_Spine3: [0.0, 0.0, 10.0]\} \\
& bent\_forward\_slightly & \{m\_avg\_Spine1: [30.0, 0.0, 0.0], m\_avg\_Spine2: [5.0, 0.0, 0.0], m\_avg\_Spine3: [5.0, 0.0, 0.0]\} \\
& bent\_forward\_fully & \{m\_avg\_Spine1: [60.0, 0.0, 0.0], m\_avg\_Spine2: [10.0, 0.0, 0.0], m\_avg\_Spine3: [10.0, 0.0, 0.0]\} \\
& bent\_backward & \{m\_avg\_Spine1: [-20.0, 0.0, 0.0], m\_avg\_Spine2: [0.0, 0.0, 0.0], m\_avg\_Spine3: [0.0, 0.0, 0.0]\} \\
& tilted\_left\_slightly & \{m\_avg\_Spine1: [0.0, 0.0, 10.0], m\_avg\_Spine2: [0.0, 0.0, 5.0], m\_avg\_Spine3: [0.0, 0.0, 5.0]\} \\
& tilted\_left\_fully & \{m\_avg\_Spine1: [0.0, 0.0, 20.0], m\_avg\_Spine2: [0.0, 0.0, 10.0], m\_avg\_Spine3: [0.0, 0.0, 10.0]\} \\
& tilted\_right\_slightly & \{m\_avg\_Spine1: [0.0, 0.0, -10.0], m\_avg\_Spine2: [0.0, 0.0, -5.0], m\_avg\_Spine3: [0.0, 0.0, -5.0]\} \\
& tilted\_right\_fully & \{m\_avg\_Spine1: [0.0, 0.0, -20.0], m\_avg\_Spine2: [0.0, 0.0, -10.0], m\_avg\_Spine3: [0.0, 0.0, -10.0]\} \\
\hline

\multirow{21}{*}{LeftUpperArm} & neutral & \{m\_avg\_L\_Collar: [0.0, 0.0, 20.0], m\_avg\_L\_Shoulder: [0.0, 0.0, 70.0]\} \\
& forward\_elbowpit\_inward & \{m\_avg\_L\_Collar: [0.0, 0.0, 0.0], m\_avg\_L\_Shoulder: [0.0, 90.0, 0.0]\} \\
& forward\_elbowpit\_upward & \{m\_avg\_L\_Collar: [0.0, 0.0, 20.0], m\_avg\_L\_Shoulder: [-70.0, 90.0, 0.0]\} \\
& upward & \{m\_avg\_L\_Collar: [0.0, 0.0, -45.0], m\_avg\_L\_Shoulder: [0.0, 0.0, -40.0]\} \\
& side\_elbowpit\_forward & \{m\_avg\_L\_Collar: [0.0, 0.0, 0.0], m\_avg\_L\_Shoulder: [0.0, 0.0, 0.0]\} \\
& side\_elbowpit\_upward & \{m\_avg\_L\_Collar: [-45.0, 0.0, 0.0], m\_avg\_L\_Shoulder: [-45.0, 0.0, 0.0]\} \\
& neutral\_to\_forward & \{m\_avg\_L\_Collar: [0.0, 0.0, 10.0], m\_avg\_L\_Shoulder: [-60.0, 20.0, 70.0]\} \\
& forward\_to\_upward & \{m\_avg\_L\_Collar: [0.0, 0.0, -10.0], m\_avg\_L\_Shoulder: [0.0, 90.0, -35.0]\} \\
& neutral\_to\_back & \{m\_avg\_L\_Collar: [0.0, 0.0, 10.0], m\_avg\_L\_Shoulder: [10.0, 90.0, 120.0]\} \\
& forward\_to\_midline & \{m\_avg\_L\_Collar: [0.0, 10.0, 0.0], m\_avg\_L\_Shoulder: [0.0, 120.0, 0.0]\} \\
& forward\_to\_side & \{m\_avg\_L\_Collar: [0.0, 10.0, 0.0], m\_avg\_L\_Shoulder: [0.0, 30.0, 0.0]\} \\
& side\_to\_back & \{m\_avg\_L\_Collar: [0.0, -10.0, 0.0], m\_avg\_L\_Shoulder: [0.0, -30.0, 0.0]\} \\
& neutral\_to\_side & \{m\_avg\_L\_Collar: [0.0, 0.0, 10.0], m\_avg\_L\_Shoulder: [0.0, 0.0, 45.0]\} \\
& upward\_to\_side & \{m\_avg\_L\_Collar: [0.0, 0.0, -10.0], m\_avg\_L\_Shoulder: [0.0, 0.0, -45.0]\} \\
& forward\_to\_upward\_side & \{m\_avg\_L\_Collar: [0.0, 0.0, -10.0], m\_avg\_L\_Shoulder: [0.0, 45.0, -30.0]\} \\
& forward\_to\_upward\_midline & \{m\_avg\_L\_Collar: [0.0, 15.0, -10.0], m\_avg\_L\_Shoulder: [0.0, 110.0, -30.0]\} \\
& side\_to\_upward\_back & \{m\_avg\_L\_Collar: [0.0, -15.0, -10.0], m\_avg\_L\_Shoulder: [0.0, -30.0, -30.0]\} \\
& neutral\_to\_forward\_midline & \{m\_avg\_L\_Collar: [0.0, 0.0, 10.0], m\_avg\_L\_Shoulder: [-45.0, 0.0, 100.0]\} \\
& neutral\_to\_forward\_side & \{m\_avg\_L\_Collar: [0.0, 0.0, 10.0], m\_avg\_L\_Shoulder: [45.0, 90.0, 45.0]\} \\
& neutral\_to\_backward\_side & \{m\_avg\_L\_Collar: [20.0, 0.0, 10.0], m\_avg\_L\_Shoulder: [45.0, 90.0, 100.0]\} \\
\hline

\multirow{21}{*}{RightUpperArm} & neutral & \{m\_avg\_R\_Collar: [0.0, 0.0, -20.0], m\_avg\_R\_Shoulder: [0.0, 0.0, -70.0]\} \\
& forward\_elbowpit\_inward & \{m\_avg\_R\_Collar: [0.0, 0.0, 0.0], m\_avg\_R\_Shoulder: [0.0, -90.0, 0.0]\} \\
& forward\_elbowpit\_upward & \{m\_avg\_R\_Collar: [0.0, 0.0, -20.0], m\_avg\_R\_Shoulder: [-70.0, -90.0, 0.0]\} \\
& upward & \{m\_avg\_R\_Collar: [0.0, 0.0, 45.0], m\_avg\_R\_Shoulder: [0.0, 0.0, 40.0]\} \\
& side\_elbowpit\_forward & \{m\_avg\_R\_Collar: [0.0, 0.0, 0.0], m\_avg\_R\_Shoulder: [0.0, 0.0, 0.0]\} \\
& side\_elbowpit\_upward & \{m\_avg\_R\_Collar: [-45.0, 0.0, 0.0], m\_avg\_R\_Shoulder: [-45.0, 0.0, 0.0]\} \\
& neutral\_to\_forward & \{m\_avg\_R\_Collar: [0.0, 0.0, -10.0], m\_avg\_R\_Shoulder: [-60.0, -20.0, -70.0]\} \\
& forward\_to\_upward & \{m\_avg\_R\_Collar: [0.0, 0.0, 10.0], m\_avg\_R\_Shoulder: [0.0, -90.0, 35.0]\} \\
& neutral\_to\_back & \{m\_avg\_R\_Collar: [0.0, 0.0, -10.0], m\_avg\_R\_Shoulder: [10.0, -90.0, -120.0]\} \\
& forward\_to\_midline & \{m\_avg\_R\_Collar: [0.0, -10.0, 0.0], m\_avg\_R\_Shoulder: [0.0, -120.0, 0.0]\} \\
& forward\_to\_side & \{m\_avg\_R\_Collar: [0.0, -10.0, 0.0], m\_avg\_R\_Shoulder: [0.0, -30.0, 0.0]\} \\
& side\_to\_back & \{m\_avg\_R\_Collar: [0.0, 10.0, 0.0], m\_avg\_R\_Shoulder: [0.0, 30.0, 0.0]\} \\
& neutral\_to\_side & \{m\_avg\_R\_Collar: [0.0, 0.0, -10.0], m\_avg\_R\_Shoulder: [0.0, 0.0, -45.0]\} \\
& upward\_to\_side & \{m\_avg\_R\_Collar: [0.0, 0.0, 10.0], m\_avg\_R\_Shoulder: [0.0, 0.0, 45.0]\} \\
& forward\_to\_upward\_side & \{m\_avg\_R\_Collar: [0.0, 0.0, 10.0], m\_avg\_R\_Shoulder: [0.0, -45.0, 30.0]\} \\
& forward\_to\_upward\_midline & \{m\_avg\_R\_Collar: [0.0, -15.0, 10.0], m\_avg\_R\_Shoulder: [0.0, -110.0, 30.0]\} \\
& side\_to\_upward\_back & \{m\_avg\_R\_Collar: [0.0, 15.0, 10.0], m\_avg\_R\_Shoulder: [0.0, 30.0, 30.0]\} \\
& neutral\_to\_forward\_midline & \{m\_avg\_R\_Collar: [0.0, 0.0, -10.0], m\_avg\_R\_Shoulder: [-45.0, 0.0, -100.0]\} \\
& neutral\_to\_forward\_side & \{m\_avg\_R\_Collar: [0.0, 0.0, -10.0], m\_avg\_R\_Shoulder: [45.0, -90.0, -45.0]\} \\
& neutral\_to\_backward\_side & \{m\_avg\_R\_Collar: [20.0, 0.0, -10.0], m\_avg\_R\_Shoulder: [45.0, -90.0, -100.0]\} \\
\hline

\multirow{4}{*}{LeftElbow} & neutral & \{m\_avg\_L\_Elbow: [0.0, 0.0, 0.0]\} \\
& slightly\_bent\_in & \{m\_avg\_L\_Elbow: [0.0, 45.0, 0.0]\} \\
& bent\_in\_90\_degrees & \{m\_avg\_L\_Elbow: [0.0, 90.0, 0.0]\} \\
& fully\_bent & \{m\_avg\_L\_Elbow: [0.0, 135.0, 0.0]\} \\
\hline

\multirow{4}{*}{RightElbow} & neutral & \{m\_avg\_R\_Elbow: [0.0, 0.0, 0.0]\} \\
& slightly\_bent\_in & \{m\_avg\_R\_Elbow: [0.0, -45.0, 0.0]\} \\
& bent\_in\_90\_degrees & \{m\_avg\_R\_Elbow: [0.0, -90.0, 0.0]\} \\
& fully\_bent & \{m\_avg\_R\_Elbow: [0.0, -135.0, 0.0]\} \\
\hline

\newpage

\multirow{6}{*}{LeftWrist} & neutral & \{m\_avg\_L\_Wrist: [0.0, 0.0, 0.0]\} \\
& bent\_upward & \{m\_avg\_L\_Wrist: [0.0, 0.0, -45.0]\} \\
& bent\_slightly\_downward & \{m\_avg\_L\_Wrist: [0.0, 0.0, 45.0]\} \\
& fully\_bent\_downward & \{m\_avg\_L\_Wrist: [0.0, 0.0, 80.0]\} \\
& tilted\_towards\_thumb\_side & \{m\_avg\_L\_Wrist: [0.0, 30.0, 0.0]\} \\
& tilted\_towards\_pinky\_side & \{m\_avg\_L\_Wrist: [0.0, -30.0, 0.0]\} \\
\hline

\multirow{6}{*}{RightWrist} & neutral & \{m\_avg\_R\_Wrist: [0.0, 0.0, 0.0]\} \\
& bent\_upward & \{m\_avg\_R\_Wrist: [0.0, 0.0, 45.0]\} \\
& bent\_slightly\_downward & \{m\_avg\_R\_Wrist: [0.0, 0.0, -45.0]\} \\
& fully\_bent\_downward & \{m\_avg\_R\_Wrist: [0.0, 0.0, -80.0]\} \\
& tilted\_towards\_thumb\_side & \{m\_avg\_R\_Wrist: [0.0, -30.0, 0.0]\} \\
& tilted\_towards\_pinky\_side & \{m\_avg\_R\_Wrist: [0.0, 30.0, 0.0]\} \\
\hline

\multirow{15}{*}{LeftUpperLeg} & neutral & \{m\_avg\_L\_Hip: [0.0, 0.0, 0.0]\} \\
& forward & \{m\_avg\_L\_Hip: [-90.0, 0.0, 0.0]\} \\
& side & \{m\_avg\_L\_Hip: [0.0, 0.0, -80.0]\} \\
& forward\_to\_side & \{m\_avg\_L\_Hip: [-90.0, 0.0, -45.0]\} \\
& forward\_to\_midline & \{m\_avg\_L\_Hip: [-90.0, 0.0, 45.0]\} \\
& neutral\_to\_forward & \{m\_avg\_L\_Hip: [-45.0, 0.0, 0.0]\} \\
& neutral\_to\_backward & \{m\_avg\_L\_Hip: [45.0, 0.0, 0.0]\} \\
& forward\_to\_upward & \{m\_avg\_L\_Hip: [-120.0, 0.0, 0.0]\} \\
& neutral\_to\_side & \{m\_avg\_L\_Hip: [0.0, 0.0, -45.0]\} \\
& neutral\_to\_forward\_side & \{m\_avg\_L\_Hip: [-45.0, 0.0, -45.0]\} \\
& neutral\_to\_forward\_midline & \{m\_avg\_L\_Hip: [-45.0, 0.0, 45.0]\} \\
& neutral\_to\_backward\_side & \{m\_avg\_L\_Hip: [45.0, 0.0, -45.0]\} \\
& neutral\_to\_backward\_midline & \{m\_avg\_L\_Hip: [45.0, 0.0, 45.0]\} \\
& forward\_to\_upward\_side & \{m\_avg\_L\_Hip: [-120.0, 0.0, -45.0]\} \\
& forward\_to\_upward\_midline & \{m\_avg\_L\_Hip: [-120.0, 0.0, 45.0]\} \\
\hline

\multirow{15}{*}{RightUpperLeg} & neutral & \{m\_avg\_R\_Hip: [0.0, 0.0, 0.0]\} \\
& forward & \{m\_avg\_R\_Hip: [-90.0, 0.0, 0.0]\} \\
& side & \{m\_avg\_R\_Hip: [0.0, 0.0, 80.0]\} \\
& forward\_to\_side & \{m\_avg\_R\_Hip: [-90.0, 0.0, 45.0]\} \\
& forward\_to\_midline & \{m\_avg\_R\_Hip: [-90.0, 0.0, -45.0]\} \\
& neutral\_to\_forward & \{m\_avg\_R\_Hip: [-45.0, 0.0, 0.0]\} \\
& neutral\_to\_backward & \{m\_avg\_R\_Hip: [45.0, 0.0, 0.0]\} \\
& forward\_to\_upward & \{m\_avg\_R\_Hip: [-120.0, 0.0, 0.0]\} \\
& neutral\_to\_side & \{m\_avg\_R\_Hip: [0.0, 0.0, 45.0]\} \\
& neutral\_to\_forward\_side & \{m\_avg\_R\_Hip: [-45.0, 0.0, 45.0]\} \\
& neutral\_to\_forward\_midline & \{m\_avg\_R\_Hip: [-45.0, 0.0, -45.0]\} \\
& neutral\_to\_backward\_side & \{m\_avg\_R\_Hip: [45.0, 0.0, 45.0]\} \\
& neutral\_to\_backward\_midline & \{m\_avg\_R\_Hip: [45.0, 0.0, -45.0]\} \\
& forward\_to\_upward\_side & \{m\_avg\_R\_Hip: [-120.0, 0.0, 45.0]\} \\
& forward\_to\_upward\_midline & \{m\_avg\_R\_Hip: [-120.0, 0.0, -45.0]\} \\
\hline

\multirow{4}{*}{LeftKnee} & neutral & \{m\_avg\_L\_Knee: [0.0, 0.0, 0.0]\} \\
& slightly\_bent & \{m\_avg\_L\_Knee: [45.0, 0.0, 0.0]\} \\
& bent\_at\_90\_degrees & \{m\_avg\_L\_Knee: [90.0, 0.0, 0.0]\} \\
& fully\_bent & \{m\_avg\_L\_Knee: [135.0, 0.0, 0.0]\} \\
\hline

\multirow{4}{*}{RightKnee} & neutral & \{m\_avg\_R\_Knee: [0.0, 0.0, 0.0]\} \\
& slightly\_bent & \{m\_avg\_R\_Knee: [45.0, 0.0, 0.0]\} \\
& bent\_at\_90\_degrees & \{m\_avg\_R\_Knee: [90.0, 0.0, 0.0]\} \\
& fully\_bent & \{m\_avg\_R\_Knee: [135.0, 0.0, 0.0]\} \\
\hline

\multirow{5}{*}{LeftAnkle} & neutral & \{m\_avg\_L\_Ankle: [0.0, 0.0, 0.0]\} \\
& bent\_upward & \{m\_avg\_L\_Ankle: [-20.0, 0.0, 0.0]\} \\
& bent\_downward & \{m\_avg\_L\_Ankle: [45.0, 0.0, 0.0]\} \\
& tilted\_inward & \{m\_avg\_L\_Ankle: [0.0, 0.0, 30.0]\} \\
& tilted\_outward & \{m\_avg\_L\_Ankle: [0.0, 0.0, -10.0]\} \\
\hline

\multirow{5}{*}{RightAnkle} & neutral & \{m\_avg\_R\_Ankle: [0.0, 0.0, 0.0]\} \\
& bent\_upward & \{m\_avg\_R\_Ankle: [-20.0, 0.0, 0.0]\} \\
& bent\_downward & \{m\_avg\_R\_Ankle: [45.0, 0.0, 0.0]\} \\
& tilted\_inward & \{m\_avg\_R\_Ankle: [0.0, 0.0, -30.0]\} \\
& tilted\_outward & \{m\_avg\_R\_Ankle: [0.0, 0.0, 10.0]\} \\
\hline

\multirow{3}{*}{LeftToes} & neutral & \{m\_avg\_L\_Foot: [0.0, 0.0, 0.0]\} \\
& curled\_up & \{m\_avg\_L\_Foot: [-30.0, 0.0, 0.0]\} \\
& curled\_down & \{m\_avg\_L\_Foot: [30.0, 0.0, 0.0]\} \\
\hline

\multirow{3}{*}{RightToes} & neutral & \{m\_avg\_R\_Foot: [0.0, 0.0, 0.0]\} \\
& curled\_up & \{m\_avg\_R\_Foot: [-30.0, 0.0, 0.0]\} \\
& curled\_down & \{m\_avg\_R\_Foot: [30.0, 0.0, 0.0]\} \\
\hline

\end{longtable}
\end{scriptsize}

\twocolumn

\newpage
\subsection{Details of Human Evaluation}
We conduct human evaluation as part of our research methodology. The nine human evaluators are graduate students, technical staff or researchers working on artificial intelligence at the same university, and participate voluntarily with above-average-wage compensation. Tasks are designed to be safe and unbiased, with clear instructions and reasonable time commitments. Participants are informed of the study's purpose and withdrawal rights. Prior to evaluation, we explicitly explain how evaluators' responses and feedback will be used in our research, including potential publication in academic venues, and obtain written consent from all evaluators. No personal data are collected. The protocol is approved by our institution's ethics review board and adheres to human subject research guidelines.

\onecolumn
\subsubsection{High-level Planning}
\label{appendix:human_evaluation_details_high_level_planning}

\begin{table}[h]
  \centering
  \begin{tabular}{lp{0.8\textwidth}}
    \toprule
    \textbf{Score} & \textbf{Judgement} \\
    \midrule
    \makecell{5} & The high-level plan follows the motion instruction well and specifies all important details. \\
    \makecell{4} & The high-level plan generally follows the motion instruction (80--90\%), but contains some minor errors. \\
    \makecell{3} & The high-level plan follows the motion instruction 50--70\% and contains one or two major errors that prevent it from achieving the goal \\
    \makecell{2} & The high-level plan shows some sign of following the motion instruction (20--40\%), but contains so many errors that it is far from the goal state \\
    \makecell{1} & The high-level plan does not follow the motion instruction at all \\
    \bottomrule
  \end{tabular}
  \caption{Rubric for High-level Plan Score (HPS). The instruction to the annotators: \texttt{Given the motion instruction, to what extent do you think the high-level plan of body part movements appropriately specifies the instructed motion? Score from 1 (poor) to 5 (excellent). Possible shortcomings usually include wrong and incomplete action descriptions.}}
  \label{tab:high_level_plan_score}
\end{table}

\begin{figure}[htbp]
    \centering
    \includegraphics[width=0.7\columnwidth]{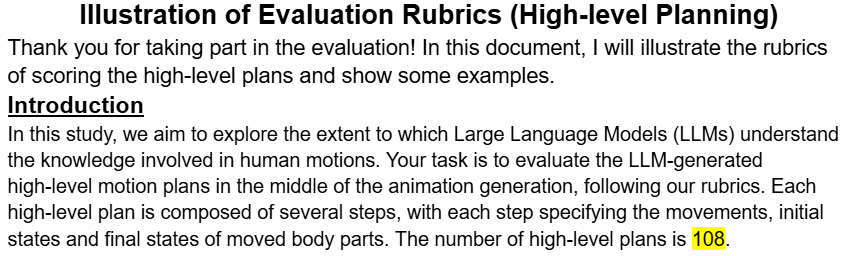}
    \caption{The illustrative document for each human evaluator to read. It is followed by evaluation rubrics and four examples covering different scores.}
    \label{fig:illustration_hp}
\end{figure}

\newpage
\subsubsection{Complete Generation}
\label{appendix:human_evaluation_details_complete_generation}

\begin{table}[h]
  \centering
  \begin{tabular}{lp{0.8\textwidth}}
    \toprule
    \textbf{Score} & \textbf{Judgement} \\
    \midrule
    \makecell{5} & The animation follows the motion instruction well without redundant or strange movements \\
    \makecell{4} & The animation generally follows the motion instruction (70--90\%), but contains some minor errors (e.g., redundant or strange movements) \\
    \makecell{3} & The animation follows the motion instruction 40--60\% and contains one or two major errors that prevent it from achieving the goal \\
    \makecell{2} & The animation shows some sign of following the motion instruction (20--30\%), but contains so many errors that it is far from the goal state \\
    \makecell{1} & The animation does not follow the motion instruction at all \\
    \bottomrule
  \end{tabular}
  \caption{Rubric for Whole Body Score (WBS). The instruction to the annotators: \texttt{Given the motion instruction, to what extent do you think the animation is appropriately following the instructed motion? Score from 1 (poor) to 5 (excellent).}}
  \label{tab:whole_body_score}
\end{table}

\begin{table}[h]
  \centering
  \begin{tabular}{lp{0.8\textwidth}}
    \toprule
    \textbf{Label} & \textbf{Judgement} \\
    \midrule
    Good & The body part follows the given motion instruction well \\
    Partially Good & The body part follows the given motion instruction partially, but has errors \\
    Bad & The body part does not follow the given motion instruction at all. Or, the body part is not absolutely necessary to this motion but is ridiculously moved \\
    Not Relevant & The body part is not absolutely necessary to this motion and is not ridiculously moved \\
    \bottomrule
  \end{tabular}
  \caption{Rubric for Body Part Quality (BPQ). The instruction to the annotators: \texttt{For each body part, choose one from “Good”, “Partially Good”, “Bad” and “Not Relevant”}.}
  \label{tab:body_part_quality}
\end{table}
\twocolumn

\begin{figure}[htbp]
    \centering
    \includegraphics[width=\columnwidth]{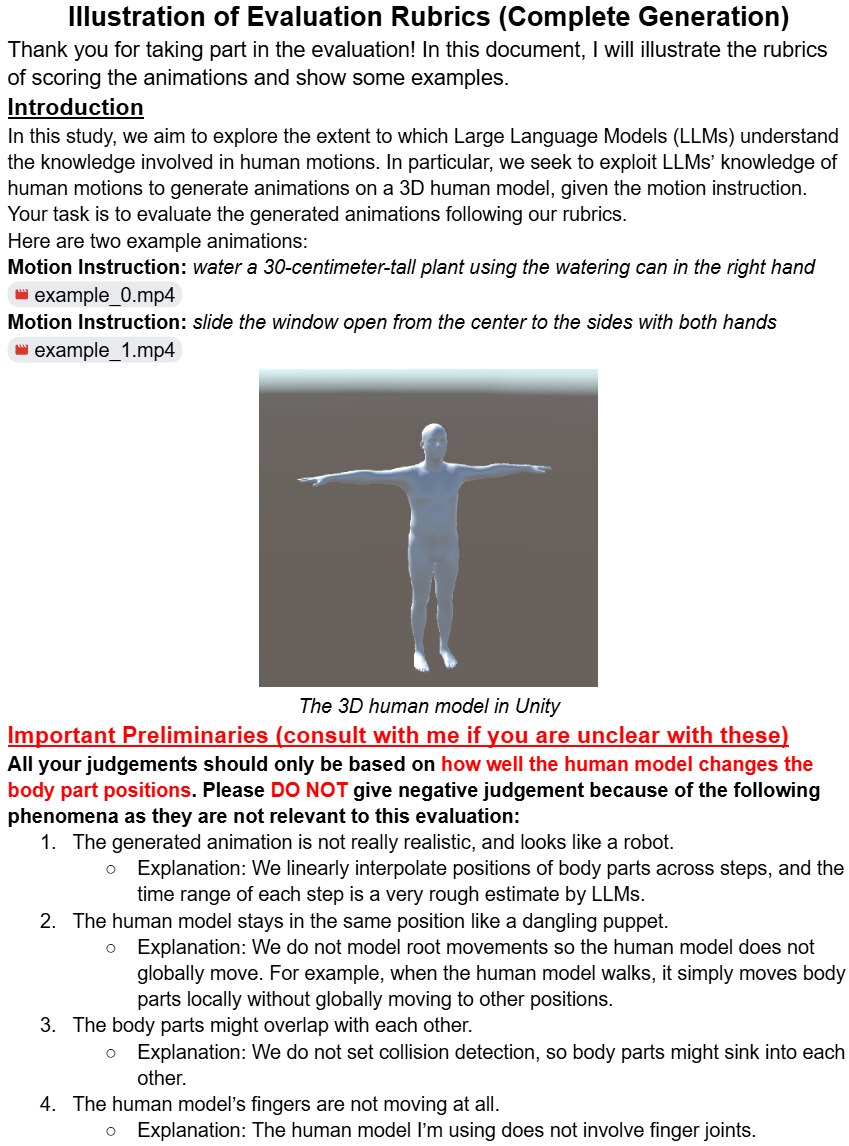}
    \caption{The illustrative document for each human evaluator to read. This page of the document is followed by evaluation rubrics and 6 examples covering different WBS and BPQ.}
    \label{fig:illustration_animation}
\end{figure}

\begin{figure}[htbp]
    \centering
    \includegraphics[width=\columnwidth]{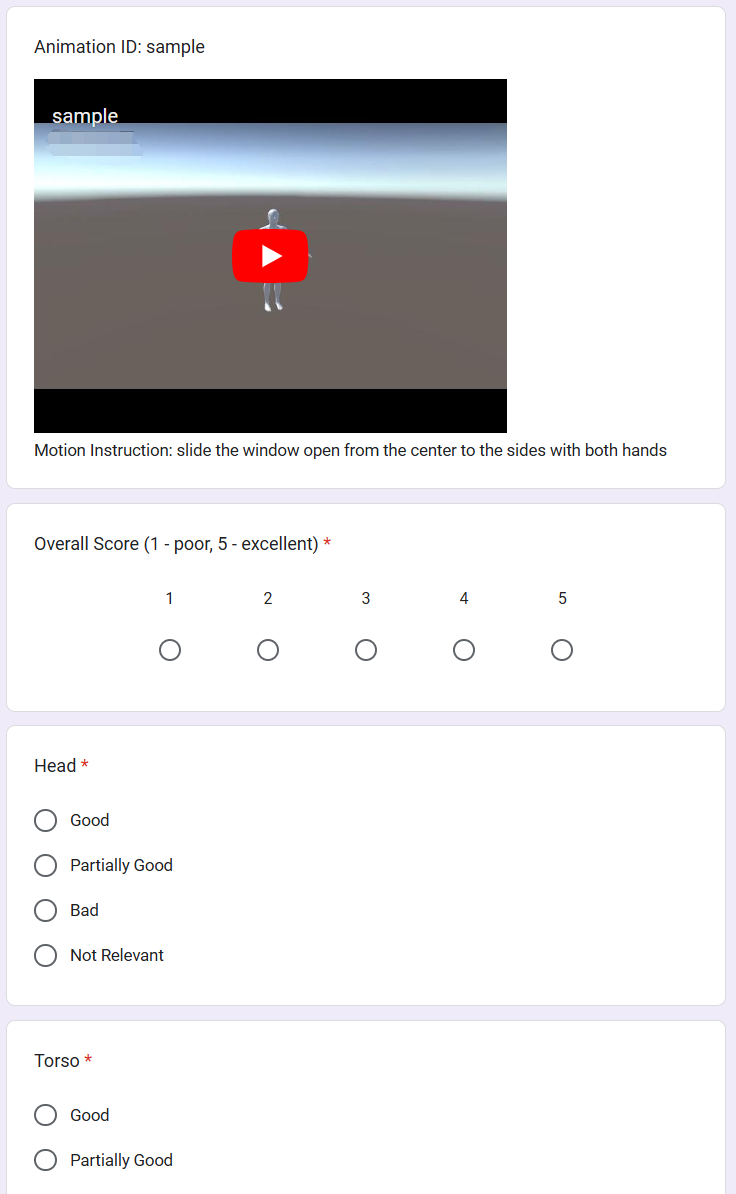}
    \caption{Sample form in human evaluation of the animations from the complete generation. At the header of each form, we provide a link to the illustrative document shown in Figure \ref{fig:illustration_animation}. We ensure that each annotator evaluates a balanced mix of animations from different settings including LLMs and querying strategies.}
    \label{fig:sample_form}
\end{figure}

\newpage
\subsection{Motion Instructions}
\label{appendix:motion_instructions}

We create twenty motion instructions for the main experiments. They are freely available for research purposes only. Researchers working on language and human motion are welcome to access and use these instructions for future investigation, which aligns with our intended use case. Derivative works or applications should remain within research contexts. The motion instructions are carefully curated to be neutral in nature, containing no personally identifiable information or controversial content.

\begin{table}[htbp]
    \centering
    \small  
    \begin{tabular}{p{0.15\columnwidth}p{0.75\columnwidth}}
    \toprule
    \textbf{\makecell{ID}} & \textbf{Motion Instruction} \\
    \midrule
    \makecell{1} & Slide the window open from the center to the sides with both hands. \\\hline
    \makecell{2} & Water a 30-centimeter-tall plant using the watering can in the right hand. \\\hline
    \makecell{3} & Look down to check the time of the watch on the left wrist. \\\hline
    \makecell{4} & Pat a 30-centimeter-tall dog in front of you on the head with the right hand. \\\hline
    \makecell{5} & Lean back fully and toss the ball into the air at a 45-degree angle using both hands. \\\hline
    \makecell{6} & Wipe down the 1-meter-high table in front of you with a cloth in the left hand. \\\hline
    \makecell{7} & Hold the glass with the left hand and pour the juice with the right hand. \\\hline
    \makecell{8} & Put a book on the 2-meter-high shelf with both hands. \\\hline
    \makecell{9} & Lift a 20-centimeter-high box from the ground to the table on your left with both hands. \\\hline
    \makecell{10} & Swing the golf club from right to left. \\\hline
    \makecell{11} & Close the 2-meter-high store shutter door from top to bottom. \\\hline
    \makecell{12} & Squat to pick up litter by the right foot with the right hand. \\\hline
    \makecell{13} & Lift the right shoe with both hands and put it on in the air. \\\hline
    \makecell{14} & Perform a left-leg high side kick in Karate. \\\hline
    \makecell{15} & Kneel in a traditional Japanese bow. \\\hline
    \makecell{16} & Roll out a yoga mat on the ground. \\\hline
    \makecell{17} & Crouch to check a car tyre. \\\hline
    \makecell{18} & Arch the back 60 degrees to relieve tension in the lower back muscles with two hands on the waist. \\\hline
    \makecell{19} & Bend to the left to reach for an item by the left foot without moving or bending the left leg. \\\hline
    \makecell{20} & Walk through while ducking under a low-hanging branch. \\
    \bottomrule
    \end{tabular}
    \caption{The motion instructions for main experiments. Each instruction specifies necessary contextual elements to eliminate ambiguity while testing LLMs' ability to infer implicit motion details. To show the potential of application, we devise each instruction to be related to a practical scene, while deliberately avoiding common game animations to focus on challenging scenarios requiring genuine motion understanding.}
    \label{tab:motion_instructions}
\end{table}

\newpage
\subsection{Hyperparameters and Computational Costs}
\label{appendix:model_running_details}

\begin{table}[h]
  \centering
  \small
    \begin{tabular}{cc}
      \toprule
      \textbf{Hyperparameter} &
      \textbf{Value} \\
      \midrule
      \texttt{temperature} & 1 \\
      \texttt{max\_tokens} & 4095 \\
      \texttt{timeout} & 60 \\
      \texttt{max\_retries} & 3 \\
      \bottomrule
    \end{tabular}
  \caption{Hyperparameter configuration for LLMs}
  \label{tab:hyperparameters}
\end{table}

The average cost per animation generation ranges from \$2.70 (Claude 3.5 Sonnet) to \$0.07 (GPT-4o-mini), with GPT-4o at \$1.20 and GPT-3.5-turbo at \$0.25. The open-source Llama-3.1-70B requires 1.5-2 hours on one 48GB NVIDIA RTX A6000.

\newpage

\subsection{Self-reflection Analysis}
\label{appendix:reflection}

\begin{figure}[htbp]
    \centering
    \includegraphics[width=\columnwidth]{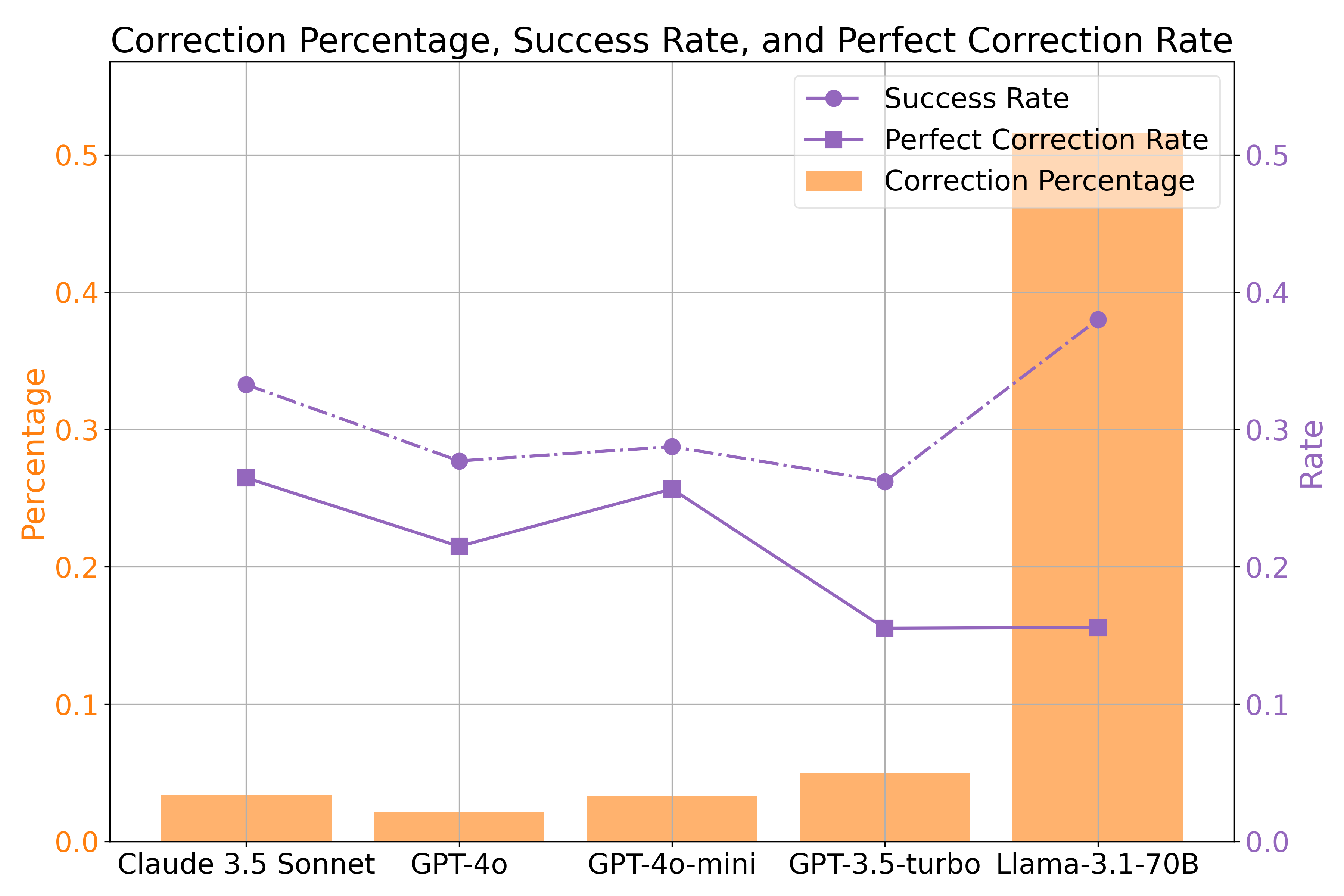}
    \caption{Reflection Statistics.}
    \label{fig:auto_model_comparison_reflection}
\end{figure}

We report the reflection statistics in Figure \ref{fig:auto_model_comparison_reflection}. To investigate the effect of self-reflection, we calculate the percentage of corrections among all body parts of all steps (\textbf{Correction Percentage}), the percentage of body parts with finally correct positions among all corrected body parts (\textbf{Success Rate}), and the percentage of corrected body parts where the last selected position is correct and previous selected positions are wrong, among all corrected body parts (\textbf{Perfect Correction Rate}).

\hspace*{1em} Llama-3.1-70B has an extraordinarily high correction percentage, while other models seldom correct after reflection. While Llama-3.1-70B's Success Rate is the highest, its Perfect Correction Rate is the lowest relative to Success Rate. This phenomenon might be attributed to Llama-3.1-70B's lack of proper instruction following capabilities, i.e., it tends to reflect and correct when asked, no matter whether the selected position should be corrected.

\newpage
\subsection{Complementary Agreement Information}

\begin{figure}[htbp]
    \centering
    \includegraphics[width=\columnwidth]{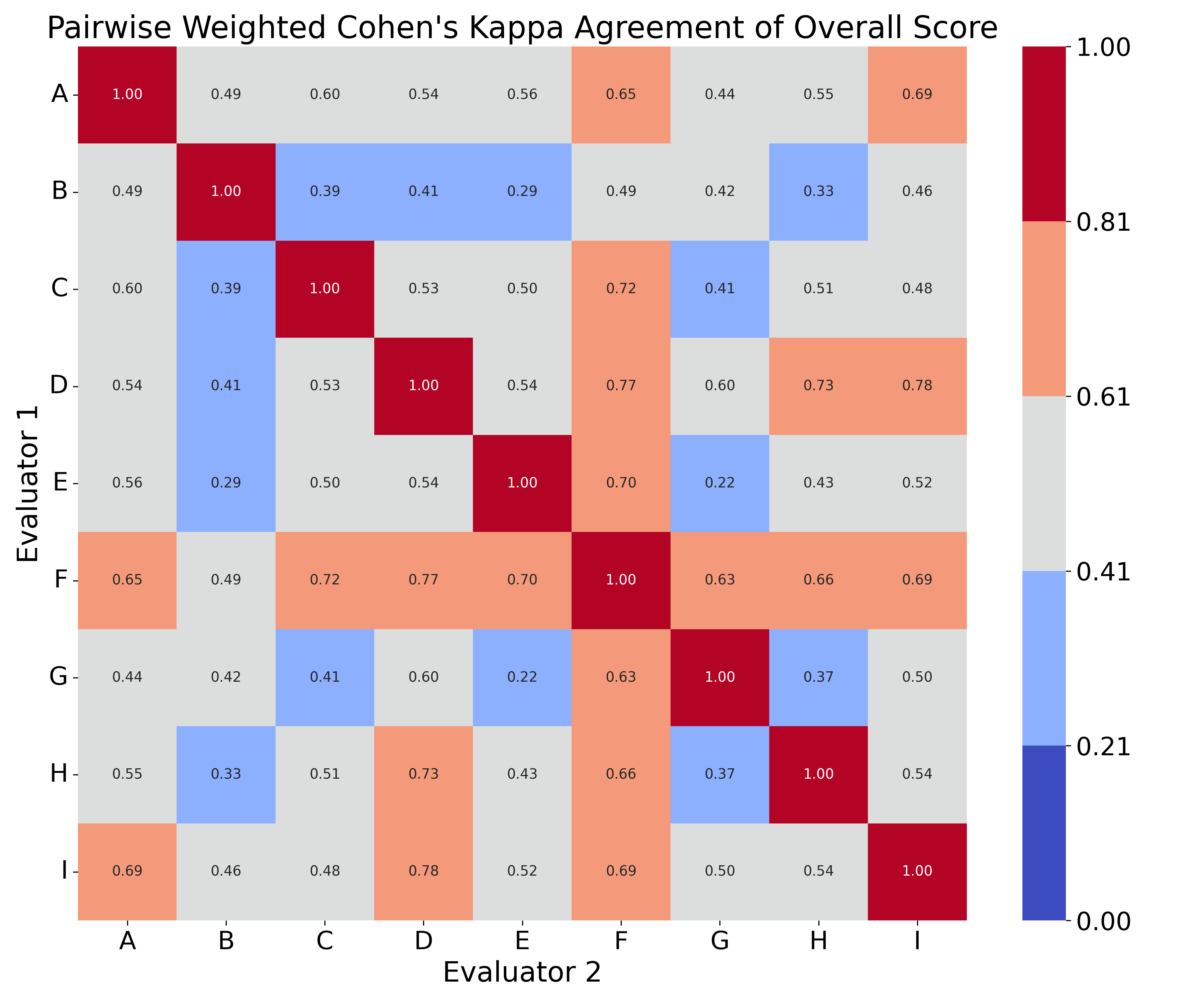}
    \caption{Pairwise weighted kappa scores of nine evaluators for WBS agreement. Based on the interpretation of \citet{kappa_interpretation}, the ranges 0.21-0.40, 0.41-0.60, and 0.61-0.80 respectively correspond to fair, moderate, and substantial levels of agreement.}
    \label{fig:agreement_overall_scores}
\end{figure}

\begin{table}[h]
  \centering
  \begin{tabular}{cc}
    \toprule
    \textbf{Body Part} & \textbf{APA} \\
    \midrule
    \texttt{Head} & 0.510 / 0.758 \\
    \texttt{Torso} & 0.536 / 0.667 \\
    \texttt{Left Arm} & 0.569 / 0.558 \\
    \texttt{Right Arm} & 0.553 / 0.550 \\
    \texttt{Left Leg} & 0.637 / 0.792 \\
    \texttt{Right Leg} & 0.638 / 0.717 \\
    \bottomrule
  \end{tabular}
  \caption{Average pairwise agreement (APA) on BPQ for each body part. Each score pair contains the agreement among human annotators (left), and between human and Gemini 2.5 Pro judgement (right). Human agreement is computed as the mean percentage of matching categories across annotator pairs; human--Gemini agreement reflects the percentage of matched categories between the majority votes of humans and Gemini 2.5 Pro.}
  \label{tab:agreement_body_parts}
\end{table}

\onecolumn
\newpage
\subsection{Statistical Measures}
\label{appendix:statistical_measures}

\subsubsection{HPS Scores}
\label{appendix:statistical_measures_hps}

\begin{table}[htbp]
\centering
\resizebox{0.9\textwidth}{!}{%
\begin{tabular}{c|c|c|c|c|c}
\toprule
\textbf{Motion ID} & \textbf{Claude 3.5 Sonnet} & \textbf{GPT-4o} & \textbf{GPT-4o-mini} & \textbf{GPT-3.5-turbo} & \textbf{Llama-3.1-70B} \\
\midrule
1 & 5.00 (0.00, 0.00) & 5.00 (0.00, 0.00) & 3.67 (0.47, 0.22) & 2.33 (0.47, 0.22) & 5.00 (0.00, 0.00) \\
2 & 4.67 (0.47, 0.22) & 4.33 (0.47, 0.22) & 4.33 (0.47, 0.22) & 4.33 (0.47, 0.22) & 4.67 (0.47, 0.22) \\
3 & 5.00 (0.00, 0.00) & 3.33 (0.47, 0.22) & 5.00 (0.00, 0.00) & 3.33 (0.47, 0.22) & 3.00 (0.00, 0.00) \\
4 & 4.33 (0.47, 0.22) & 4.67 (0.47, 0.22) & 5.00 (0.00, 0.00) & 2.33 (0.47, 0.22) & 2.33 (0.47, 0.22) \\
5 & 4.33 (0.47, 0.22) & 5.00 (0.00, 0.00) & 5.00 (0.00, 0.00) & 3.33 (0.47, 0.22) & 4.67 (0.47, 0.22) \\
6 & 5.00 (0.00, 0.00) & 5.00 (0.00, 0.00) & 5.00 (0.00, 0.00) & 3.00 (0.00, 0.00) & 4.33 (0.47, 0.22) \\
7 & 4.33 (0.47, 0.22) & 4.67 (0.47, 0.22) & 4.67 (0.47, 0.22) & 4.33 (0.47, 0.22) & 4.67 (0.47, 0.22) \\
\bottomrule
\end{tabular}
}
\caption{Human-scored HPS (piece\_by\_piece) by motion ID for each LLM. Values represent the mean score with associated standard deviation and variance.}
\end{table}

\begin{table}[htbp]
\centering
\resizebox{0.9\textwidth}{!}{%
\begin{tabular}{c|c|c|c|c|c}
\toprule
\textbf{Motion ID} & \textbf{Claude 3.5 Sonnet} & \textbf{GPT-4o} & \textbf{GPT-4o-mini} & \textbf{GPT-3.5-turbo} & \textbf{Llama-3.1-70B} \\
\midrule
1 & 5.00 (0.00, 0.00) & 5.00 (0.00, 0.00) & 2.67 (0.47, 0.22) & 3.00 (0.00, 0.00) & 4.67 (0.47, 0.22) \\
2 & 5.00 (0.00, 0.00) & 4.33 (0.47, 0.22) & 5.00 (0.00, 0.00) & 4.00 (0.00, 0.00) & 4.00 (0.00, 0.00) \\
3 & 5.00 (0.00, 0.00) & 3.00 (0.00, 0.00) & 4.00 (0.00, 0.00) & 3.00 (0.00, 0.00) & 3.00 (0.00, 0.00) \\
4 & 4.00 (0.00, 0.00) & 4.00 (0.00, 0.00) & 5.00 (0.00, 0.00) & 3.00 (0.00, 0.00) & 3.67 (0.47, 0.22) \\
5 & 4.00 (0.00, 0.00) & 4.00 (0.00, 0.00) & 4.00 (0.00, 0.00) & 3.00 (0.00, 0.00) & 3.00 (0.00, 0.00) \\
6 & 4.00 (0.00, 0.00) & 5.00 (0.00, 0.00) & 4.00 (0.00, 0.00) & 3.00 (0.00, 0.00) & 4.00 (0.00, 0.00) \\
7 & 5.00 (0.00, 0.00) & 5.00 (0.00, 0.00) & 4.00 (0.00, 0.00) & 4.33 (0.47, 0.22) & 4.00 (0.00, 0.00) \\
\bottomrule
\end{tabular}
}
\caption{GPT-4.1-scored HPS (piece\_by\_piece) by motion ID for each LLM. Values represent the mean score with associated standard deviation and variance.}
\end{table}

\begin{table}[htbp]
\centering
\resizebox{\textwidth}{!}{%
\begin{tabular}{c|c|c|c|c}
\toprule
\textbf{Motion ID} & \textbf{Claude 3.5 Sonnet} & \textbf{GPT-4o} & \textbf{GPT-4o-mini} & \textbf{GPT-3.5-turbo} \\
\midrule
1 & 5.00 (0.00, 0.00) & 4.67 (0.47, 0.22) & 5.00 (0.00, 0.00) & 4.67 (0.47, 0.22) \\
2 & 3.67 (0.47, 0.22) & 4.67 (0.47, 0.22) & 3.33 (0.47, 0.22) & 4.33 (0.47, 0.22) \\
3 & 4.67 (0.47, 0.22) & 3.00 (0.00, 0.00) & 3.00 (0.00, 0.00) & 3.33 (0.47, 0.22) \\
4 & 5.00 (0.00, 0.00) & 5.00 (0.00, 0.00) & 5.00 (0.00, 0.00) & 4.33 (0.47, 0.22) \\
5 & 5.00 (0.00, 0.00) & 5.00 (0.00, 0.00) & 4.67 (0.47, 0.22) & 4.33 (0.47, 0.22) \\
6 & 3.67 (0.47, 0.22) & 5.00 (0.00, 0.00) & 3.67 (0.47, 0.22) & 3.67 (0.47, 0.22) \\
7 & 3.33 (0.47, 0.22) & 4.67 (0.47, 0.22) & 3.33 (0.47, 0.22) & 3.67 (0.47, 0.22) \\
8 & 3.67 (0.47, 0.22) & 4.67 (0.47, 0.22) & 4.67 (0.47, 0.22) & 3.00 (0.00, 0.00) \\
9 & 4.67 (0.47, 0.22) & 4.67 (0.47, 0.22) & 2.00 (0.00, 0.00) & 3.33 (0.47, 0.22) \\
10 & 3.33 (0.47, 0.22) & 4.67 (0.47, 0.22) & 3.00 (0.00, 0.00) & 2.00 (0.82, 0.67) \\
11 & 4.33 (0.47, 0.22) & 3.33 (0.47, 0.22) & 3.67 (0.47, 0.22) & 4.67 (0.47, 0.22) \\
12 & 4.67 (0.47, 0.22) & 5.00 (0.00, 0.00) & 5.00 (0.00, 0.00) & 2.33 (0.47, 0.22) \\
13 & 4.67 (0.47, 0.22) & 4.33 (0.47, 0.22) & 4.00 (0.00, 0.00) & 2.67 (0.47, 0.22) \\
14 & 4.67 (0.47, 0.22) & 3.33 (0.47, 0.22) & 4.67 (0.47, 0.22) & 3.00 (0.00, 0.00) \\
15 & 5.00 (0.00, 0.00) & 4.67 (0.47, 0.22) & 2.67 (0.47, 0.22) & 2.67 (0.47, 0.22) \\
16 & 4.33 (0.47, 0.22) & 5.00 (0.00, 0.00) & 2.67 (0.47, 0.22) & 1.00 (0.00, 0.00) \\
17 & 4.33 (0.47, 0.22) & 5.00 (0.00, 0.00) & 4.67 (0.47, 0.22) & 4.67 (0.47, 0.22) \\
18 & 5.00 (0.00, 0.00) & 5.00 (0.00, 0.00) & 4.67 (0.47, 0.22) & 3.33 (0.47, 0.22) \\
19 & 4.33 (0.47, 0.22) & 4.67 (0.47, 0.22) & 4.33 (0.47, 0.22) & 3.67 (0.47, 0.22) \\
20 & 5.00 (0.00, 0.00) & 4.67 (0.47, 0.22) & 4.67 (0.47, 0.22) & 2.00 (0.00, 0.00) \\
\bottomrule
\end{tabular}
}
\caption{Human-scored HPS (in\_one\_go) by motion ID for each LLM. Values represent the mean score with associated standard deviation and variance.}
\end{table}

\begin{table}[htbp]
\centering
\resizebox{\textwidth}{!}{%
\begin{tabular}{c|c|c|c|c}
\toprule
\textbf{Motion ID} & \textbf{Claude 3.5 Sonnet} & \textbf{GPT-4o} & \textbf{GPT-4o-mini} & \textbf{GPT-3.5-turbo} \\
\midrule
1 & 5.00 (0.00, 0.00) & 4.00 (0.00, 0.00) & 4.67 (0.47, 0.22) & 5.00 (0.00, 0.00) \\
2 & 4.00 (0.00, 0.00) & 4.67 (0.47, 0.22) & 4.00 (0.00, 0.00) & 4.00 (0.00, 0.00) \\
3 & 5.00 (0.00, 0.00) & 2.00 (0.00, 0.00) & 3.00 (0.00, 0.00) & 2.33 (0.47, 0.22) \\
4 & 5.00 (0.00, 0.00) & 4.67 (0.47, 0.22) & 5.00 (0.00, 0.00) & 4.00 (0.00, 0.00) \\
5 & 4.00 (0.00, 0.00) & 4.00 (0.00, 0.00) & 3.67 (0.47, 0.22) & 3.00 (0.00, 0.00) \\
6 & 4.00 (0.00, 0.00) & 4.00 (0.00, 0.00) & 3.33 (0.47, 0.22) & 3.00 (0.00, 0.00) \\
7 & 4.33 (0.47, 0.22) & 4.67 (0.47, 0.22) & 3.00 (0.00, 0.00) & 3.00 (0.00, 0.00) \\
8 & 4.00 (0.00, 0.00) & 3.67 (0.47, 0.22) & 3.00 (0.00, 0.00) & 2.00 (0.00, 0.00) \\
9 & 5.00 (0.00, 0.00) & 4.67 (0.47, 0.22) & 2.67 (0.47, 0.22) & 3.33 (0.47, 0.22) \\
10 & 4.00 (0.00, 0.00) & 4.00 (0.00, 0.00) & 4.00 (0.00, 0.00) & 2.00 (0.00, 0.00) \\
11 & 5.00 (0.00, 0.00) & 5.00 (0.00, 0.00) & 3.00 (0.00, 0.00) & 4.33 (0.47, 0.22) \\
12 & 4.00 (0.00, 0.00) & 4.00 (0.00, 0.00) & 4.00 (0.00, 0.00) & 4.00 (0.00, 0.00) \\
13 & 5.00 (0.00, 0.00) & 4.67 (0.47, 0.22) & 2.00 (0.00, 0.00) & 2.67 (0.47, 0.22) \\
14 & 5.00 (0.00, 0.00) & 3.67 (0.47, 0.22) & 4.00 (0.00, 0.00) & 3.00 (0.00, 0.00) \\
15 & 4.67 (0.47, 0.22) & 4.33 (0.47, 0.22) & 4.00 (0.00, 0.00) & 4.00 (0.00, 0.00) \\
16 & 4.33 (0.47, 0.22) & 5.00 (0.00, 0.00) & 4.00 (0.00, 0.00) & 1.00 (0.00, 0.00) \\
17 & 4.33 (0.47, 0.22) & 5.00 (0.00, 0.00) & 4.33 (0.47, 0.22) & 4.00 (0.00, 0.00) \\
18 & 5.00 (0.00, 0.00) & 5.00 (0.00, 0.00) & 5.00 (0.00, 0.00) & 2.00 (0.00, 0.00) \\
19 & 5.00 (0.00, 0.00) & 4.67 (0.47, 0.22) & 4.00 (0.00, 0.00) & 4.00 (0.00, 0.00) \\
20 & 4.00 (0.00, 0.00) & 4.00 (0.00, 0.00) & 4.00 (0.00, 0.00) & 2.00 (0.00, 0.00) \\
\bottomrule
\end{tabular}
}
\caption{GPT-4.1-scored HPS (in\_one\_go) by motion ID for each LLM. Values represent the mean score with associated standard deviation and variance.}
\end{table}

\newpage
\subsubsection{BPPA Scores}
\label{appendix:statistical_measures_bppa}

\begin{table}[htbp]
\centering
\resizebox{\textwidth}{!}{%
\begin{tabular}{c|c|c|c|c|c}
\toprule
\textbf{Motion ID} & \textbf{Claude 3.5 Sonnet} & \textbf{GPT-4o} & \textbf{GPT-4o-mini} & \textbf{GPT-3.5-turbo} & \textbf{Llama-3.1-70B} \\
\midrule
1 & 0.7812 (0.03, 0.00) & 0.7188 (0.00, 0.00) & 0.7188 (0.03, 0.00) & 0.7031 (0.02, 0.00) & 0.5781 (0.02, 0.00) \\
2 & 0.9000 (0.00, 0.00) & 0.9125 (0.01, 0.00) & 0.9062 (0.02, 0.00) & 0.8250 (0.01, 0.00) & 0.6750 (0.04, 0.00) \\
3 & 0.9531 (0.02, 0.00) & 0.9062 (0.03, 0.00) & 0.8750 (0.00, 0.00) & 0.9531 (0.02, 0.00) & 0.6719 (0.08, 0.01) \\
4 & 0.7396 (0.03, 0.00) & 0.5312 (0.03, 0.00) & 0.5938 (0.07, 0.01) & 0.6562 (0.01, 0.00) & 0.4896 (0.01, 0.00) \\
5 & 0.7812 (0.00, 0.00) & 0.7812 (0.00, 0.00) & 0.6328 (0.01, 0.00) & 0.6797 (0.05, 0.00) & 0.5079 (0.09, 0.01) \\
6 & 0.8938 (0.01, 0.00) & 0.8625 (0.01, 0.00) & 0.7875 (0.01, 0.00) & 0.7875 (0.00, 0.00) & 0.7375 (0.01, 0.00) \\
7 & 0.7916 (0.02, 0.00) & 0.8021 (0.01, 0.00) & 0.7396 (0.01, 0.00) & 0.8646 (0.01, 0.00) & 0.7812 (0.01, 0.00) \\
\bottomrule
\end{tabular}
}
\caption{BPPA (hierarchical) scores by motion ID for each LLM. Values represent the mean score with associated standard deviation and variance.}
\label{tab:statistical_measures_bppa_hierarchical}
\end{table}

\begin{table}[htbp]
\centering
\resizebox{\textwidth}{!}{%
\begin{tabular}{c|c|c|c|c|c}
\toprule
\textbf{Motion ID} & \textbf{Claude 3.5 Sonnet} & \textbf{GPT-4o} & \textbf{GPT-4o-mini} & \textbf{GPT-3.5-turbo} & \textbf{Llama-3.1-70B} \\
\midrule
1  & 0.8438 (0.03, 0.00) & 0.7812 (0.00, 0.00) & 0.7656 (0.08, 0.01) & 0.7344 (0.02, 0.00) & 0.6250 (0.02, 0.00) \\
2  & 0.8750 (0.01, 0.00) & 0.8562 (0.01, 0.00) & 0.8625 (0.00, 0.00) & 0.7937 (0.03, 0.00) & 0.5813 (0.08, 0.01) \\
3  & 0.9062 (0.03, 0.00) & 0.9062 (0.03, 0.00) & 0.9062 (0.00, 0.00) & 0.8125 (0.09, 0.01) & 0.7500 (0.03, 0.00) \\
4  & 0.7188 (0.01, 0.00) & 0.6562 (0.01, 0.00) & 0.6041 (0.02, 0.00) & 0.5938 (0.01, 0.00) & 0.5625 (0.06, 0.00) \\
5  & 0.6719 (0.02, 0.00) & 0.7734 (0.01, 0.00) & 0.6406 (0.00, 0.00) & 0.6641 (0.01, 0.00) & 0.4766 (0.07, 0.00) \\
6  & 0.8562 (0.02, 0.00) & 0.8687 (0.02, 0.00) & 0.8438 (0.02, 0.00) & 0.8188 (0.02, 0.00) & 0.6687 (0.07, 0.00) \\
7  & 0.7812 (0.01, 0.00) & 0.8438 (0.01, 0.00) & 0.7396 (0.01, 0.00) & 0.8021 (0.01, 0.00) & 0.6562 (0.01, 0.00) \\
\bottomrule
\end{tabular}
}
\caption{BPPA (one\_by\_one) scores by motion ID for each LLM. Values represent the mean score with associated standard deviation and variance.}
\label{tab:statistical_measures_bppa_obo}
\end{table}

\begin{table}[htbp]
\centering
\resizebox{\textwidth}{!}{%
\begin{tabular}{c|c|c|c|c|c}
\toprule
\textbf{Motion ID} & \textbf{Claude 3.5 Sonnet} & \textbf{GPT-4o} & \textbf{GPT-4o-mini} & \textbf{GPT-3.5-turbo} & \textbf{Llama-3.1-70B} \\
\midrule
1  & 0.7656 (0.02, 0.00) & 0.7578 (0.01, 0.00) & 0.8125 (0.00, 0.00) & 0.2188 (0.03, 0.00) & 0.6250 (0.08, 0.01) \\
2  & 0.8062 (0.02, 0.00) & 0.9062 (0.01, 0.00) & 0.9000 (0.04, 0.00) & 0.3187 (0.07, 0.00) & 0.6125 (0.05, 0.00) \\
3  & 0.8906 (0.02, 0.00) & 0.7969 (0.02, 0.00) & 0.8594 (0.02, 0.00) & 0.2344 (0.02, 0.00) & 0.7344 (0.02, 0.00) \\
4  & 0.6562 (0.01, 0.00) & 0.4896 (0.01, 0.00) & 0.6146 (0.05, 0.00) & 0.2083 (0.02, 0.00) & 0.4584 (0.04, 0.00) \\
5  & 0.7109 (0.04, 0.00) & 0.7188 (0.00, 0.00) & 0.6875 (0.02, 0.00) & 0.1641 (0.09, 0.01) & 0.4062 (0.03, 0.00) \\
6  & 0.8187 (0.01, 0.00) & 0.7688 (0.01, 0.00) & 0.8000 (0.06, 0.00) & 0.2062 (0.03, 0.00) & 0.4750 (0.04, 0.00) \\
7  & 0.8021 (0.03, 0.00) & 0.8021 (0.01, 0.00) & 0.6979 (0.01, 0.00) & 0.1354 (0.05, 0.00) & 0.5104 (0.07, 0.01) \\
\bottomrule
\end{tabular}
}
\caption{BPPA (all) scores by motion ID for each LLM. Values represent the mean score with associated standard deviation and variance.}
\label{tab:statistical_measures_bppa_all}
\end{table}

\newpage

\subsubsection{WBS Scores}
\label{appendix:statistical_measures_wbs}

\begin{table}[htbp]
\centering
\resizebox{\textwidth}{!}{%
\begin{tabular}{c|c|c|c|c|c|c}
\toprule
\textbf{Motion ID} & \textbf{Claude 3.5 Sonnet} & \textbf{GPT-4o} & \textbf{GPT-4o-mini} & \textbf{GPT-3.5-turbo} & \textbf{Llama-3.1-70B} & \textbf{Oracle Annotation} \\
\midrule
1  & 1.80 (0.75, 0.56) & 4.40 (0.49, 0.24) & 1.80 (0.75, 0.56) & 1.80 (0.75, 0.56) & 2.20 (0.75, 0.56) & 5.00 (0.00, 0.00) \\
2  & 4.80 (0.40, 0.16) & 4.20 (0.40, 0.16) & 2.80 (0.75, 0.56) & 1.80 (0.40, 0.16) & 1.60 (0.80, 0.64) & 5.00 (0.00, 0.00) \\
3  & 4.00 (0.00, 0.00) & 1.00 (0.00, 0.00) & 2.60 (0.49, 0.24) & 1.00 (0.00, 0.00) & 2.40 (1.02, 1.04) & 5.00 (0.00, 0.00) \\
4  & 2.60 (1.50, 2.24) & 3.20 (0.40, 0.16) & 3.60 (1.02, 1.04) & 1.00 (0.00, 0.00) & 1.40 (0.49, 0.24) & 4.60 (0.49, 0.24) \\
5  & 3.40 (1.20, 1.44) & 3.20 (0.40, 0.16) & 3.40 (0.80, 0.64) & 2.60 (0.49, 0.24) & 1.60 (0.80, 0.64) & 4.80 (0.40, 0.16) \\
6  & 4.20 (0.40, 0.16) & 2.60 (0.80, 0.64) & 2.80 (1.17, 1.36) & 3.20 (0.40, 0.16) & 1.60 (0.49, 0.24) & 4.80 (0.40, 0.16) \\
7  & 3.20 (0.75, 0.56) & 2.80 (0.75, 0.56) & 2.00 (0.89, 0.80) & 2.40 (0.49, 0.24) & 3.20 (0.98, 0.96) & 5.00 (0.00, 0.00) \\
\bottomrule
\end{tabular}
}
\caption{Human-scored WBS by motion ID for each LLM. Values represent the mean score with associated standard deviation and variance.}
\end{table}

\begin{table}[htbp]
\centering
\resizebox{\textwidth}{!}{%
\begin{tabular}{c|c|c|c|c|c|c}
\toprule
\textbf{Motion ID} & \textbf{Claude 3.5 Sonnet} & \textbf{GPT-4o} & \textbf{GPT-4o-mini} & \textbf{GPT-3.5-turbo} & \textbf{Llama-3.1-70B} & \textbf{Oracle Annotation} \\
\midrule
1 & 4.20 (1.17, 1.36) & 5.00 (0.00, 0.00) & 3.20 (0.98, 0.96) & 2.20 (0.75, 0.56) & 1.80 (0.75, 0.56) & 5.00 (0.00, 0.00) \\
2 & 4.20 (0.75, 0.56) & 5.00 (0.00, 0.00) & 1.00 (0.00, 0.00) & 1.00 (0.00, 0.00) & 1.20 (0.40, 0.16) & 5.00 (0.00, 0.00) \\
3 & 4.20 (0.98, 0.96) & 1.00 (0.00, 0.00) & 2.60 (1.20, 1.44) & 1.80 (1.60, 2.56) & 1.40 (0.49, 0.24) & 3.40 (1.96, 3.84) \\
4 & 4.00 (1.26, 1.60) & 1.60 (0.49, 0.24) & 2.60 (0.80, 0.64) & 1.00 (0.00, 0.00) & 1.00 (0.00, 0.00) & 4.60 (0.80, 0.64) \\
5 & 2.60 (0.49, 0.24) & 4.60 (0.49, 0.24) & 2.60 (1.02, 1.04) & 2.20 (0.40, 0.16) & 1.80 (0.75, 0.56) & 1.20 (0.40, 0.16) \\
6 & 4.00 (0.89, 0.80) & 1.20 (0.40, 0.16) & 2.20 (0.40, 0.16) & 4.40 (0.80, 0.64) & 2.60 (0.49, 0.24) & 1.40 (0.49, 0.24) \\
7 & 4.60 (0.49, 0.24) & 2.60 (0.49, 0.24) & 1.00 (0.00, 0.00) & 4.00 (1.55, 2.40) & 4.60 (0.49, 0.24) & 4.60 (0.80, 0.64) \\
\bottomrule
\end{tabular}
}
\caption{Gemini 2.5 Pro-scored WBS by motion ID for each LLM. Values represent the mean score with associated standard deviation and variance.}
\end{table}

\onecolumn
\newpage
\subsection{High-level Planning Errors}
\label{appendix:high_level_planning_errors}

\captionof{figure}{Example of wrong action descriptions, generated by GPT-3.5-turbo. In this example, the plan only specifies the process of lifting the foot and grabbing the shoe, deviating from the intention of lifting the shoe and then putting it on.}
\label{fig:high_level_plan_case_wrong}

\lstset{style=MyListingStyle_0}
\begin{lstlisting}[escapeinside={(*@}{@*)}]
![motion instruction]!
lift the right shoe with both hands and put it on in the air

![high-level plan]!
[
  {
    @"step_number"@: 1,
    @"time_range"@: [0, 3],
    @"movement"@: "In Step 1, the relevant body parts and their single-directional movements are:\n1. Right foot: Lifted towards the hands\n2. Hands: Reach down and grasp the right shoe",
    @"initial_state"@: "In Step 1, the initial states of relevant body parts are:\n1. Right foot: Initially on the ground\n2. Hands: Initially hanging beside the body",
    @"final_state"@: "In Step 1, the final states of relevant body parts are:\n1. Right foot: Lifted in the air\n2. Hands: Holding the right shoe"
  }
]
\end{lstlisting}

\captionof{figure}{Example of incomplete action descriptions, generated by GPT-4o. In this example, the plan does not specify the action of the left arm in order for the eyes to see the watch.}
\label{fig:high_level_plan_case_incomplete}

\lstset{style=MyListingStyle_0}
\begin{lstlisting}[escapeinside={(*@}{@*)}]
![motion instruction]!
look down to check the time of the watch on the left wrist

![high-level plan]!
[
  {
    @"step_number"@: 1,
    @"time_range"@: [0, 2],
    @"movement"@: "In Step 1, the relevant movements are:\n\n1. The head tilts downward.\n2. The eyes move downward to focus on the wrist.",
    @"initial_state"@: "The initial states of the relevant body parts in Step 1 are:\n\n1. The head is upright and facing forward.\n2. The eyes are looking straight ahead.\n3. The left arm is hanging naturally beside the body.\n4. The left wrist is aligned with the left arm, facing inward.",
    @"final_state"@: "The final states of the relevant body parts in Step 1 are:\n\n1. The head is tilted downward.\n2. The eyes are directed downward, focusing on the left wrist.\n3. The left arm remains hanging naturally beside the body.\n4. The left wrist remains aligned with the left arm, facing inward."
  }
]
\end{lstlisting}

\twocolumn

\newpage
\subsection{Complexity Analysis}

\begin{figure}[htbp]
    \centering
    \includegraphics[width=\columnwidth]{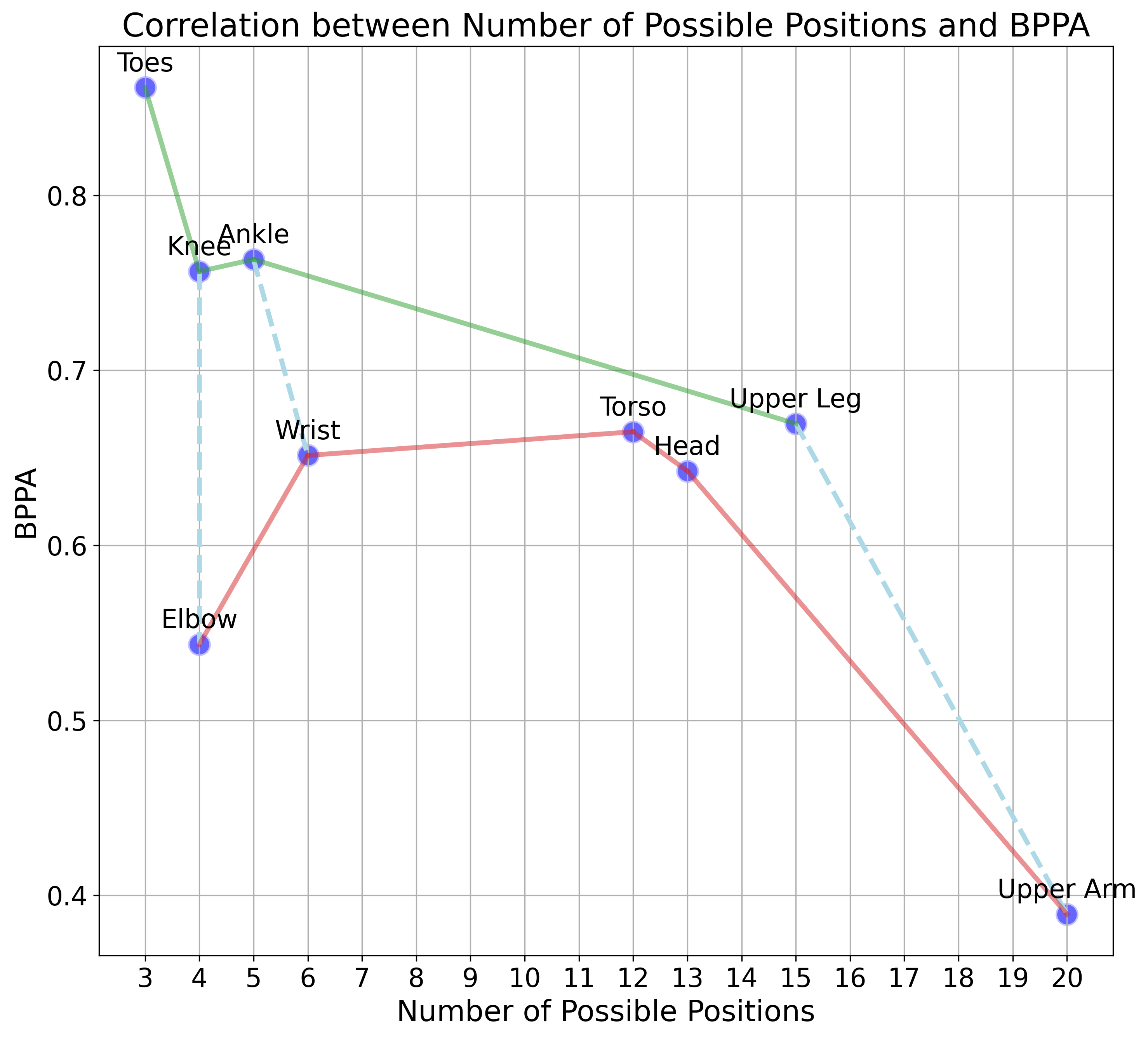}
    \caption{The body-part-wise correlation between the number of possible positions and BPPA. We average the BPPA for paired body parts. Comparison of the lower body performance (green line) and upper body performance (red line) demonstrates that LLMs achieve higher accuracy for lower body parts versus their upper body counterparts.}
    \label{fig:low_level_planning_body_part_DoF_accuracy_correlation}
\end{figure}

\begin{figure}[htbp]
    \centering
    \includegraphics[width=\columnwidth]{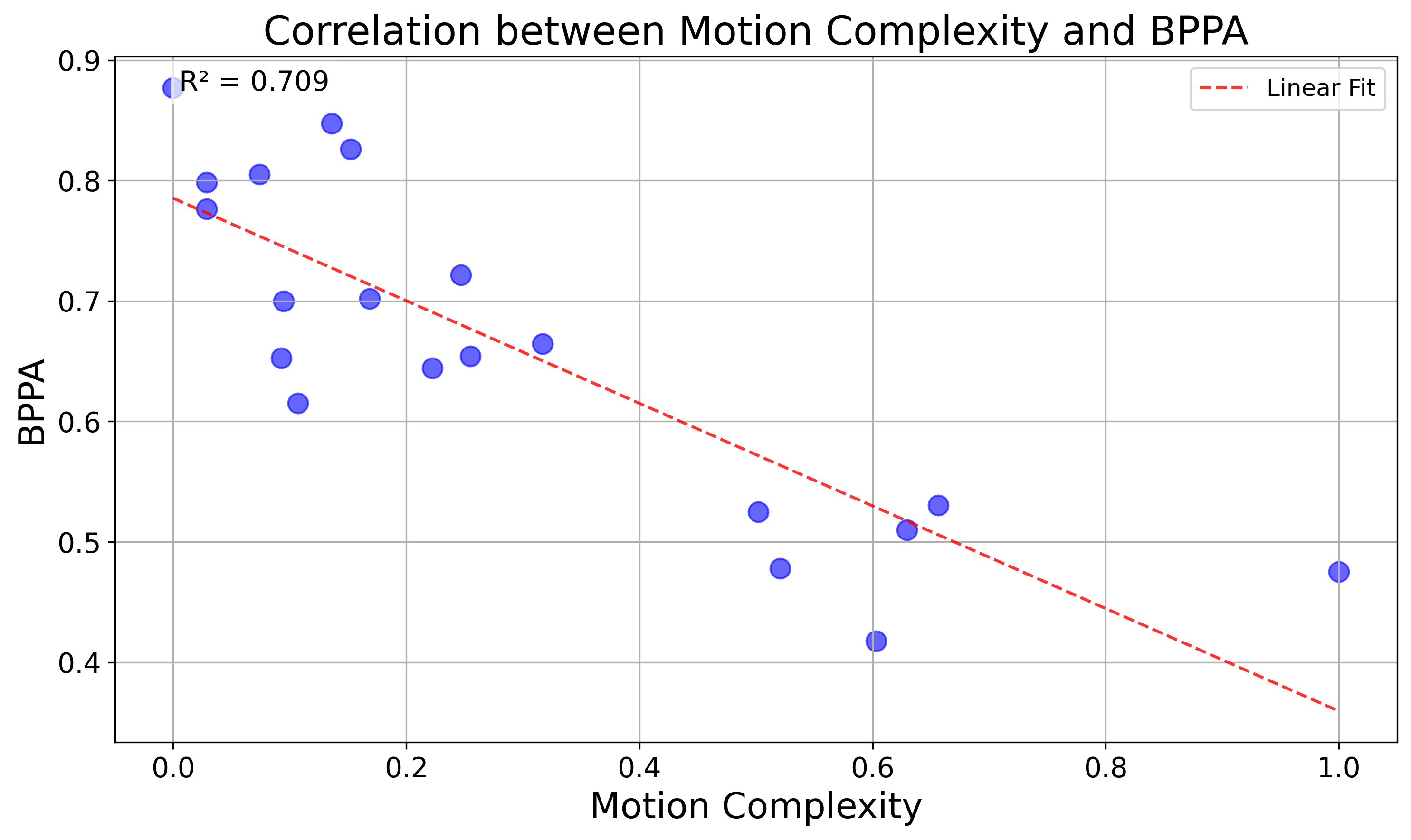}
    \caption{The motion-wise correlation between the motion complexity and BPPA. The motion complexity is calculated as $\sum_{s=1}^{N} \frac{|moved_s|}{|unmoved_s|}$, where $s$ denotes the step number and $|\cdot|$ represents the count of body parts.}
    \label{fig:low_level_planning_motion_complexity_accuracy_correlation}
\end{figure}

\subsection{Generating Raw Avatar Control Parameters}
\label{appendix:raw_control_params}

To generate the raw avatar control parameters, we extend our approach by incorporating the SMPL model specifications into the textual summarization, including the initial joint coordinates and descriptions of the joint rotational directions. For querying strategies, we query the LLM to generate an animation plan by sequentially determining relevant joints, rotational directions, and movement quantities which can directly be executed.

\newpage

\captionof{figure}{Example high-level plan to show imprecise timing (in whole seconds), generated by GPT-4o.}
\label{fig:raw_control_params_timing_0}

\lstset{style=MyListingStyle_0}
\begin{lstlisting}[escapeinside={(*@}{@*)}]
![motion instruction]!
look down to check the time of the watch on the left wrist

![high-level plan]!
[
  {
    "step_number": 1,
    @"time_range": [0, 1]@,
    "movement": "Tilt the head downward.",
    "initial_state": "Head facing forward.",
    "final_state": "Head tilted downward."
  },
  {
    "step_number": 2,
    @"time_range": [1, 2]@,
    "movement": "Move eyes downward to check the time on the watch on the left wrist.",
    "initial_state": "Eyes looking forward.",
    "final_state": "Eyes looking at the left wrist watch."
  }
]
\end{lstlisting}

\captionof{figure}{Example high-level plan to show imprecise timing (in whole and half seconds), generated by GPT-4o.}
\label{fig:raw_control_params_timing_1}

\lstset{style=MyListingStyle_0}
\begin{lstlisting}[escapeinside={(*@}{@*)}]
![motion instruction]!
swing the golf club from right to left

![high-level plan]!
[
  {
    "step_number": 1,
    @"time_range": [0, 1],@
    "movement": "Raise the golf club upwards with a clockwise twist from behind the right shoulder.",
    "initial_state": "The initial position of the golf club is behind the right shoulder, with the right arm extended back.",
    "final_state": "The golf club is raised upwards and begins moving forward."
  },
  {
    "step_number": 2,
    @"time_range": [1, 2.5],@
    "movement": "Swing the golf club downwards and across the front body, moving towards the left side.",
    "initial_state": "The golf club is raised upwards and twisting forward.",
    "final_state": "The golf club is positioned over the left shoulder, with arms crossed in front of the body."
  },
  {
    "step_number": 3,
    @"time_range": [2.5, 3],@
    "movement": "Complete the follow-through of the swing by bringing the club above and behind the left shoulder.",
    "initial_state": "The golf club is over the left shoulder with arms crossed.",
    "final_state": "The golf club has completed its swing and is resting over the left shoulder."
  }
]
\end{lstlisting}

\begin{figure}[htbp]
    \centering
    \begin{subfigure}[b]{0.45\textwidth}
        \centering
        \includegraphics[width=\textwidth]{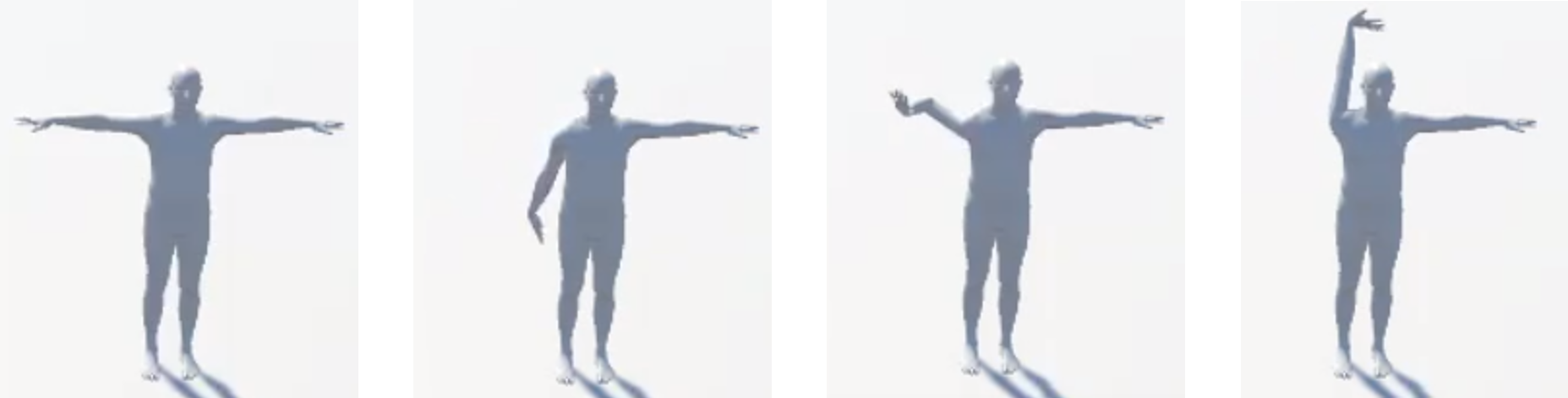}
        \caption{``Toss a ball in the air.''}
        \label{fig:complete_generation_raw_params_case_0_0}
    \end{subfigure}
    \hspace{0.5\textwidth}
    \begin{subfigure}[b]{0.45\textwidth}
        \centering
        \includegraphics[width=\textwidth]{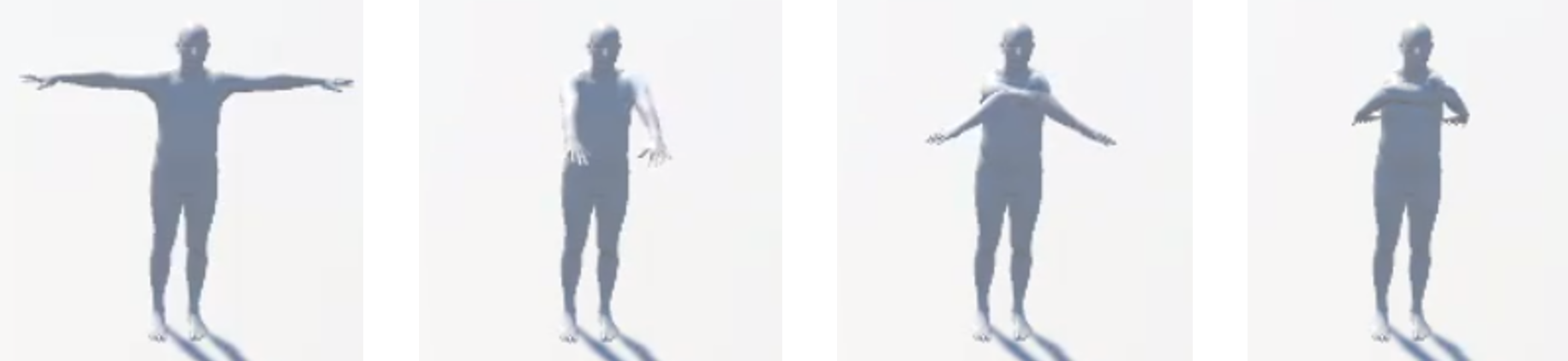}
        \caption{``Hug a person.''}
        \label{fig:complete_generation_raw_params_case_0_1}
    \end{subfigure}
    \caption{Key frames from example animations generated by GPT-4o that struggle to handle SMPL joint rotations, i.e., direction recognition and movement quantity generation. The avatar starts with a T-pose extending arms straight to sides. While the movement directions of joints often deviate from the intended goals, the rotation angles are frequently exaggerated beyond anatomical limits, as shown in the third frame of (a) and the last two frames of (b).}
    \label{fig:complete_generation_raw_params_case_0}
\end{figure}

\begin{figure}[htbp]
    \centering
    \includegraphics[width=\columnwidth]{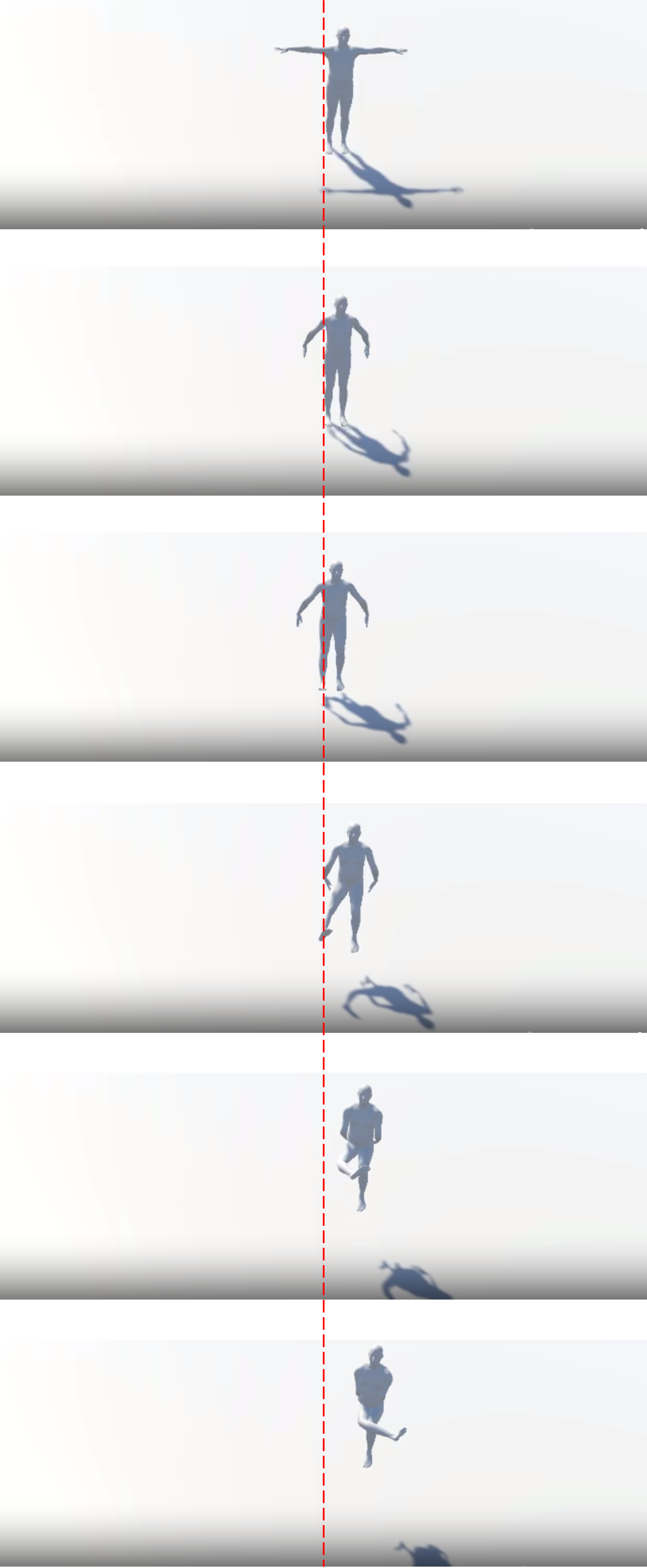}
    \caption{Key frames from an example animation generated by GPT-4o that struggles to handle root movements. The avatar starts with a T-pose extending arms straight to sides. Motion instruction: ``Mount a horse.'' The generated root movements are linearly approximated, resulting in unrealistic global movements that largely deviate from expected real-world values.}
    \label{fig:complete_generation_raw_params_case_1}
\end{figure}

\end{document}